%% file: neurips_2020.tex
\documentclass{article}

    \PassOptionsToPackage{round}{natbib}

\usepackage[final]{neurips_2020}

\usepackage[utf8]{inputenc} 
\usepackage[T1]{fontenc}    
\usepackage{hyperref}       
\usepackage{url}            
\usepackage{booktabs}       
\usepackage{amsfonts}       
\usepackage{nicefrac}       
\usepackage{microtype}      

\usepackage{multirow}
\usepackage{todonotes}

\usepackage{xcolor}

\newcommand{\hbest}[1]{\mathbf{\textcolor{blue}{#1}}}
\newcommand{\hworst}[1]{\mathbf{\textcolor{red}{#1}}}

\usepackage{amsmath,amsfonts,amssymb,amsthm}
\usepackage{marvosym}

\usepackage{gensymb}
\usepackage{subcaption}

\usepackage[ruled, vlined, linesnumbered]{algorithm2e}
\SetKwInput{KwData}{Inputs}
\SetKwInput{KwResult}{Output}

\usepackage{cleveref}
\crefformat{equation}{(#2#1#3)}
\crefformat{figure}{Figure #2#1#3}
\crefformat{table}{Table #2#1#3}
\crefformat{appendix}{Appendix #2#1#3}
\crefformat{section}{Section #2#1#3}

\usepackage{listings}
\usepackage{lmodern}
\definecolor{deepblue}{rgb}{0,0,0.5}
\definecolor{deepred}{rgb}{0.6,0,0}
\definecolor{deepgreen}{rgb}{0,0.5,0}

\newcommand{\lstbg}[3][0pt]{{\fboxsep#1\colorbox{#2}{\strut #3}}}
\lstdefinestyle{pydiff}{
    language=Python,
    basicstyle=\ttfamily\small,
    morecomment=[f][\lstbg{red!20}]-,
    morecomment=[f][\lstbg{green!20}]+,
    morecomment=[f][\textit]{@@},
    frame=tb,
}

\usepackage[super]{nth}
\usepackage{mathtools}
\usepackage{amsmath,amsfonts,amssymb,amsthm, bm, bbm}
\usepackage{textcomp}
\usepackage{nicefrac}       
\usepackage{placeins}  
\usepackage{wrapfig}

\newcommand{\bff}{\mathbf}
\DeclarePairedDelimiterX{\norm}[1]{\lVert}{\rVert}{#1}
\DeclarePairedDelimiter{\abs}{\lvert}{\rvert}
\newcommand{\dset}{{\mathfrak D}}

\newcommand{\ind}{\mathbbm{1}} 
\DeclareMathOperator*{\argmax}{arg\,max}

\newcommand{\balpha}{\bm{\alpha}}
\newcommand{\bbeta}{\boldsymbol{\beta}}

\newcommand{\btheta}{\boldsymbol{\theta}}

\newcommand{\bmu}{\boldsymbol{\mu}}
\newcommand{\bsigma}{\boldsymbol{\sigma}}

\newcommand{\RN}[1]{%
  \textup{\uppercase\expandafter{\romannumeral#1}}%
}
\DeclareMathOperator{\EX}{\mathbb{E}}

\usepackage{tikz}
\usetikzlibrary{arrows, arrows.meta, bayesnet, chains, positioning, decorations.pathreplacing, fit, calc, decorations.pathmorphing}

\usepackage{layouts}  
\tikzset{scaled/.style={scale=#1}}
\tikzset{scaled/.default=1.}

\definecolor{depth1}{HTML}{ff7f0e}
\definecolor{depth2}{HTML}{2ca02c}
\definecolor{depth3}{HTML}{d62728}
\definecolor{depth4}{HTML}{9467bd}
\definecolor{depth5}{HTML}{17becf}
\tikzset{nicecolor/.style={draw=#1!50!black, fill=#1!40!white}}
    
\tikzset{block/.style={rectangle, text width=2ex, align=center, minimum height=2ex, nicecolor=#1, font=\scshape, rounded corners}}
\tikzset{block/.default=red}

\tikzset{opnode/.style={circle, draw, minimum size=1ex, text width=1ex}}
\tikzset{plus/.style ={opnode, font={$+$}}}

\tikzset{line/.style={line width=.15ex, draw=black}}
\tikzset{arrow/.style={-{Latex[length=1.3ex,width=0.8ex]},line}}
\tikzset{arrowsm/.style={-{Latex[length=1.ex,width=0.66ex]},line}}
\tikzset{snake/.style={decoration={snake, pre length=0.01mm, segment length=2mm, amplitude=0.2mm, post length=1mm}, decorate}}

\usepackage[acronym]{glossaries}
\glsdisablehyper

\input{glossary}

\title{Depth Uncertainty in Neural Networks}

\author{%
  Javier Antorán\thanks{equal contribution} \\ 
University of Cambridge \\
\texttt{ja666@cam.ac.uk} \\
\And
James Urquhart Allingham\footnotemark[\value{footnote}] \\ 
University of Cambridge \\
\texttt{jua23@cam.ac.uk} \\
\And
José Miguel Hernández-Lobato \\ 
University of Cambridge \\
Microsoft Research \\
The Alan Turing Institute \\
\texttt{jmh233@cam.ac.uk}
}

\begin{document}

\maketitle


\begin{abstract}
Existing methods for estimating uncertainty in deep learning tend to require multiple forward passes, making them unsuitable for applications where computational resources are limited.
To solve this, we perform probabilistic reasoning over the depth of neural networks. Different depths correspond to subnetworks which share weights and whose predictions are combined via marginalisation, yielding model uncertainty. By exploiting the sequential structure of feed-forward networks, we are able to both evaluate our training objective and make predictions \emph{with a single forward pass}. We validate our approach on real-world regression and image classification tasks. Our approach provides uncertainty calibration, robustness to dataset shift, and accuracies competitive with more computationally expensive baselines. 
\end{abstract}
\section{Introduction}\label{sec:introduction}
Despite the widespread adoption of deep learning, building models that provide robust uncertainty estimates remains a challenge.
This is especially important for real-world applications, where we cannot expect the distribution of observations to be the same as that of the training data.
Deep models tend to be pathologically overconfident, even when their predictions are incorrect \citep{nguyen2015fooled,amodei2016concrete}. If \gls{AI} systems would reliably identify cases in which they expect to underperform, and request human intervention, they could more safely be deployed in medical scenarios \citep{oatml2019bdlb} or self-driving vehicles \citep{fridman2019arguing}, for example. 

In response, a rapidly growing subfield has emerged seeking to build uncertainty aware neural networks \citep{hernandez2015probabilistic,gal2016dropout,lakshminarayanan2017simple}. Regrettably, these methods rarely make the leap from research to production due to a series of shortcomings. \emph{1)~Implementation Complexity:} they can be technically complicated and sensitive to hyperparameter choice. \emph{2)~Computational cost:} they can take orders of magnitude longer to converge than regular networks or require training multiple networks. At test time, averaging the predictions from multiple models is often required. \emph{3)~Weak performance:} they rely on crude approximations to achieve scalability, often resulting in limited or unreliable uncertainty estimates \citep{foong2019pathologies}.

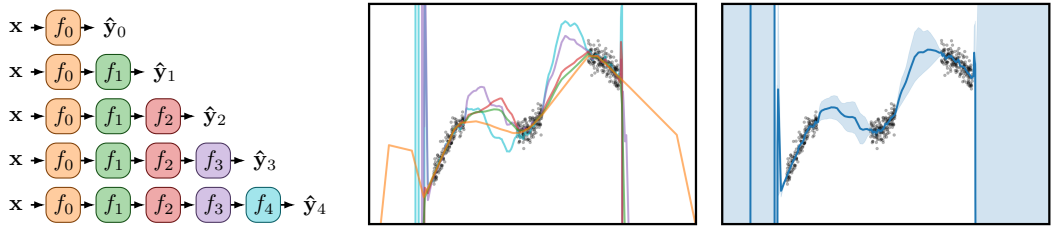
\begin{figure}
    \vspace{-0.1in}
    \begin{subfigure}{0.33 \linewidth}
    \begin{flushleft}
        \begin{tikzpicture}[every node/.style={scale=0.85}]
        	\node[block=depth1] (IB) {$f_0$};
        
        	\node[left=1.3ex of IB] (X) {$\mathbf{x}$};
        	
        	\node[right=1.3ex of IB] (Y) {$\mathbf{\hat y}_0$};
        
        	\draw[arrowsm] (X) -- (IB);
        	\draw[arrowsm] (IB) -- (Y);
        \end{tikzpicture}
        
        \vspace{0.5ex}
        
        \begin{tikzpicture}[every node/.style={scale=0.85}]
        	\node[block=depth1] (IB) {$f_0$};
        	\node[block=depth2, right=1.3ex of IB] (RB1) {$f_1$};
        
        	\node[left=1.3ex of IB] (X) {$\mathbf{x}$};
        	
        	\node[right=1.3ex of RB1] (Y) {$\mathbf{\hat y}_1$};
        
        	\draw[arrowsm] (X) -- (IB);
        	\draw[arrowsm] (IB) -- (RB1);
        	\draw[arrowsm] (RB1) -- (Y);
        \end{tikzpicture}
        
        \vspace{0.5ex}
        
        \begin{tikzpicture}[every node/.style={scale=0.85}]
        	\node[block=depth1] (IB) {$f_0$};
        	\node[block=depth2, right=1.3ex of IB] (RB1) {$f_1$};
        	\node[block=depth3, right=1.3ex of RB1] (RB2) {$f_2$};
        
        	\node[left=1.3ex of IB] (X) {$\mathbf{x}$};
        	
        	\node[right=1.3ex of RB2] (Y) {$\mathbf{\hat y}_2$};
        
        	\draw[arrowsm] (X) -- (IB);
        	\draw[arrowsm] (IB) -- (RB1);
        	\draw[arrowsm] (RB1) -- (RB2);
        	\draw[arrowsm] (RB2) -- (Y);
        \end{tikzpicture}
        
        \vspace{0.5ex}
        
        \begin{tikzpicture}[every node/.style={scale=0.85}]
        	\node[block=depth1] (IB) {$f_0$};
        	\node[block=depth2, right=1.3ex of IB] (RB1) {$f_1$};
        	\node[block=depth3, right=1.3ex of RB1] (RB2) {$f_2$};
        	\node[block=depth4, right=1.3ex of RB2] (RB3) {$f_3$};
        
        	\node[left=1.3ex of IB] (X) {$\mathbf{x}$};
        	
        	\node[right=1.3ex of RB3] (Y) {$\mathbf{\hat y}_3$};
        
        	\draw[arrowsm] (X) -- (IB);
        	\draw[arrowsm] (IB) -- (RB1);
        	\draw[arrowsm] (RB1) -- (RB2);
        	\draw[arrowsm] (RB2) -- (RB3);
        	\draw[arrowsm] (RB3) -- (Y);
        \end{tikzpicture}
        
        \vspace{0.5ex}
        
        \begin{tikzpicture}[every node/.style={scale=0.85}]
        	\node[block=depth1] (IB) {$f_0$};
        	\node[block=depth2, right=1.3ex of IB] (RB1) {$f_1$};
        	\node[block=depth3, right=1.3ex of RB1] (RB2) {$f_2$};
        	\node[block=depth4, right=1.3ex of RB2] (RB3) {$f_3$};
        	\node[block=depth5, right=1.3ex of RB3] (RB4) {$f_4$};
        
        	\node[left=1.3ex of IB] (X) {$\mathbf{x}$};
        	
        	\node[right=1.3ex of RB4] (Y) {$\mathbf{\hat y}_4$};
        
        	\draw[arrowsm] (X) -- (IB);
        	\draw[arrowsm] (IB) -- (RB1);
        	\draw[arrowsm] (RB1) -- (RB2);
        	\draw[arrowsm] (RB2) -- (RB3);
        	\draw[arrowsm] (RB3) -- (RB4);
        	\draw[arrowsm] (RB4) -- (Y);
        \end{tikzpicture}
        
        \vspace{0.5ex}
        
    \end{flushleft}
    \end{subfigure}
    \begin{subfigure}{0.66\linewidth}
    \includegraphics[width=\linewidth]{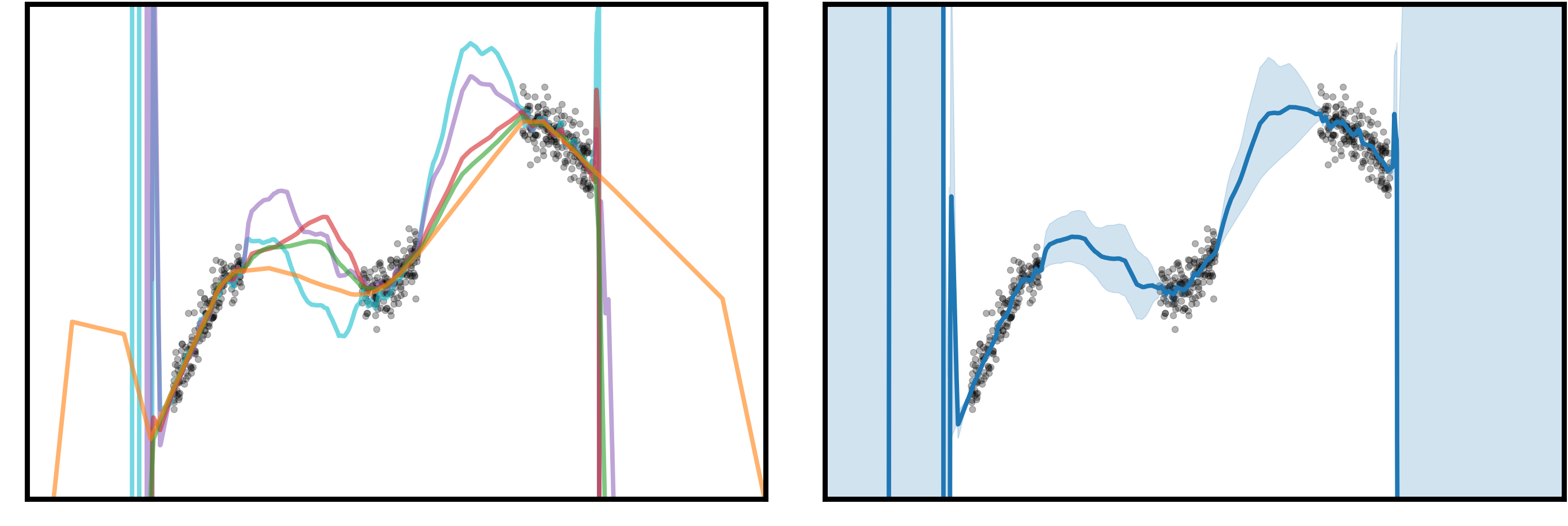}
    \end{subfigure}
    \caption{A \gls{DUN} is composed of subnetworks of increasing depth (\emph{left}, colors denote layers with shared parameters). These correspond to increasingly complex functions (\emph{centre}, colors denote depth at which predictions are made). Marginalising over depth yields model uncertainty through disagreement of these functions (\emph{right}, error bars denote 1 std. dev.).}
    \label{fig:layerwise_DUN}
    \vspace{-0.15in}
\end{figure}

In this work, we introduce \glspl{DUN}, a probabilistic model that treats the depth of a \gls{NN} as a random variable over which to perform inference. In contrast to more typical weight-space approaches for Bayesian inference in \glspl{NN}, ours reflects a lack of knowledge about how deep our network should be. We treat network weights as learnable hyperparameters. In \glspl{DUN}, marginalising over depth is equivalent to performing \gls{BMA} over an ensemble of progressively deeper \glspl{NN}. As shown in \cref{fig:layerwise_DUN}, \glspl{DUN} exploit the overparametrisation of a single deep network to generate diverse explanations of the data. The key advantages of \glspl{DUN} are:
\begin{enumerate}
    \item \emph{Implementation simplicity}:  requiring only minor additions to vanilla deep learning code, and no changes to the hyperparameters or training regime. 
    \item \emph{Cheap deployment}: computing exact predictive posteriors with a single forward pass.
    \item \emph{Calibrated uncertainty}: our experiments show that \glspl{DUN} are competitive with strong baselines in terms of predictive performance, \gls{OOD} detection and robustness to corruptions.
\end{enumerate}


    

\section{Related Work}\label{sec:related_work}
 
Traditionally, Bayesians tackle overconfidence in deep networks by treating their weights as random variables. Through marginalisation, uncertainty in weight-space is translated to predictions. Alas, the weight posterior in \glspl{BNN} is intractable. Hamiltonian Monte Carlo \citep{neal1995bayesian} remains the gold standard for inference in \glspl{BNN} but is limited in scalability. The Laplace approximation \citep{mackay1992practical,ritter2018scalable}, \gls{VI} \citep{hinton1993keeping,graves2011practical,blundell2015weight} and expectation propagation \citep{hernandez2015probabilistic} have all been proposed as alternatives. More recent methods are scalable to large models \citep{khan2018fast,osawa2019practical,dusenberry2020efficient}. \citet{gal2016dropout} re-interpret dropout as \gls{VI}, dubbing it MC Dropout. Other stochastic regularisation techniques can also be viewed in this light \citep{kingma2015variational,Gal2016Uncertainty,teye2018bayesian}. These can be seamlessly applied to vanilla networks. Regrettably, most of the above approaches rely on factorised, often Gaussian, approximations resulting in pathological overconfidence \citep{foong2019pathologies}.

It is not clear how to place reasonable priors over network weights \citep{wenzel2020good}. 
\glspl{DUN} avoid this issue by targeting depth. \gls{BNN} inference can also be performed directly in function space \citep{hafner2018reliable,sun2018functional,ma2019variational,wang2018function}. However, this requires crude approximations to the KL divergence between stochastic processes. The equivalence between infinitely wide \glspl{NN} and \glspl{GP} \citep{neal1995bayesian,matthews2018gaussian,garriga2018deep} can be used to perform exact inference in \glspl{BNN}. Unfortunately, exact \gls{GP} inference scales poorly in dataset size. 


Deep ensembles is a non-Bayesian method for uncertainty estimation in \glspl{NN} that trains multiple independent networks and aggregates their predictions \citep{lakshminarayanan2017simple}. Ensembling provides very strong results but is limited by its computational cost. \citet{huang2017snapshot}, \citet{garipov2018loss}, and \citet{maddox2019simple} reduce the cost of training an ensemble by leveraging different weight configurations found in a single SGD trajectory. However, this comes at the cost of reduced predictive performance~\citep{ashukha2020pitfalls}.
Similarly to deep ensembles, \glspl{DUN} combine the predictions from a set of deep models. However, this set stems from treating depth as a random variable. Unlike ensembles, \gls{BMA} assumes the existence of a single correct model \citep{minka2000bayesian}. In \glspl{DUN}, uncertainty arises due to a lack of knowledge about how deep the correct model is. It is worth noting that deep ensembles can also be interpreted as approximate \gls{BMA} \citep{wilson2020case}.

All of the above methods, except \glspl{DUN}, require multiple forward passes to produce uncertainty estimates. This is problematic in low-latency settings or those in which computational resources are limited.
Note that certain methods, such as MC Dropout, can be parallelised via batching. This allows for some computation time / memory usage trade-off.
Alternatively, \citet{postels2019sampling} use error propagation to approximate the dropout predictive posterior with a single forward pass. Although efficient, this approach shares pathologies with MC Dropout. \citet{van2020simple} combine deep RBF networks with a Jacobian regularisation term to deterministically detect \gls{OOD} points. \citet{nalisnick2019hybrid} and \citet{meinke2020towards} use generative models to detect \gls{OOD} data without multiple predictor evaluations. Unfortunately, deep generative models can be unreliable for \gls{OOD} detection \citep{nalisnick2018do} and simpler alternatives might struggle to scale.

There is a rich literature on probabilistic inference for \gls{NN} structure selection, starting with the Automatic Relevance Detection prior \citep{mackay1994bayesian}. Since then, a number of approaches have been introduced \citep{lawrence2002note,ghosh2019model}. Perhaps the closest to our work is that of \citet{nalisnick2019dropout}, which interprets dropout as a \emph{structured shrinkage prior} that reduces the influence of residual blocks in ResNets. Conversely, a \gls{DUN} can be constructed for any feed-forward neural network and marginalizes predictions at different depths. 
Similar work from \citet{dikov2019bayesian} uses \gls{VI} to learn both the width and depth of a \gls{NN} by leveraging continuous relaxations of discrete probability distributions. For depth, they use a Bernoulli distribution to model the probability that any layer is used for a prediction.
In contrast to their approach, \glspl{DUN} use a Categorical distribution to model depth, do not require sampling for evaluation of the training objective or making predictions, and can be applied to a wider range of \gls{NN} architectures, such as \glspl{CNN}.
\citet{huang2016deep} stochastically drop layers as a ResNet training regularisation approach. On the other hand, \glspl{DUN} perform exact marginalisation over architectures at train and test time, translating depth uncertainty into uncertainty over a broad range of functional complexities.

\glspl{DUN} rely on two insights that have recently been demonstrated elsewhere in the literature.
The first is that a single over-parameterised \gls{NN} is capable of learning multiple, diverse, representations of a dataset. This is also a key insight for a subsequent work MIMO~\citep{havasi2020training}. The second is that ensembling \glspl{NN} with varying hyperparameters, in our case depth, leads to improved robustness in predictions. Concurrently to our work, hyper-deep ensembles~\citep{wenzel2020hyper} demonstrate this property for a large range of parameters.


\section{\glsentrydescplural{DUN}}\label{sec:DUNs}

\begin{figure}
\vspace{-0.1in}
\centering
\begin{subfigure}[b]{0.18\textwidth}
    \centering
    \begin{tikzpicture}[every node/.style={scale=1}]
      \node[obs] (xn) {\scriptsize $\mathbf{x}^{(n)}$};
      \node[latent, below=4.5ex of xn] (yn) {\scriptsize $\mathbf{y}^{(n)}$};
      \node[const, left=3ex of yn] (theta) {\footnotesize $\theta \ $};
      \node[latent, right=3ex of yn] (d) {\footnotesize $d$};
      \node[const, above=3ex of d] (beta) {\footnotesize $\boldsymbol{\beta}$};
      
      \edge[arrow] {xn} {yn} ; %
      \edge[arrow] {theta} {yn} ;
      \edge[arrow] {d} {yn} ;
      \edge[arrow] {beta} {d} ;
      
      \plate[inner sep=1.7ex] {} {(yn)(xn)} {\scriptsize $N$} ;
    \end{tikzpicture}
\end{subfigure}
\quad
\begin{subfigure}[b]{0.763\textwidth}
    \centering
    \begin{tikzpicture}[every node/.style={scale=0.85}]
    	\node[block=depth1] (RB0) {$f_{0}$};
    	\node[block=depth2, below right=.0ex and 4ex of RB0] (RB1) {$f_{1}$};
    	\node[block=depth3, below right=.0ex and 4ex of RB1] (RB2) {$f_{2}$};
    	\node[block=depth4, below right=.0ex and 4ex of RB2] (RB3) {$f_{3}$};
    	\node[block=depth5, below right=.0ex and 4ex of RB3] (RB4) {$f_{4}$};
    	\node[block=teal!80!white, below right=2ex and 7ex of RB4, text width=2.5ex] (RBD) {$f_{D}$};

    	\node[left=4ex of RB0] (X) {$\mathbf{x}$};

    	\draw[arrow] (X) -- (RB0);
    	\draw[arrow] (RB0) |- (RB1);
    	\draw[arrow] (RB1) |- (RB2);
    	\draw[arrow] (RB2) |- (RB3);
    	\draw[arrow] (RB3) |- (RB4);
    	\draw[line] (RB4) -- ([yshift=-.5ex] RB4.south);
    	\draw[line, dashed] ([yshift=-.5ex] RB4.south) |- ([xshift=-4.ex] RBD.west);
    	\draw[arrow] ([xshift=-4.ex] RBD.west) -- (RBD);
    
        \draw[arrow, snake] (RBD) -- ([xshift=4ex] RBD.east) node[coordinate] (end) {};
        \draw[arrow, snake] (RB0) -- (RB0 -| end);
        \draw[arrow, snake] (RB1) -- (RB1 -| end);
        \draw[arrow, snake] (RB2) -- (RB2 -| end);
        \draw[arrow, snake] (RB3) -- (RB3 -| end);
        \draw[arrow, snake] (RB4) -- (RB4 -| end);
        
        \draw [decorate,decoration={brace,amplitude=1ex, raise=.5ex},xshift=2ex, thick] (RB0 -| end.west) -- (end) node [block=blue!60!white, midway, xshift=6ex, text width=5ex, thin] (OB) {$f_{D+1}$};
        
    	\node[right=4ex of OB] (YD) {$\mathbf{\hat y}_{i}$};
        	
    	\draw[arrow] (OB.east) -- (YD);
         
    \end{tikzpicture}
\end{subfigure}
\caption{Left: graphical model under consideration. Right: computational model. Each layer's activations are passed through the output block, producing per-depth predictions. }
    \label{fig:nn_structure_graph_model}
\vspace{-0.15in}
\end{figure}
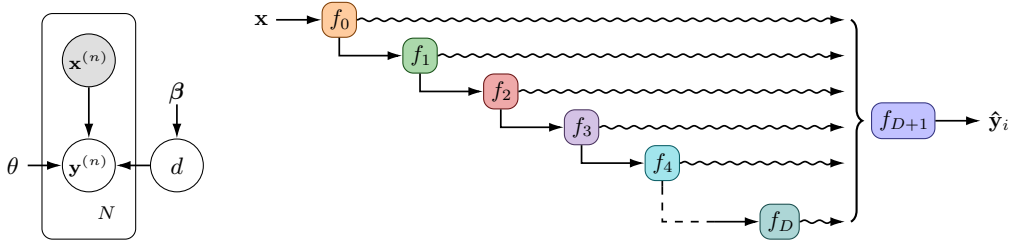

Consider a dataset $\dset\,{=}\,\{\bff{x}^{(n)}, \bff{y}^{(n)}\}_{n=1}^{N}$ and a neural network composed of an input block $f_{0}(\cdot)$, $D$ intermediate blocks $\{f_{i}(\cdot)\}_{i=1}^{D}$, and an output block $f_{D+1}(\cdot)$. Each block is a group of one or more stacked linear and non-linear operations.
The activations at depth $i \in [0, D]$, $\bff{a}_{i}$, are obtained recursively as $\bff{a}_{i}\,{=}\, f_{i}(\bff{a}_{i-1})$, $\bff{a}_{0}\,{=}\, f_{0}(\bff{x})$.

A forward pass through the network is an iterative process, where each successive block $f_{i}(\cdot)$ refines the previous block's activation. Predictions can be made at each step of this procedure by applying the output block to each intermediate block's activations: $\hat{\bff{y}}_{i} = f_{D+1}(\bff{a}_{i})$. This computational model is displayed in \cref{fig:nn_structure_graph_model}. It can be implemented by changing 8 lines in a vanilla PyTorch \gls{NN}, as shown in \cref{app:dun_impl}. Recall, from \cref{fig:layerwise_DUN}, that we can leverage the disagreement among intermediate blocks' predictions to quantify model uncertainty.


\subsection{Probabilistic Model: Depth as a Random Variable}\label{sec:prob_model}

We place a categorical prior over network depth $p_{\bbeta}(d)\,{=}\,\mathrm{Cat}(d| \{\beta_{i}\}_{i=0}^{D})$. Referring to network weights as $\btheta$, we parametrise the likelihood for each depth using the corresponding subnetwork's output: $p(\bff{y}|\bff{x}, d{=}i; \btheta) = p(\bff{y}|f_{D+1}(\bff{a}_{i}; \btheta))$. A graphical model is shown in \cref{fig:nn_structure_graph_model}. For a given weight configuration, the likelihood for every depth, and thus our model's \gls{MLL}:
\begin{align}\label{eq:marginal_likelihood}
\log p(\dset; \btheta) = \log \sum^{D}_{i=0}\left(p_{\bbeta}(d{=}i)\cdot\prod_{n=1}^{N} p(\bff{y}^{(n)}|\bff{x}^{(n)}, d{=}i; \btheta)\right),
\end{align}
can be obtained with a \emph{single forward pass} over the training set by exploiting the sequential nature of feed-forward \glspl{NN}. The posterior over depth, ${p(d|\dset; \btheta){=}p(\dset | d; \btheta)p_{\bbeta}(d)/p(\dset; \btheta)}$ is a categorical distribution that tells us about how well each subnetwork explains the data. 

A key advantage of deep neural networks lies in their capacity for automatic feature extraction and representation learning. For instance, \citet{zeiler2014visualizing} demonstrate that \glspl{CNN} detect successively more abstract features in deeper layers. 
Similarly, \citet{frosst2019analyzing} find that maximising the entanglement of different class representations in intermediate layers yields better generalisation. Given these results, using all of our network's intermediate blocks for prediction might be suboptimal. Instead, we infer whether each block should be used to learn representations or perform predictions, which we can leverage for ensembling, by treating network depth as a random variable. As shown in \cref{fig:maintext_VI_MLE_training}, subnetworks too shallow to explain the data are assigned low posterior probability; they perform feature~extraction.

\subsection{Inference in DUNs}\label{sec:inference}
We consider learning network weights by directly maximising \cref{eq:marginal_likelihood} with respect to $\btheta$, using backpropagation and the \textit{log-sum-exp} trick. In \cref{app:vI_vs_MLE}, we show that the gradients of \cref{eq:marginal_likelihood} reaching each subnetwork are weighted by the corresponding depth's posterior mass. This leads to local optima where all but one subnetworks' gradients vanish. The posterior collapses to a delta function over an arbitrary depth, leaving us with a deterministic \gls{NN}. 
When working with large datasets, one might indeed expect the true posterior over~depth~to~be a delta.
However, because modern \glspl{NN} are underspecified even for large datasets, multiple depths should be able to explain the data simultaneously (shown in \cref{fig:maintext_VI_MLE_training} and \cref{app:vI_vs_MLE}).

We can avoid the above pathology by decoupling the optimisation of network weights $\btheta$ from the posterior distribution. In latent variable models, the \gls{EM} algorithm \citep{bishop_prml} allows us to optimise the \gls{MLL} by iteratively computing~$p(d | \dset; \btheta)$~and then updating $\btheta$. 
We propose to use stochastic gradient variational inference as an alternative more amenable to \gls{NN} optimisation. We introduce a surrogate categorical distribution over depth $q_{\balpha}(d) \,{=}\, \mathrm{Cat}(d| \{\alpha_{i}\}_{i=0}^{D})$. In \cref{app:Derivation}, we derive the following lower bound on \cref{eq:marginal_likelihood}:
\begin{gather}\label{eq:var_objective}
    \log p(\dset; \btheta) \geq \mathcal{L}(\balpha, \btheta) = \sum_{n=1}^{N} \EX_{q_{\balpha}(d)}\left[ \log p(\bff{y}^{(n)}|\bff{x}^{(n)}, d; \btheta)\right] - \text{KL}( q_{\balpha}(d)\,\|\,p_{\bbeta}(d)).
\end{gather} 
This \gls{ELBO} allows us to optimise the variational parameters $\balpha$ and network weights $\btheta$ simultaneously using gradients. Because both our variational and true posteriors are categorical, \cref{eq:var_objective} is convex with respect to $\balpha$. At the optima, $q_{\balpha}(d)\,{=}\,p(d|\dset; \btheta)$ and the bound is tight. Thus, we perform exact rather than approximate inference.


$\EX_{q_{\balpha}(d)}[\log p(\bff{y}|\bff{x}, d; \btheta)]$ can be computed from the activations at every depth. Consequently, both terms in \cref{eq:var_objective} can be evaluated exactly, with only a single forward pass. This removes the need for high variance Monte Carlo gradient estimators, often required by \gls{VI} methods for \glspl{NN}. When using mini-batches of size $B$, we stochastically estimate the \gls{ELBO} in \cref{eq:var_objective} as
\begin{gather}\label{eq:detailed_objective}
    \mathcal{L}(\balpha, \btheta) \approx \frac{N}{B} \sum^{B}_{n=1} \sum_{i=0}^{D} \left(\log p(\bff{y}^{(n)} | \bff{x}^{(n)}, d{=}i; \btheta) \cdot \alpha_{i} \right) - \sum_{i=0}^{D} \left( \alpha_{i} \log \frac{\alpha_{i}}{\beta_{i}} \right).
\end{gather}
Predictions for new data $\bff{x}^{*}$ are made by marginalising depth with the variational posterior: 
\begin{align}\label{eq:marginalisation}
    p(\bff{y}^{*} | \bff{x}^{*}, \dset; \btheta) = \sum^{D}_{i=0} p(\bff{y}^{*} | \bff{x}^{*}, d{=}i; \btheta) q_{\balpha}(d{=}i).
\end{align}

\section{Experiments}\label{sec:experiments}
First, we compare the \gls{MLL} and \gls{VI} training approaches for \glspl{DUN}. We then evaluate \glspl{DUN} on toy-regression, real-world regression, and image classification tasks. 
As baselines, we provide results for vanilla \glspl{NN} (denoted as `SGD'), MC Dropout \citep{gal2016dropout}, and deep ensembles \citep{lakshminarayanan2017simple}, arguably the strongest approach~for uncertainty estimation in deep learning \citep{snoek1029can,ashukha2020pitfalls}. For regression tasks, we also include Gaussian \gls{MFVI} \citep{blundell2015weight} with the local reparametrisation trick \citep{kingma2015variational}. 
For the image classification tasks, we include stochastic depth ResNets (S-ResNets) \citep{huang2016deep}, which can be viewed as MC Dropout applied to whole residual blocks, and deep ensembles of different depth networks (depth-ensembles). We include the former as an alternate method for converting uncertainty over depth into predictive uncertainty. We include the latter to investigate the hypothesis that the different classes of functions produced at different depths help provide improved disagreement and, in turn, predictive uncertainty. 
We study all methods in terms of accuracy, uncertainty quantification, and robustness to corrupted or \gls{OOD} data. We place a uniform prior over DUN depth. See \cref{app:computing_uncertainties}, \cref{app:evaluating_uncertainties}, and \cref{app:experimental_setup} for detailed descriptions of the techniques we use to compute, and evaluate uncertainty estimates, and our experimental setup, respectively. Additionally, in \cref{app:duns_for_nas}, we explore the effects of architecture hyperparameters on the depth posterior and the use of DUNs for architecture search. Code is available at \href{https://github.com/cambridge-mlg/DUN}{\texttt{https://github.com/cambridge-mlg/DUN}}.

\subsection{Comparing MLL and VI training}\label{sec:VI_vs_MLE}
\begin{figure}[t]
\vspace{-0.15in}
 \centering
    \includegraphics[width=0.85\linewidth]{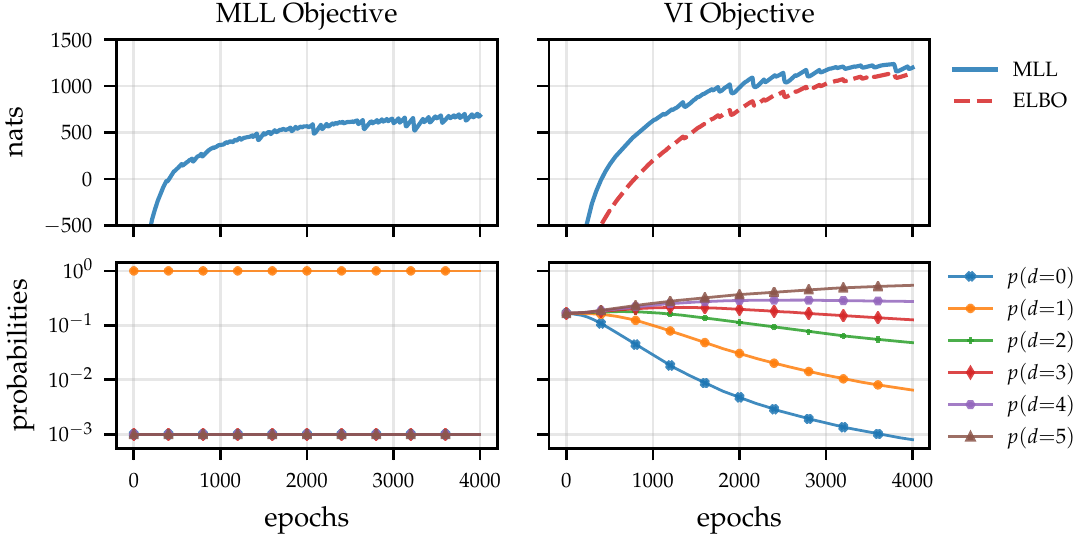}
    \vspace{-0.05in}
    \caption{Top row: progression of \gls{MLL} and \gls{ELBO} during training. Bottom: progression of all six depth posterior probabilities. The left column corresponds to optimising the \gls{MLL} directly and the right to \gls{VI}. For the latter, variational posterior probabilities $q(d)$ are shown.}
    \label{fig:maintext_VI_MLE_training}
\vspace{-0.15in}
\end{figure}

\Cref{fig:maintext_VI_MLE_training} compares the optimisation of a 5 hidden layer fully connected \gls{DUN} on the concrete dataset using estimates of the MLL \cref{eq:marginal_likelihood} and ELBO \cref{eq:detailed_objective}. The former approach converges to a local optima where all but one depth's probabilities go to 0. With \gls{VI}, the surrogate posterior converges slower than the network weights. This allows $\btheta$ to reach a configuration where multiple depths can be used for prediction. Towards the end of training, the variational gap vanishes. The surrogate distribution approaches the true posterior without collapsing to a delta. The \gls{MLL} values obtained with \gls{VI} are larger than those obtained with \cref{eq:marginal_likelihood}, i.e. our proposed approach finds better explanations for the data. In \cref{app:vI_vs_MLE}, we optimise \cref{eq:marginal_likelihood} after reaching a local optima with VI \cref{eq:detailed_objective}. This does not cause posterior collapse, showing that \gls{MLL} optimisation's poor performance is due to a propensity for poor local optima.

\subsection{Toy Datasets}\label{sec:toy_datasets}

\begin{figure}[t]
\vspace{-0.1in}
 \centering
    \includegraphics[width=\linewidth]{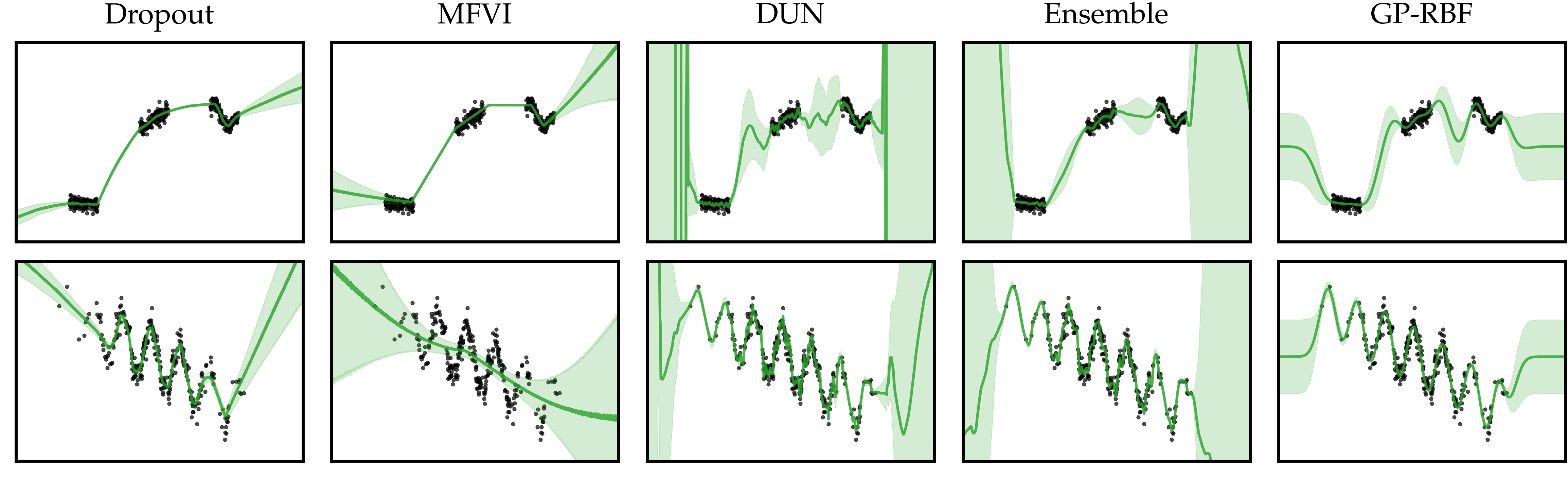}
    \vspace{-0.15in}
    \caption{Top row: toy dataset from \cite{izmailov2019subspace}.
    Bottom: Wiggle dataset. Black dots denote data points. Error bars represent standard deviation among mean predictions.
    }
    \label{fig:maintext_toy}
\vspace{-0.15in}
\end{figure}

We consider two synthetic 1D datasets, shown in \cref{fig:maintext_toy}. We use 3 hidden layer, 100 hidden unit, fully connected networks with residual connections for our baselines.
DUNs use the same architecture but with 15 hidden layers. GPs use the RBF kernel. We found these configurations to work well empirically. 
In \cref{app:toy_datasets}, we perform experiments with different toy datasets, architectures and hyperparameters. \glspl{DUN}' performance increases with depth but often 5 layers are sufficient to produce reasonable uncertainty estimates.


The first dataset, which is taken from \cite{izmailov2019subspace}, contains three disjoint clusters of data.
Both MFVI and Dropout present error bars that are similar in the data dense and in-between regions. MFVI underfits slightly, not capturing smoothness in the data. DUNs  perform most similarly to Ensembles. They are both able to fit the data well and express~in-between uncertainty. Their error bars become large very quickly in the extrapolation regime due to different ensemble elements' and depths' predictions diverging in different directions. 

Our second dataset consists of 300 samples from $y{=}\sin(\pi x){+}0.2 \cos(4 \pi x){-}0.3 x{+}\epsilon$, where $\epsilon \sim \mathcal{N}(0, 0.25)$ and $x \sim \mathcal{N}(5, 2.5)$. We dub it ``Wiggle''. Dropout struggles to fit this faster varying function outside of the data-dense regions. MFVI fails completely. \glspl{DUN} and Ensembles both fit the data well and provide error bars that grow as the data becomes~sparse. 

\subsection{Tabular Regression} \label{sec:tabresres}

We evaluate all methods on UCI regression datasets using standard \citep{hernandez2015probabilistic} and gap splits \citep{foong2019inbetween}. We also use the large-scale non-stationary flight delay dataset, preprocessed by \citet{hensman2013Gaussian}. Following \citet{deisenroth2015distributed}, we train on the first 2M data points and test on the subsequent 100k. We select all hyperparameters, including \gls{NN} depth, using Bayesian optimisation with HyperBand \citep{falkner2018bohb}. See \cref{app:setup_regression} for details. We evaluate methods with \gls{RMSE}, \gls{LL} and \gls{TCE}. The latter measures the calibration of the $10\%$ and $90\%$ confidence intervals, and is described in \cref{app:evaluating_uncertainties}.

\begin{figure}[t]
\vspace{-0.1in}
    \centering
    \includegraphics[width=\textwidth]{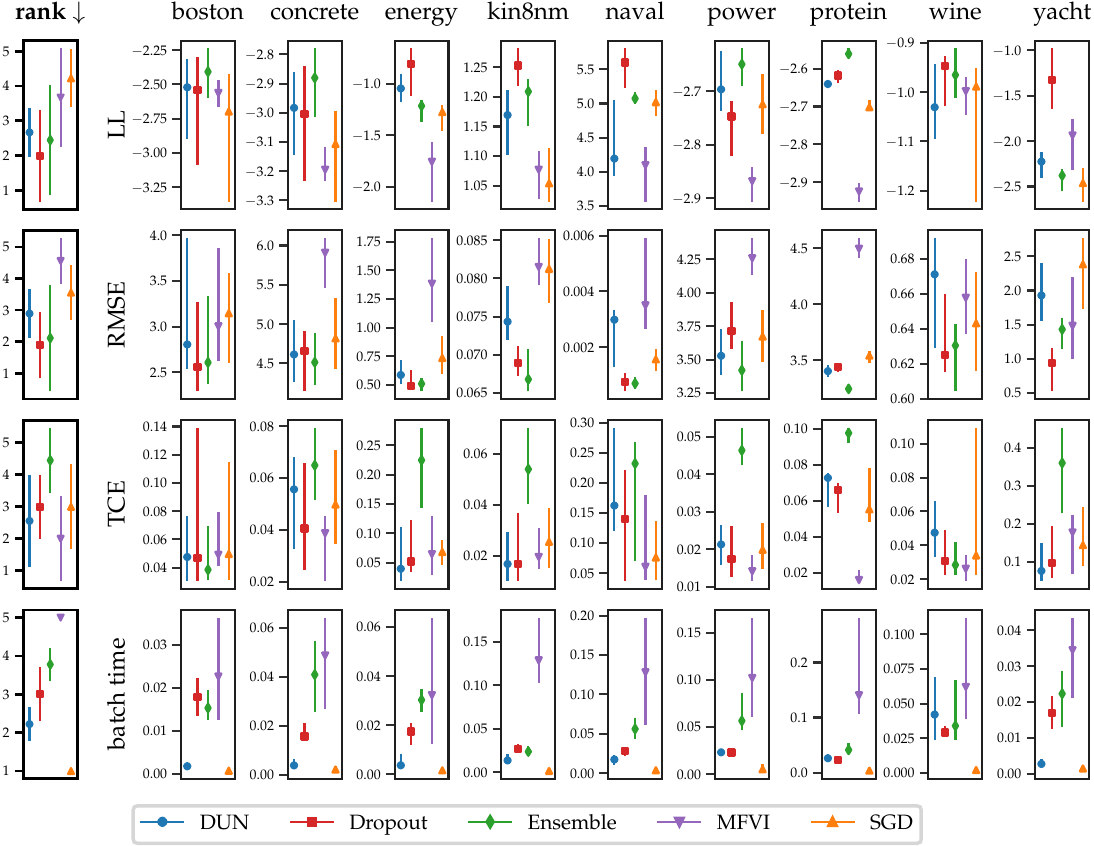}
    \caption{Quartiles for results on UCI regression datasets across standard splits. Average ranks are computed across datasets. For LL, higher is better. Otherwise, lower is better.}
    \label{fig:maintext_UCI_regression}
    \vspace{-0.15in}
\end{figure}

UCI standard split results are found in \cref{fig:maintext_UCI_regression}. For each dataset and metric, we rank methods from 1 to 5 based on mean performance. We report mean ranks and standard deviations. Dropout obtains the best mean rank in terms of \gls{RMSE}, followed closely by Ensembles. \glspl{DUN} are third, significantly ahead of MFVI and SGD. Even so, \glspl{DUN} outperform Dropout and Ensembles in terms of \gls{TCE}, i.e. \glspl{DUN} more reliably assign large error bars to points on which they make incorrect predictions. Consequently, in terms of LL, a metric which considers both uncertainty and accuracy, \glspl{DUN} perform competitively (the LL rank distributions for all three methods overlap almost completely). MFVI provides the best calibrated uncertainty estimates. Despite this, its mean predictions are inaccurate, as evidenced by it being last in terms of RMSE. This leads to \gls{MFVI}'s LL rank only being better than SGD's. Results for gap splits, designed to evaluate methods' capacity to express in-between uncertainty, are given in \cref{app:additional_regression_results}. Here, \glspl{DUN} outperform Dropout in terms of LL rank. However, they are both outperformed by MFVI and ensembles.

The flights dataset is known for strong covariate shift between its train and test sets, which are sampled from contiguous time periods. \gls{LL} values are strongly dependent on calibrated uncertainty. As shown in \cref{tab:flights}, \glspl{DUN}' RMSE is similar to that of SGD, with Dropout and Ensembles performing best. Again, \glspl{DUN} present superior uncertainty calibration. This allows them to achieve the best \gls{LL}, tied with Ensembles and Dropout. We speculate that \glspl{DUN}' calibration stems from being able to perform exact inference, albeit in depth space.

In terms of prediction time, \glspl{DUN} clearly outrank Dropout, Ensembles, and MFVI on UCI.
Due to depth, or maximum depth $D$ for \glspl{DUN}, being chosen with Bayesian optimisation, methods' batch times vary across datasets.
\glspl{DUN} are often deeper because the quality of their uncertainty estimates improves with additional explanations of the data. As a result, SGD clearly outranks \glspl{DUN}. On flights, increased depth causes \glspl{DUN}' prediction time to lie in between Dropout's and Ensembles'.

\begin{table}[t]
\centering
\vspace{-0.1in}
\caption{Results obtained on the flights dataset (2M). Mean and standard deviation values are computed across 5 independent training runs.}
\small
\begin{tabular}{@{}l|ccccc@{}}
\toprule
\textsc{Metric} & \textsc{DUN}                           & \textsc{Dropout}                      & \textsc{Ensemble}                     & \textsc{MFVI}                        & \textsc{SGD}                         \\ \midrule
LL              & $\mathbf{-4.95 \scriptstyle \pm 0.01}$ & $\mathbf{-4.95 \scriptstyle \pm0.02}$ & $\mathbf{-4.95 \scriptstyle \pm0.01}$ & $-5.02 \scriptstyle \pm0.05$         & $-4.97 \scriptstyle \pm0.01$         \\
RMSE            & $34.69 \scriptstyle \pm0.28$           & $\mathbf{34.28 \scriptstyle \pm0.11}$ & $34.32 \scriptstyle \pm0.13$          & $36.72 \scriptstyle \pm1.84$         & $34.61 \scriptstyle \pm0.19$         \\
TCE             & $.087 \scriptstyle \pm.009$            & $.096 \scriptstyle \pm.017$           & $.090 \scriptstyle \pm.008$           & $\mathbf{.068 \scriptstyle \pm.014}$ & $.084 \scriptstyle \pm.010$          \\
Time            & $.026 \scriptstyle \pm.001$            & $.016 \scriptstyle \pm.001$           & $.031 \scriptstyle \pm.001$           & $.547 \scriptstyle \pm.003$          & $\mathbf{.002 \scriptstyle \pm.000}$ \\ \bottomrule
\end{tabular}
\label{tab:flights}
\vspace{-0.2cm}
\end{table}

\subsection{Image Classification}\label{sec:imgres}

\begin{figure}[t]
    \centering
    \vspace{-0.15in}
    \includegraphics[width=\linewidth]{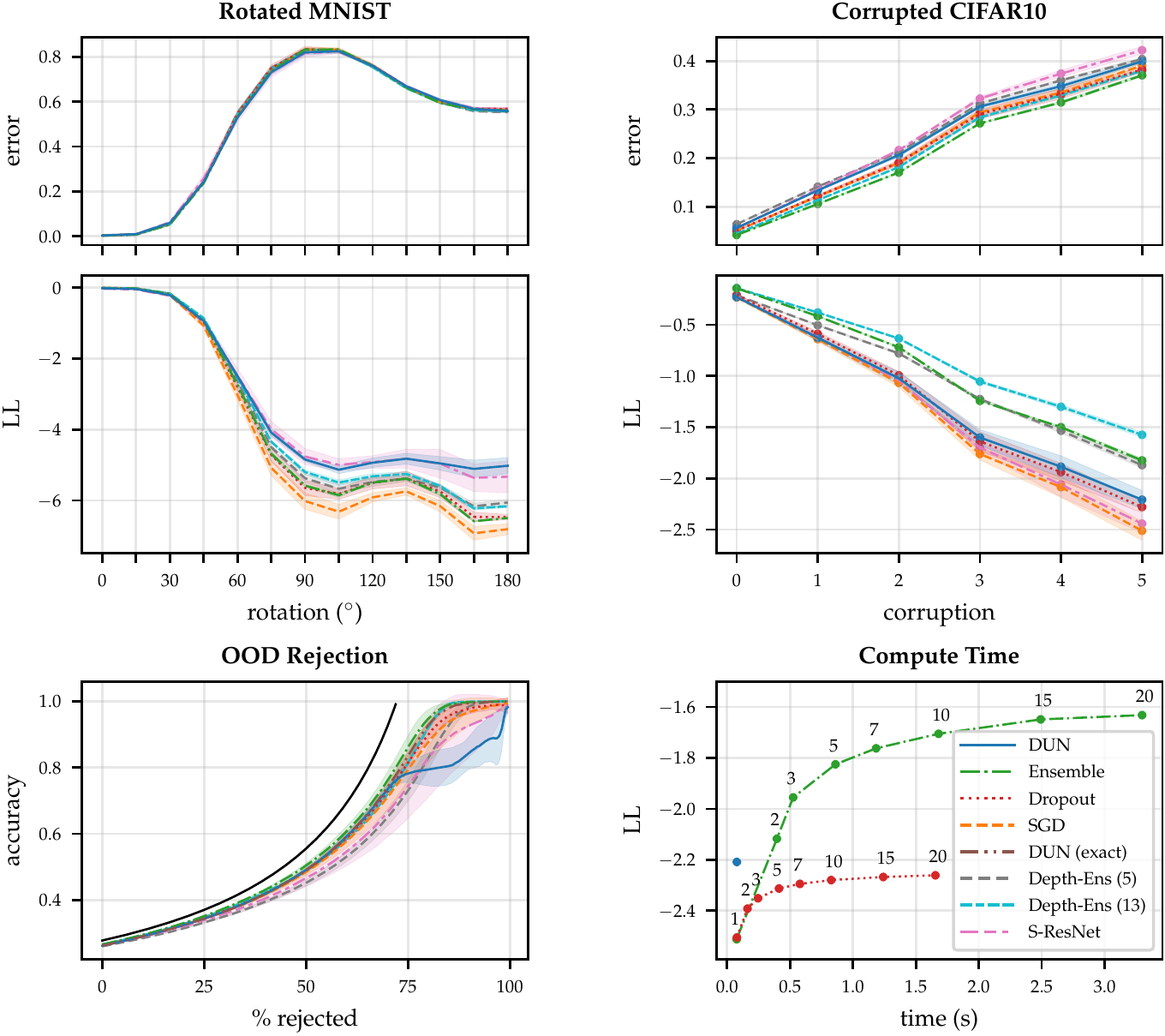}
    \vspace{-0.1in}
    \caption{Top left: error and \gls{LL} for MNIST at varying degrees of rotation. Top right: error and \gls{LL} for CIFAR10 at varying corruption severities. Bottom left: CIFAR10-SVHN rejection-classification plot. The black line denotes the theoretical maximum performance; all in-distribution samples are correctly classified and OOD samples are rejected first. Bottom right: Pareto frontiers showing \gls{LL} for corrupted CIFAR10 (severity 5) vs batch prediction time. Batch size is 256, split over 2 Nvidia P100 GPUs. Annotations show ensemble elements and Dropout samples. Note that a single element ensemble is equivalent to SGD.}
    \label{fig:img_res_main}
    \vspace{-0.2in}
\end{figure}

We train ResNet-50 \citep{he2016deep} using all methods under consideration. This model is composed of an input convolutional block, 16 residual blocks and a linear layer. For \glspl{DUN}, our prior over depth is uniform over the first 13 residual blocks. The last 3 residual blocks and linear layer form the output block, providing the flexibility to make predictions from activations at multiple resolutions. We use $1\times1$ convolutions to adapt the number of channels between earlier blocks and the output block. We use default PyTorch training hyperparameters\footnote{\url{https://github.com/pytorch/examples/blob/master/imagenet/main.py}} for all methods. We set per-dataset LR schedules. We use 5 element (standard) deep ensembles, as suggested by \cite{snoek1029can}, and 10 dropout samples. 
We use two variants of depth-ensembles. The first is composed of five elements corresponding to the five most shallow \gls{DUN} sub-networks. The second depth-ensemble is composed of 13 elements, one for each depth used by \glspl{DUN}.
Similarly, S-ResNets are uncertain over only the first 13 layers.
\Cref{fig:img_res_main} contains results for all experiments described below. Mean values and standard deviations are computed across 5 independent training runs. Full details are given in \cref{app:setup_images}.

\paragraph{Rotated MNIST} Following \cite{snoek1029can}, we train all methods on MNIST and evaluate their predictive distributions on increasingly rotated digits. Although all methods perform well on the original test-set, their accuracy degrades quickly for rotations larger than 30\degree. Here, \glspl{DUN} and S-ResNets differentiate themselves by being the least overconfident. Additionally, depth-ensembles improve over standard ensembles. We hypothesize that predictions based on features at diverse resolutions allow for increased disagreement.

\paragraph{Corrupted CIFAR} Again following \cite{snoek1029can}, we train models on CIFAR10 and evaluate them on data subject to 16 different corruptions with 5 levels of intensity each \citep{hendrycks2019benchmarking}. Here, Ensembles significantly outperform all single network methods in terms of error and LL at all corruption levels, with depth-ensembles being notably better than standard ensembles.
This is true even for the 5 element depth-ensemble, which has relatively shallow networks with far fewer parameters.
The strong performance of depth-ensembles further validates our hypothesis that networks of different depths provide a useful diversity in predictions.
With a single network, \glspl{DUN} perform similarly to SGD and Dropout on the uncorrupted data. However, leveraging a distribution over depth allows \glspl{DUN} to be the most robust non-ensemble method.

\paragraph{\gls{OOD} Rejection}
We simulate a realistic \gls{OOD} rejection scenario \citep{oatml2019bdlb} by jointly evaluating our models on an in-distribution and an \gls{OOD} test set. We allow our methods to reject increasing proportions of the data based on predictive entropy before classifying the rest. All predictions on \gls{OOD} samples are treated as incorrect. Following \cite{nalisnick2018do}, we use CIFAR10 and SVHN as in and out of distribution datasets.
Ensembles perform best. In their standard configuration, \glspl{DUN} show underconfidence. They are incapable of separating very uncertain in-distribution inputs from \gls{OOD} points. We re-run \glspl{DUN} using the exact posterior over depth $p(d | \dset; \btheta)$ in \cref{eq:marginalisation}, instead of $q_{\balpha}(d)$. The exact posterior is computed while setting batch-norm to test mode. See \cref{app:additional_image_results} for additional discussion. This resolves underconfidence, outperforming dropout and coming second, within error, of ensembles. We don't find exact posteriors to improve performance in any other experiments. Hence we abstain from using them, as they require an additional evaluation of the train set.


\paragraph{Compute Time} We compare methods' performance on corrupted CIFAR10 (severity 5) as a function of computational budget. The \gls{LL} obtained by a \gls{DUN} matches that of a $\sim$1.8 element ensemble. 
A single \gls{DUN} forward pass is $\sim$1.02 times slower than a vanilla network's. On average, \gls{DUN}s' computational budget matches that of $\sim$0.47 ensemble elements or $\sim$0.94 dropout samples. These values are smaller than one due to overhead such as ensemble element loading. Thus, making predictions with \glspl{DUN} is 10$\times$ faster than with five element~ensembles.
Note that we include loading times for ensembles to reflect that it is often impractical to store multiple ensemble elements in memory. Without loading times, ensemble timing would match Dropout. For \emph{single-element ensembles} (SGD) we report only the prediction time.


\section{Discussion and Future Work}\label{sec:discussion}

We have re-cast \gls{NN} depth as a random variable, rather than a fixed parameter. This treatment allows us to optimise weights as model hyperparameters, preserving much of the simplicity of non-Bayesian \glspl{NN}. Critically, both the model evidence and predictive posterior for \glspl{DUN} can be evaluated with a single forward pass. Our experiments show that networks of different depths obtain diverse fits. As result, \glspl{DUN} produce well calibrated uncertainty estimates, performing well relative to their computational budget on uncertainty-aware tasks. They scale to modern architectures and large datasets. 



In \glspl{DUN}, network weights have dual roles: fitting the data well and expressing diverse predictive functions at each depth. In future work, we would like to develop optimisation schemes that better ensure both roles are fulfilled, and investigate the relationship between excess model capacity and \gls{DUN} performance. We would also like to investigate the effects of \gls{DUN} depth on uncertainty estimation, allowing for more principled model selection. Additionally, because depth uncertainty is orthogonal to weight uncertainty, both could potentially be combined to expand the space of hypothesis over which we perform inference.
Furthermore, it would be interesting to investigate the application of \glspl{DUN} to a wider range of \gls{NN} architectures, for example stacked RNNs or Transformers.


\section*{Broader Impact}

We have introduced a general method for training neural networks to capture model uncertainty. These models are fairly flexible and can be applied to a large number of applications, including potentially malicious ones. Perhaps, our method could have the largest impact on critical decision making applications, where reliable uncertainty estimates are as important as the predictions themselves. Financial default prediction and medical diagnosis would be examples of these. 

We hope that this work will contribute to increased usage of uncertainty aware deep learning methods in production. \glspl{DUN} are trained with default hyperparameters and easy to make converge to reasonable solutions. The computational cost of inference in \glspl{DUN} is similar to that of vanilla \glspl{NN}. This makes \glspl{DUN} especially well suited for applications with real-time requirements or low computational resources, such as self driving cars or sensor fusion on embedded devices. More generally, \glspl{DUN} make leveraging uncertainty estimates in deep learning more accessible for researchers or practitioners who lack extravagant computational resources.

Despite the above, a hypothetical failure of our method, e.g. providing miscalibrated uncertainty estimates, could have large negative consequences. This is particularly the case for critical decision making applications, such as medical diagnosis.



\begin{ack}
We would like to thank Eric Nalisnick and John Bronskill for helpful discussions.
We also thank Pablo Morales-Álvarez, Stephan Gouws, Ulrich Paquet, Devin Taylor, Shakir Mohamed, Avishkar Bhoopchand and Taliesin Beynon for giving us feedback on this work.
Finally, we thank Marc Deisenroth and Balaji Lakshminarayanan for helping us acquire the flights dataset and Andrew Foong for providing us with the UCI gap datasets.

JA acknowledges support from Microsoft Research, through its PhD Scholarship Programme, and from the EPSRC. JUA acknowledges funding from the EPSRC and the Michael E. Fisher Studentship in Machine Learning. This work has been performed using resources provided by the Cambridge Tier-2 system operated by the University of Cambridge Research Computing Service (http://www.hpc.cam.ac.uk) funded by EPSRC Tier-2 capital grant EP/P020259/1.
\end{ack}

\bibliographystyle{plainnat}
{\small\bibliography{neurips_2020}}

\newpage
\appendix

\section*{Appendix}
This appendix is arranged as follows:
\begin{itemize}
    \item We derive the lower bound used to train \glspl{DUN} in \cref{app:Derivation}.
    \item We analyse the proposed MLE \cref{eq:marginal_likelihood} and VI \cref{eq:var_objective} objectives in \cref{app:vI_vs_MLE}.
    \item We discuss how to compute uncertainty estimates with all methods under consideration in \cref{app:computing_uncertainties}.
    \item We discuss approaches to evaluate the quality of uncertainty estimates in \cref{app:evaluating_uncertainties}.
    \item We detail the experimental setup used for training and evaluation in \cref{app:experimental_setup}.
    \item We provide additional experimental results in \cref{app:experiment_appendix}.
    \item We discuss the application of \glspl{DUN} to neural architecture search in \cref{app:duns_for_nas}.
    \item We show how standard PyTorch \glspl{NN} can be adapted into \glspl{DUN} in \cref{app:dun_impl}.
    \item We provide some negative results in \cref{app:neg_res}.
\end{itemize}

\section{Derivation of \cref{eq:var_objective} and link to the EM algorithm}\label{app:Derivation}

Referring to $\dset{=}\{\bff{X}, \bff{Y}\}$ with $\bff{X} = \{\bff{x}^{(n)}\}_{n=1}^{N} \text{, and } \bff{Y}=\{\bff{y}^{(n)}\}_{n=1}^{N}$, we show that \cref{eq:var_objective} is a lower bound on $\log p(\dset; \btheta) = \log p(\bff{Y} | \bff{X}; \btheta)$:
\begin{align}\label{eq:ELBO_derivation}
    \text{KL}(q_{\balpha}(d)\,\|\,p(d|\dset; \btheta)) &= \EX_{q_{\balpha}(d)}[\log q_{\balpha}(d) - \log p(d|\dset)] \notag \\
    &= \EX_{q_{\balpha}(d)}\left[\log q_{\balpha}(d) -  \log \frac{p(\bff{Y} | \bff{X}, d; \btheta) \,  p(d)}{p(\bff{Y} | \bff{X}; \btheta)}\right]\notag \\
    &= \EX_{q_{\balpha}(d)}[\log q_{\balpha}(d) - \log p(\bff{Y} | \bff{X}, d; \btheta) -\log p(d) + \log p(\bff{Y} | \bff{X}; \btheta)]\notag\\
    &= \EX_{q_{\balpha}(d)}[-\log p(\bff{Y} | \bff{X}, d; \btheta)] + \text{KL}(q_{\balpha}(d)\,\|\,p(d)) + \log p(\bff{Y} | \bff{X}; \btheta)\notag\\
    &= -\mathcal{L}(\balpha, \btheta) + \log p(\bff{Y} | \bff{X}; \btheta).
\end{align}
Using the non-negativity of the KL divergence, we can see that: $\mathcal{L}(\balpha, \btheta) \leq \log p(\bff{Y} | \bff{X}; \btheta)$.

We now discuss the link to the EM algorithm introduced in \cref{sec:inference}. Recall that, in our model, network depth $d$ acts as the latent variable and network weights $\btheta$ are parameters. For a given setting of network weights $\btheta^{k}$, at optimisation step $k$, we can apply Bayes rule to perform the \emph{E step}, obtaining the exact posterior over $d$:
\begin{gather}\label{eq:e_step}
     \alpha^{k+1}_{j} = p(d{=}j|\dset; \btheta^{k}) = \frac{p(d{=}j)\cdot\prod_{n=1}^{N} p(\bff{y}^{(n)}|\bff{x}^{(n)}, d{=}j; \btheta^{k})}{\sum^{D}_{i=0}p(d{=}i)\cdot\prod_{n=1}^{N} p(\bff{y}^{(n)}|\bff{x}^{(n)}, d{=}i; \btheta^{k})}
\end{gather}
The posterior depth probabilities can now be used to marginalise this latent variable and perform maximum likelihood estimation of network parameters. This is the \emph{M step}:
\begin{align}\label{eq:m_step}
     \btheta^{k+1} &=  \argmax_{\btheta} \EX_{p(d|\dset; \btheta^{k})}\left[\prod_{n=1}^{N} p(\bff{y}^{(n)}|\bff{x}^{(n)}, d; \btheta^{k})\right]\notag \\
     &= \argmax_{\btheta} \sum_{i=0}^{D} p(d{=}i|\dset; \btheta^{k}) \prod_{n=1}^{N} p(\bff{y}^{(n)}|\bff{x}^{(n)}, d{=}i; \btheta^{k})
\end{align}
The E step \cref{eq:e_step} requires calculating the likelihood of the complete training dataset. The M step requires optimising the weights of the \gls{NN}. Both operations are expensive when dealing with large networks and big data. The EM algorithm is not practical in this case, as requires performing both steps multiple times. We sidestep this issue through the introduction of an approximate posterior $q(d)$, parametrised by ${\balpha}$, and a variational lower bound on the marginal log-likelihood \cref{eq:ELBO_derivation}. The corresponding variational E step is given by:
\begin{gather}\label{eq:var_E_step}
    \balpha^{k+1} = \argmax_{\balpha} \textstyle{\sum}_{n=1}^{N} \EX_{q_{\balpha}(d)}\left[ \log p(\bff{y}^{(n)}|\bff{x}^{(n)}, d; \btheta^{k})\right] - \text{KL}( q_{\balpha^{(k)}}(d)\,\|\,p_{\bbeta}(d))
\end{gather}
Because our variational family contains the exact posterior distribution -- they are both categorical -- the ELBO is tight at the optima with respect to the variational parameters $\balpha$. Solving \cref{eq:var_E_step} recovers $\balpha$ such that $q_{\balpha^{k+1}}(d)\,{=}\,p(d|\dset; \btheta^{k})$. This step can be performed with stochastic gradient optimisation. 

We can now combine the variational E step \cref{eq:var_E_step} and M step \cref{eq:m_step} updates, recovering \cref{eq:var_objective}, where $\balpha$ and $\btheta$ are updated simultaneously through gradient steps:
\begin{align*}
     \mathcal{L}(\balpha, \btheta) = \textstyle{\sum}_{n=1}^{N} \EX_{q_{\balpha}(d)}\left[ \log p(\bff{y}^{(n)}|\bff{x}^{(n)}, d; \btheta)\right] - \text{KL}( q_{\balpha}(d)\,\|\,p(d))
\end{align*}
This objective is amenable to minibatching. The variational posterior tracks the true posterior during gradient updates. Thus, \cref{eq:var_objective}, allows us to optimise a lower bound on the data's marginal log-likelihood which is unbiased in the limit.

\section{Comparing VI and MLL Training Objectives}\label{app:vI_vs_MLE}

In this section, we further compare the \gls{MLL} \cref{eq:marginal_likelihood} and \gls{VI} \cref{eq:detailed_objective} training objectives presented in \cref{sec:inference}. Our probabilistic model is atypical in that it can have millions of hyperparameters, \gls{NN} weights, while having a single latent variable, depth. For moderate to large datasets, the posterior distribution over depth is determined almost completely by the setting of the network weights. The success of \glspl{DUN} is largely dependent on being able to optimise these hyperparameters well. Even so, our probabilistic model tells us nothing about how to do this. We investigate the gradients of both objectives with respect to the hyperparameters. For \gls{MLL}:
\begin{align}
\frac{\partial}{\partial \btheta} \log p(\dset; \btheta) &=  \frac{\partial}{\partial \btheta} {\mathrm{logsumexp}}_{d}(\log p(\dset| d; \btheta) + \log p(d))\notag\\
&= \sum_{i=0}^{D} \frac{p(\dset| d{=}i; \btheta) p(d{=}i)}{\sum_{j=0}^{D}p(\dset| d{=}j; \btheta) p(d{=}j)}  \frac{\partial}{\partial \btheta} \log p(\dset | d{=}i; \btheta)\notag\\
&= \sum_{i=0}^{D} p(d{=}i | \dset; \btheta) \frac{\partial}{\partial \btheta} \log p(\dset | d{=}i; \btheta)\notag\\
&= \EX_{p(d | \dset; \btheta)}[\frac{\partial}{\partial \btheta} \log p(\dset | d; \btheta)] \label{eq:MLL_gradient}
\end{align}
The gradient of the marginal log-likelihood is equivalent to expectation, under the posterior over depth, of the gradient of the log-likelihood conditioned on depth. The weights of the subnetwork which is able to best explain the data at initialisation will receive larger gradients. This will result in this depth fitting the data even better and receiving larger gradients in successive iterations while the gradients for subnetworks of different depths vanish, i.e. the rich get richer. We conjecture that the \gls{MLL} objective is prone to hard-to-escape local optima, at which a single depth is used. This can be especially problematic if the initial posterior distribution has its maximum over shallow depths, as this will reduce the capacity of the \gls{NN}.

On the other hand, \gls{VI} decouples the likelihood at each depth from the approximate posterior during optimisation: 
\begin{gather}
    \frac{\partial}{\partial \btheta} \mathcal{L}(\btheta, \balpha) = \sum^{D}_{i=0} q_{\balpha}(d{=}i)\frac{\partial}{\partial \btheta} \log p(\dset | d{=}i; \btheta)\notag\\
    \frac{\partial}{\partial \alpha_{i}} \mathcal{L}(\btheta, \balpha) = \underbrace{\log p(\dset | d{=}i; \btheta)\frac{\partial}{\partial \alpha_{i}} q_{\balpha}(d{=}i)}_{\RN{1}} - (\log q_{\balpha}(d{=}i) - \log p(d{=}i) + 1) \frac{\partial}{\partial \alpha_{i}} q_{\balpha}(d{=}i) \label{eq:ELBO_gradient}
\end{gather}
For moderate to large datasets, when updating the variational parameters $\balpha$, the data dependent term ($\RN{1}$) of the ELBO's gradient will dominate. However, the gradients that reach the variational parameters are scaled by the log-likelihood at each depth. In contrast, in \cref{eq:MLL_gradient}, the likelihood at each depth scales the gradients directly. We conjecture that, with \gls{VI}, $\balpha$ will converge slower than the true posterior does when optimising the \gls{MLL} directly. This allows network weights to reach to solutions that explain the data well at multiple depths.

\begin{figure}[h]
\vspace{-0.1cm}
 \centering
    \includegraphics[width=\linewidth]{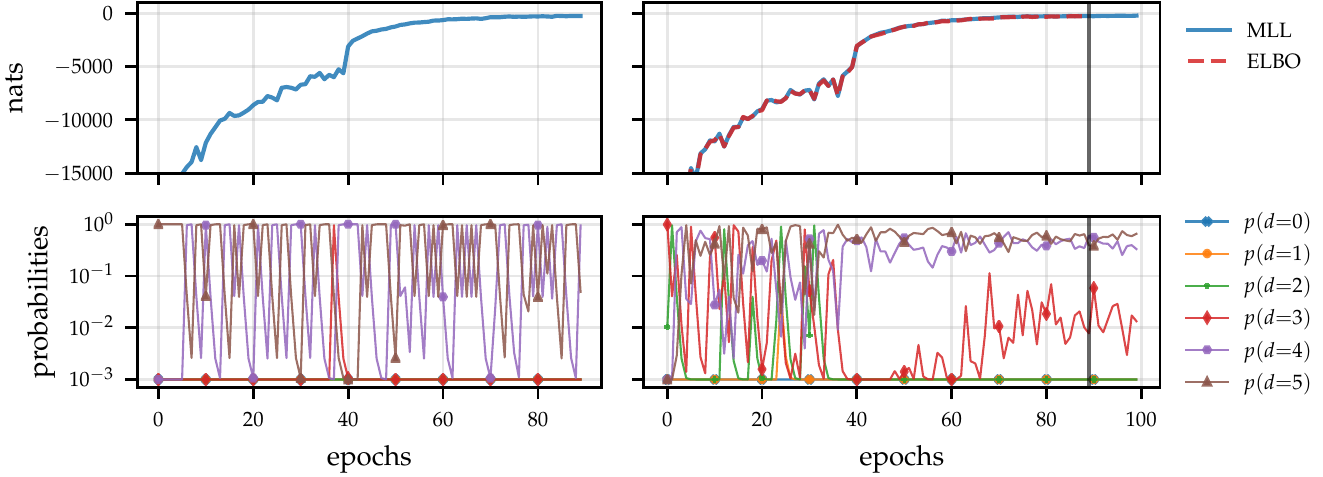}
    \caption{Top row: progression of the \gls{MLL} and \gls{ELBO} during training of ResNet-50 \glspl{DUN} on the Fashion dataset. Bottom: progression of depth posterior probabilities. The left column corresponds to \gls{MLL} optimisation and the right to \gls{VI}. For the latter, approximate posterior probabilities are shown. We perform an additional 10 epochs of ``finetunning'' on the \gls{VI} \gls{DUN} with the \gls{MLL} objective. These are separated by the vertical black line. True posterior probabilities are shown for these 10 epochs. The posterior over depth, ELBO and \gls{MLL} values shown are not stochastic estimates. They are computed using the full training~set.}
    \label{fig:appendix_train_img_resnet}
\vspace{-0.15in}
\end{figure}

We test the above hypothesis by training a \gls{ResNet}-50 \gls{DUN} on the Fashion-MNIST dataset, as shown in \cref{fig:appendix_train_img_resnet}. We treat the first 7 residual blocks of the model as the \glspl{DUN} input block and the last 3 as the output block. This leaves us with the need to infer a distribution over 5 depths (7-12). 
Both the \gls{MLL} and VI training schemes run for 90 epochs, with scheduling described in \cref{app:setup_images}. We then fine-tune the \gls{DUN} that was trained with VI for 10 additional epochs using the \gls{MLL} objective. 
Both training schemes obtain very similar \gls{MLL} values. The dataset under consideration is much larger than the one in \cref{sec:VI_vs_MLE}, but the dimensionality of the latent variable stays the same. Hence, the variational gap is small relative to the \gls{MLL}. Nevertheless, unlike with the \gls{MLL} objective, \gls{VI} training results in posteriors that avoid placing all of their mass on a single depth setting. 

\begin{figure}[h]
\vspace{-0.1cm}
 \centering
    \includegraphics[width=0.7\linewidth]{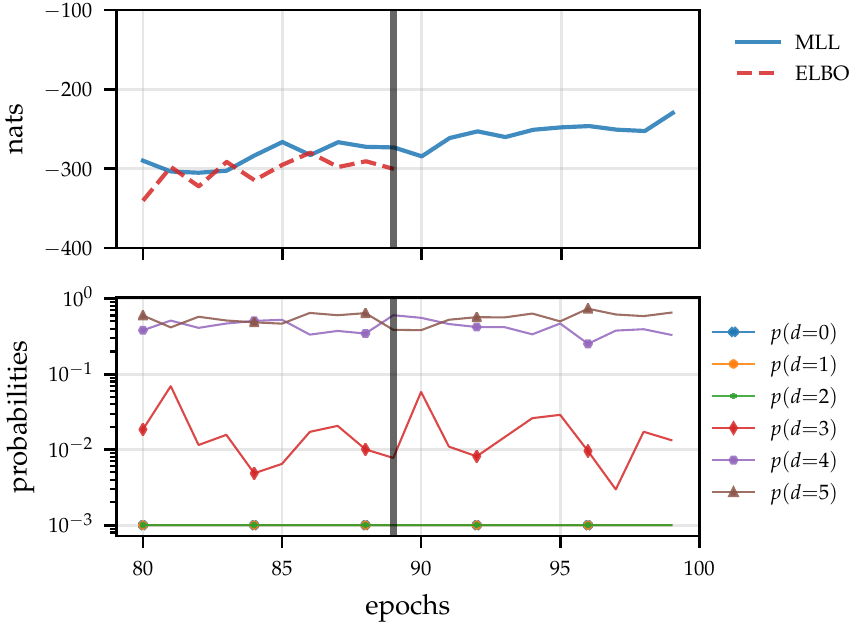}
    \caption{Zoomed-in view of the last 20 epochs of \cref{fig:appendix_train_img_resnet}. The vertical black line denotes the switch from \gls{VI} training to \gls{MLL} optimisation. Probabilities to the left of the line correspond to the variational posterior $q$. The ones to the right of the line correspond to the exact posterior. In some steps of training, the ELBO appears to be larger than the \gls{MLL} due to numerical error.}
    \label{fig:appendix_train_img_resnet_zoom}
\vspace{-0.1in}
\end{figure}

Zooming in on the last 20 epochs in \cref{fig:appendix_train_img_resnet_zoom}, we see that after converging to a favorable solution with \gls{VI}, optimising the \gls{MLL} objective directly does not result in the posterior collapsing to a single depth. Instead, it remains largely the same as the \gls{VI} surrogate posterior. \gls{VI} optimisation allowed us to find an optima of the \gls{MLL} where multiple depths explain the data similarly well.   

\begin{figure}[h]
\vspace{-0.1cm}
 \centering
    \includegraphics[width=0.85\linewidth]{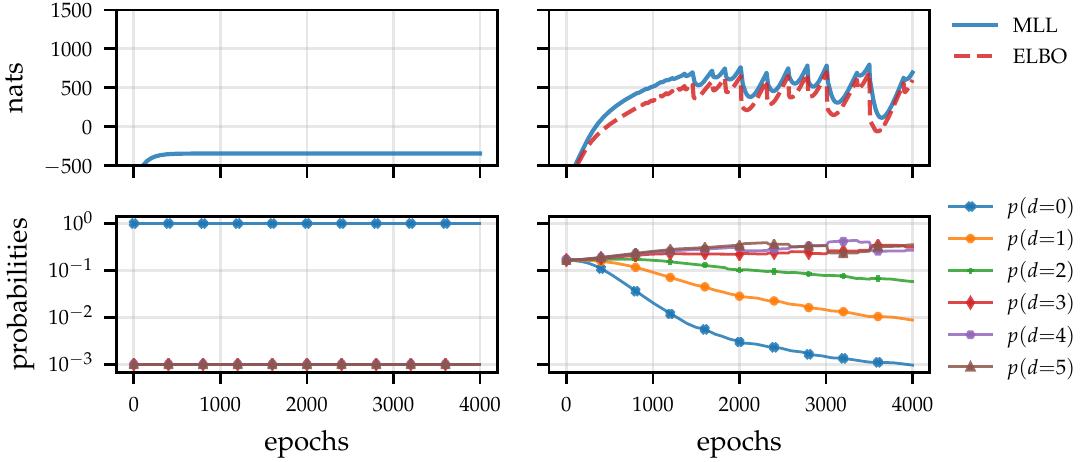}
    \caption{Top row: progression of \gls{MLL} and \gls{ELBO} during training of \glspl{DUN} on the Boston dataset. Bottom: progression of depth posterior probabilities. The left column corresponds to \gls{MLL} optimisation and the right to \gls{VI}. For the latter, approximate posterior probabilities are shown.}
    \label{fig:appendix_boston_train}
\vspace{-0.1in}
\end{figure}

\begin{figure}[h]
\vspace{-0.1cm}
 \centering
    \includegraphics[width=0.85\linewidth]{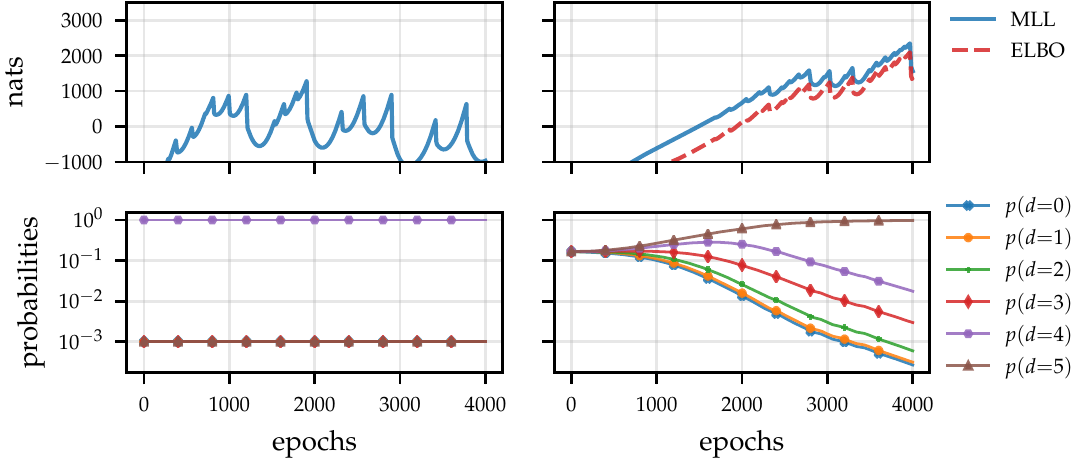}
    \caption{Top row: progression of \gls{MLL} and \gls{ELBO} during training of \glspl{DUN} on the Wine dataset. Bottom: progression of depth posterior probabilities. The left column corresponds to \gls{MLL} optimisation and the right to \gls{VI}. For the latter, approximate posterior probabilities are shown.}
    \label{fig:appendix_wine_train}
\vspace{-0.1in}
\end{figure}

In \cref{fig:appendix_boston_train} and \cref{fig:appendix_wine_train} we show the \gls{MLL}, ELBO and posterior probabilities obtained with our two optimisation objectives, \cref{eq:marginal_likelihood} and \cref{eq:detailed_objective}, on the Boston and Wine datasets respectively. Like in \cref{sec:VI_vs_MLE}, we employ 5 hidden layer \glspl{DUN} without residual connections and 100 hidden units per layer. The input and output blocks consist of linear layers. Both approaches employ full-batch gradient descent with a step-size of $10^{-3}$ and momentum of $0.9$. 

The \gls{MLL} objective consistently reaches parameter settings for which the posterior over depth places all its mass on a single depth. We found the depth to which the posterior collapses to change depending on weight initialisation. However, converging to a network where no hidden layers were used $p(d{=}0){=}1$ seems to be the most common occurrence. Even when the chosen depth is large, as in the Wine dataset example, we are able to reach significantly larger likelihood values when optimising the ELBO. Even though the variational gap becomes very small by the end of training, the approximate posterior probabilities found with \gls{VI} place non-0 mass over more than one depth; training with \gls{VI} allows us to find weight configurations which explain the data well while being able to use multiple layers for prediction.

\section{Computing Uncertainties}\label{app:computing_uncertainties}

In this work, we consider \glspl{NN} which parametrise two types of distributions over target variables: the categorical for classification problems and the Gaussian for regression. For generality, in this section we omit references to model hyperparameters $\btheta$ and refer to the distribution over random variables that induces stochasticity in our networks as $q(\bff{w})$. In \glspl{DUN}, this is a distribution over depth. It is a distribution over weights in the case of MFVI, MC Dropout and ensembles. 

For classification models, our networks output a probability vector with elements $f_{k}(\bff{x}, \bff{w})$, corresponding to classes $\{c_{k}\}_{k=1}^{K}$. The likelihood function is $p(y | \bff{x}, \bff{w}) = \mathrm{Cat}(y; f(\bff{x}, \bff{w}))$. Through marginalisation, the uncertainty in $\bff{w}$ is translated into uncertainty in predictions. For \glspl{DUN}, computing the exact predictive posterior is tractable \cref{eq:marginalisation}. However, for our baseline approaches, we resort to approximating it with $M$ \gls{MC} samples:
\begin{align*}
    p(\bff{y}^{*} | \bff{x}^{*}, \dset) &= \EX_{p(\bff{w} | \dset)}[p(\bff{y}^{*} | \bff{x}^{*}, \bff{w})] \\
    &\approx \frac{1}{M}\sum^{M}_{m=0} f(\bff{x}^{*}, \bff{w}); \quad \bff{w}\sim q(\bff{w})
\end{align*}
In both, the exact and approximate cases, the resulting predictive distribution is categorical. We quantify its uncertainty using entropy:
\begin{gather*}
    H(\bff{y}^{*} | \bff{x}^{*}, \dset) = \sum^{K}_{k=1}  p(y^{*}{=}c_{k} | \bff{x}^{*}, \dset) \log p(y^{*}{=}c_{k} | \bff{x}^{*}, \dset)
\end{gather*}
For regression, we employ homoscedastic likelihood functions. The mean is parametrised by a \gls{NN} and the variance is learnt as a standalone parameter: $p(\bff{y}^{*} | \bff{x}^{*}, \bff{w}) = \mathcal{N}(\bff{y}; f(\bff{x}^{*}, \bff{w}), \bsigma^{2}\cdot I)$. For the models under consideration, marginalising over $\bff{w}$ induces a Gaussian mixture distribution over outputs. We approximate this mixture with a single Gaussian using moment matching: $p(\bff{y}^{*} | \bff{x}^{*}) \approx \mathcal{N}(\bff{y}; \bmu_{a}, \bsigma_{a}^{2})$. For \glspl{DUN}, the mean can be computed exactly:
\begin{gather*}
    \bmu_{a} = \sum_{i=0}^{D} f(\bff{x}^{*}, \bff{w}{=}i) q(\bff{w}{=}i)
\end{gather*}
Otherwise, we estimate it with \gls{MC}:
\begin{gather*}
    \bmu_{a} \approx \frac{1}{M}\sum^{M}_{m=0} f(\bff{x}^{*}, \bff{w}); \quad \bff{w}\sim q(\bff{w})
\end{gather*}
The predictive variance is obtained as the variance of the GMM.  For \glspl{DUN}:
\begin{gather*}
    \bsigma_{a}^{2} = \underbrace{\sum^{D}_{i=0} q(\bff{w}{=}i) f(\bff{x}^{*}, \bff{w}{=}i)^{2} - \bmu_{a}^{2}}_{\RN{1}} + \underbrace{\bsigma^{2}}_{\RN{2}}; 
\end{gather*}
Otherwise, we estimate it with \gls{MC}:
\begin{gather*}
    \bsigma_{a}^{2} \approx \underbrace{\frac{1}{M}\sum^{M}_{m=1} f(\bff{x}^{*}, \bff{w})^{2} - \bmu_{a}^{2}}_{\RN{1}} + \underbrace{\bsigma^{2}}_{\RN{2}};  \quad \bff{w}\sim q(\bff{w})
\end{gather*}
Here, $\RN{1}$ reflects model uncertainty -- our lack of knowledge about $\bff{w}$ -- while $\RN{2}$ tells us about the irreducible uncertainty or noise in our training data.

\section{Evaluating Uncertainty Estimates}\label{app:evaluating_uncertainties}

We consider the following approaches to quantify the quality of uncertainty estimates:
\begin{itemize}
    \item \textbf{Test Log Likelihood} (higher is better): This metric tells us about how probable it is that the test targets where generated using the test inputs and our model. It is a proper scoring rule \citep{gneiting2007strictly} that depends on both the accuracy of predictions and their uncertainty. We employ it in both classification and regression settings, using categorical and Gaussian likelihoods, respectively.
    \item \textbf{Brier Score} (lower is better): Proper scoring rule that measures the accuracy of predictive probabilities in classification tasks. It is computed as the mean squared distance between predicted class probabilities and one-hot class labels:
    \begin{gather*}
        \mathrm{BS} = \frac{1}{N} \sum^{N}_{n=1} \frac{1}{K}\sum^{K}_{k=1} (p(y^{*}=c_{k} | \bff{x}^{*}, \dset) - \ind[y^{*}=c_{k}])^{2}
    \end{gather*}
    Erroneous predictions made with high confidence are penalised less by Brier score than by log-likelihood. This can avoid outlier inputs from having a dominant effect on experimental results. Nevertheless, we find Brier score to be less sensitive than log-likelihood, making it harder to distinguish the approaches being compared. Hence, we favor the use of log-likelihood in \cref{sec:imgres}.
    \item \textbf{\gls{ECE}} (lower is better): This metric measures the difference between predictive confidence and empirical accuracy in classification. It is computed by dividing the [0,1] range into a set of bins $\{B_{s}\}_{s=1}^{S}$ and weighing the miscalibration in each bin by the number of points that fall into it $\abs{B_{s}}$:
    \begin{gather*}
        \mathrm{ECE} = \sum^{S}_{s=1} \frac{\abs{B_{s}}}{N} \abs{\mathrm{acc}(B_{s}) - \mathrm{conf}(B_{s}))}
    \end{gather*}
    Here, 
    \begin{gather*}
        \mathrm{acc}(B_{s})\,{=}\,\frac{1}{\abs{B_{s}}} \sum_{\bff{x} \in B_{s}} \ind[\bff{y}\,{=}\,\argmax_{c_{k}}p(\bff{y}| \bff{x}, \dset)]\quad \textrm{and}\\
        \mathrm{conf}(B_{s})\,{=}\, \frac{1}{\abs{B_{s}}} \sum_{\bff{x} \in B_{s}} \max p(\bff{y}| \bff{x}, \dset).
    \end{gather*}
    \gls{ECE} is not a proper scoring rule. A perfect \gls{ECE} score can be obtained by predicting the marginal distribution of class labels $p(\bff{y})$ for every input. A well calibrated predictor with poor accuracy would obtain low log likelihood values but also low \gls{ECE}.
    Although \gls{ECE} works well for binary classification, the naive adaption to the multi-class setting suffers from a number of pathologies \citep{nixon2019measuring}. Thus, we do not employ this metric.
    \item \textbf{Regression  Calibration Error (RCE)} (lower is better): We extend \gls{ECE} to regression settings, while avoiding the pathologies described by \cite{nixon2019measuring}: We seek to asses how well our model's predictive distribution describes the residuals obtained on the test set. It is not straight forward to define bins, like in standard \gls{ECE}, because our predictive distribution might not have finite support. We apply the \gls{CDF} of our predictive distribution to our test targets. If the predictive distribution describes the targets well, the transformed distribution should resemble a uniform with support $[0, 1]$. This procedure is common for backtesting market risk models \citep{market_risk_book}.
    
    To asses the global similarity between our targets' distribution and our predictive distribution, we separate the $[0, 1]$ interval into $S$ equal-sized bins $\{B_{s}\}_{s=1}^{S}$. We compute calibration error in each bin as the difference between the proportion of points that have fallen within that bin and $\nicefrac{1}{S}$:
    \begin{gather*}
        \mathrm{RCE} = \sum^{S}_{s=1} \frac{\abs{B_{s}}}{N}\cdot \abs{\frac{1}{S} - \frac{\abs{B_{s}}}{N}};\quad \abs{B_{s}} = \sum_{n=1}^{N} \ind[ CDF_{p(y | \bff{x}^{(n)})}(y^{(n)}) \in B_{s}]
    \end{gather*}
    Alternatively, we can asses how well our model predicts extreme values with a ``frequency of tail losses'' approach \citep{tail_frequency_paper}. It might not be realistic to assume the noise in our targets is Gaussian. Only considering calibration at the tails of the predictive distribution allows us to ignore shape mismatch between the predictive distribution and the true distribution over targets. Instead, we focus on our model's capacity to predict on which inputs it is likely to make large mistakes. This can be used to ensure our model is not overconfident \gls{OOD}. We specify two bins $\{B_{0}, B_{1}\}$, one at each tail end of our predictive distribution, and compute \textbf{Tail Calibration Error (TCE)} as:
    \begin{gather*}
        \mathrm{TCE} = \sum_{s=0}^{1} \frac{\abs{B_{s}}}{\abs{B_{0}} + \abs{B_{1}}} \cdot \abs{\frac{1}{\tau} - \frac{\abs{B_{s}}}{N}}; \\ \abs{B_{0}} = \sum_{n=1}^{N} \ind[ CDF_{p(y | \bff{x}^{(n)})}(y^{(n)}) < \tau]; \quad
        \abs{B_{1}} = \sum_{n=1}^{N} \ind[ CDF_{p(y | \bff{x}^{(n)})}(y^{(n)}) \geq (1-\tau)]
    \end{gather*}
    We specify the tail range of our distribution by selecting $\tau$. Note that this is slightly different from \citet{tail_frequency_paper}, who uses a binomial test to asses whether a model's predictive distribution agrees with the distribution over targets in the tails.
    
    RCE and \gls{TCE} are not a proper scoring rules. Additionally, they are only applicable to 1 dimensional continuous target variables.
\end{itemize}
Please see \citep{ashukha2020pitfalls,snoek1029can} for additional discussion on evaluating uncertainty estimates of predictive models. 

\section{Experimental Setup}\label{app:experimental_setup}

We implement all of our experiments in PyTorch \citep{pytorch}. Gaussian processes for toy data experiments are implemented with GPyTorch \citep{gpytorch}.

\subsection{Toy Dataset Experiments}\label{app:experimental_setup_toy}

All \glspl{NN} used for toy regression experiments in \cref{sec:toy_datasets} consist of fully connected models with ReLU activations and residual connections. Their hidden layer width is 100. Batch normalisation is applied after every layer for SGD and \glspl{DUN}. Unless specified otherwise, the same is true for the additional toy dataset experiments conducted in \cref{app:toy_datasets}. Network depths are defined on a per-experiment basis. \glspl{DUN} employ linear input and output blocks, meaning that a depth of $d{=}0$ corresponds to a linear model. We refer to depth as the number of hidden layers of a \gls{NN}.

 Ensemble elements, \glspl{DUN} and dropout models employ a weight decay value of $10^{-4}$. Ensembles are composed of 20 identical networks, trained from different initialisations. Initialisation parameters are sampled from the He initialisation \citep{he2015delving}. Dropout probabilities are fixed to $0.1$. MFVI networks use a $\mathcal{N}(\bff{0},I)$ prior. Gradients of the likelihood term in the ELBO are estimated with the local reparameterisation trick \citep{kingma2015variational} using 5 \gls{MC} samples. \glspl{DUN} employ uniform priors, assigning the same mass to each depth. 
 
 Networks are optimised using 6000 steps of full-batch gradient descent with a momentum value of $0.9$ and learning rate of $10^{-3}$. Exceptions to this are: Dropout being trained for 10000 epochs, as we found 6000 to not be enough to achieve convergence, and MFVI using a learning rate of $10^{-2}$. For MFVI and \glspl{DUN}, we scale the ELBO by one over the number of data points $N$. This makes the scale of the objective insensitive to dataset size.
 
The parameters of the predictive distributions are computed as described in \cref{app:computing_uncertainties}. For 1D datasets, we draw $10^{4}$ \gls{MC} samples with MFVI and dropout. For 2D datasets, we draw $10^{3}$. Plot error bars correspond to the standard deviations of each approach's mean predictions. Thus, they convey model uncertainty. 
 
Gaussian processes use a Gaussian likelihood function and radial basis function (RBF) kernel. For 2d toy experiments, the automatic relevance detection (ARD) version of the kernel is used, allowing for a different length-scale per dimension. A gamma prior with parameters $\alpha=1, \beta=20$ is placed on the length-scale parameter for the 1d datasets. This avoids local optima of the log-likelihood function where fast varying patterns in the data are treated like noise. Noise variance and kernel parameters are learnt by optimising the \gls{MLL} with 100 steps of Adam. Step size is set to $0.1$.

We employ 7 different toy datasets. These allow us to test methods' capacity to express uncertainty in-between clusters of data and outside the convex hull of the data. They also allow us to evaluate methods' capacity to fit differently quickly varying functions. All of them can be loaded using our provided code. 

\subsection{Regression Experiments} \label{app:setup_regression}

\subsubsection{Hyperparameter Optimisation and Training}

To obtain our results on tabular regression tasks, given in \cref{sec:tabresres}, we follow \cite{hernandez2015probabilistic} and follow-up work \citep{gal2016dropout, lakshminarayanan2017simple} in performing \gls{HPO} to determine the best configurations for each method. 
However, rather than using \gls{BO} \citep{snoek2012practical} we use \gls{BOHB} \citep{falkner2018bohb}. This method, as the name suggests, combines \gls{BO} with Hyperband, a bandit based \gls{HPO} method \citep{li2018hyperband}. \gls{BOHB} has the strengths of both \gls{BO} (strong final performance) and Hyperband (scalability, flexibility, and robustness).

In particular, we use the \texttt{HpBandSter} implementation of \gls{BOHB}: \url{https://github.com/automl/HpBandSter}. We run \gls{BOHB} for each dataset and split for 20 iterations using the same settings, shown in \cref{tab:bohb_settings}. \texttt{min\_budget} and \texttt{max\_budget} are defined on a per dataset basis, as shown in \cref{tab:hpo_per_dataset_config}. We find these values to be sufficiently large to ensure all methods' convergence. 

\begin{table}[htbp]
    \centering
    \caption{\gls{BOHB} settings.}
    \begin{tabular}{l|c}
        \toprule
        \textsc{Setting} & \textsc{Value}  \\  \midrule
        \texttt{eta} & 3 \\
        \texttt{min\_points\_in\_model} & \texttt{None} \\
        \texttt{top\_n\_percent} & 14 \\
        \texttt{num\_samples} & 64 \\
        \texttt{random\_fraction} & 1/3 \\
        \texttt{bandwidth\_factor} & 3 \\
        \texttt{min\_bandwidth} & 1e-3 \\
        \bottomrule
    \end{tabular}
    \label{tab:bohb_settings}
\end{table}

\begin{table}[htbp]
    \centering
    \caption{Per-dataset \gls{HPO} configurations.}
    \begin{tabular}{l|cccc}
        \toprule
        \textsc{Dataset} & \textsc{Min Budget} & \textsc{Max Budget} & \textsc{Early Stop Patience} & \textsc{Val Prop} \\ \midrule
        Boston & 200 & 2000 & 200 & 0.15 \\
        Concrete & 200 & 2000 & 200 & 0.15 \\
        Energy & 200 & 2000 & 200 & 0.15 \\
        Kin8nm & 50 & 500 & 50 & 0.15 \\
        Naval & 50 & 500 & 50 & 0.15 \\
        Power & 50 & 500 & 50 & 0.15 \\
        Protein & 50 & 500 & 50 & 0.15 \\
        Wine & 100 & 1000 & 100 & 0.15 \\
        Yacht & 200 & 2000 & 200 & 0.15 \\
        Boston Gap & 200 & 2000 & 200 & 0.15 \\
        Concrete Gap & 200 & 2000 & 200 & 0.15 \\
        Energy Gap & 200 & 2000 & 200 & 0.15 \\
        Kin8nm Gap & 50 & 500 & 50 & 0.15 \\
        Naval Gap & 50 & 500 & 50 & 0.15 \\
        Power Gap & 50 & 500 & 50 & 0.15 \\
        Protein Gap & 50 & 500 & 50 & 0.15  \\
        Wine Gap & 100 & 1000 & 100 & 0.15 \\
        Yacht Gap & 200 & 2000 & 200 & 0.15 \\
        Flights & 2 & 25 & 5 & 0.05 \\ 
        \bottomrule
    \end{tabular}
    \label{tab:hpo_per_dataset_config}
\end{table}

For each test-train split of each dataset, we split the original training set into a new training set and a validation set. The validation sets are taken to be the last $N$ elements of the original training set, where $N$ is calculated from the validation proportions listed in \cref{tab:hpo_per_dataset_config}. 
The training and validation sets are normalised by subtracting the mean and dividing by the variance of the new training set.
\gls{BOHB} performs minimisation on the validation \gls{NLL}. During optimisation, we perform early stopping with patience values show in \cref{tab:hpo_per_dataset_config}.

As shown in \cref{tab:method_hyperparams}, each method has a different set of hyperparameters to optimise. The \gls{BOHB} configuration for each hyperparameter is shown in \cref{tab:hyperparam_configs}. It is worth noting that maximum network depth is a hyperparater which we optimise with \gls{BOHB}. \glspl{DUN} benefit from being deeper as it allows then to perform \gls{BMA} over a larger set of functions. We prevent this from disadvantaging competing methods by choosing the depth at which each one performs best.

All methods are applied to fully-connected networks with hidden layer width of 100. We employ residual connections, allowing all approaches to better take advantage of depth. All methods are trained using SGD with momentum and a batch size of 128. No learning rate scheduling is performed. We use batch-normalisation for \glspl{DUN} and vanilla networks (labelled SGD in experiments). All \glspl{DUN} are trained using \gls{VI} \cref{eq:detailed_objective}. The likelihood term in the MFVI ELBO is estimated with 3 \gls{MC} samples per input. For \gls{MFVI} and Dropout, 10 \gls{MC} samples are used to estimate the test log-likelihood.
Ensembles use 5 elements for prediction. Ensemble elements differ from each other in their initialisation, which is sampled from the He initialisation distribution \citep{he2015delving}. We do not use adversarial training as, inline with \citet{ashukha2020pitfalls}, we do not find it to improve results.

\begin{table}[htbp]
    \centering
    \caption{Hyperparameters optimised for each method.}
    \begin{tabular}{l|cccc}
        \toprule
        \textsc{Hyperparameter} & \textsc{\gls{DUN}} & \textsc{SGD} & \textsc{MFVI} & \textsc{MC Dropout} \\ \midrule
        Learning Rate & \checkmark & \checkmark & \checkmark & \checkmark \\
        SGD Momentum & \checkmark & \checkmark & \checkmark &  \checkmark\\
        Num. Layers & \checkmark & \checkmark & \checkmark & \checkmark \\
        Weight Decay & \checkmark &\checkmark & & \checkmark\\
        Prior Std. Dev. & & & \checkmark & \\
        Drop Prob. & & & & \checkmark \\
        \bottomrule
    \end{tabular}
    \label{tab:method_hyperparams}
\end{table}

\begin{table}[htbp]
    \centering
    \caption{\gls{BOHB} hyperparameter optimisation configurations. All hyperparameters were sampled from uniform distributions.}
    \begin{tabular}{l|ccccc}
        \toprule
        \textsc{Hyperparameter} & \textsc{Lower} & \textsc{Upper} & \textsc{Default} & \textsc{Log} & \textsc{Data Type} \\ \midrule
        Learning Rate & $1 \times 10^{-4}$ & $1$ & $0.01$ & \texttt{True} & \texttt{float} \\
        SGD Momentum & $0$ & $0.99$ & $0.5$ & \texttt{False} & \texttt{float} \\
        Num. Layers & $1$ & $40$ & $5$ & \texttt{False} & \texttt{int} \\
        Weight Decay & $1 \times 10^{-6}$ & $0.1$ & $5 \times 10^{-4}$ & \texttt{True} & \texttt{float} \\
        Prior Std. Dev. & $0.01$ & $10$ & $1$ & \texttt{True} & \texttt{float} \\
        Drop Prob. & $5 \times 10^{-3}$ & $0.5$ & $0.2$ & \texttt{True} & \texttt{float} \\
        \bottomrule
    \end{tabular}
    \label{tab:hyperparam_configs}
\end{table}

\subsubsection{Evaluation}

The best configuration found for each dataset, method and split is used to re-train a model on the entire original training set. For the flights dataset, which does not come with multiple splits, we repeat this five times. We report mean and standard deviation values across all five.
Final run training and test sets are normalised using the mean and variance of the original training set.
Note, however, that the results presented in \cref{sec:tabresres} are unnormalised.
The number of epochs used for final training runs is the number of epochs at which the optimal configuration was found with \gls{HPO}.

Timing experiments for regression models are performed on a 40 core Intel Xeon CPU E5-2650 v3 \@ 2.30GHz. We report computation time for a single batch of size 512, which we evaluate across 5 runs. Ensembles, Dropout and MFVI require multiple forward passes per batch. We report the time taken for all passes to be made. For Ensembles, we also include network loading time.

\subsection{Image Experiments}\label{app:setup_images}
\subsubsection{Training}

The results shown in \cref{sec:imgres} are obtained by training \gls{ResNet}-50 models using \gls{SGD} with momentum. The initial learning rate, momentum, and weight decay are 0.1, 0.9, and $1 \times 10^{-4}$, respectively. We train on 2 Nvidia P100 GPUs with a batch size of 256 for all experiments. Each dataset is trained for a different number of epochs, shown in \cref{tab:img_exp_train_configs}. We decay the learning rate by a factor of 10 at scheduled epochs, also shown in \cref{tab:img_exp_train_configs}. Otherwise, all methods and datasets share hyperparameters. These hyperparameter settings are the defaults provided by \texttt{PyTorch} for training on ImageNet. We found them to perform well across the board. 
We report results obtained at the final training epoch. We do not use a separate validation set to determine the best epoch as we found ResNet-50 to not overfit with the chosen schedules.

\begin{table}[htbp]
    \centering
    \caption{Per-dataset training configuration for image experiments.}
    \begin{tabular}{l|cc}
        \toprule
        \textsc{Dataset} & \textsc{No. Epochs} & \textsc{LR Schedule}  \\ \midrule
        MNIST & 90 & 40, 70 \\
        Fashion & 90 & 40, 70 \\
        SVHN & 90 & 50, 70 \\
        CIFAR10 & 300 & 150, 225 \\
        CIFAR100 & 300 & 150, 225 \\
        \bottomrule
    \end{tabular}
    \label{tab:img_exp_train_configs}
\end{table}

For dropout experiments, we add dropout to the standard \gls{ResNet}-50 model \citep{he2016deep} in between the \nth{2} and \nth{3} convolutions in the bottleneck blocks. This approach follows \citet{zagoruyko2016wide} and \citet{ashukha2020pitfalls} who add dropout in-between the two convolutions of a WideResNet-50's basic block. 
Following their approach, we try a dropout probability of 0.3. However, we find that this value is too large and causes underfitting. A dropout probability of 0.1 provides stronger results. We show results with both settings in \cref{app:additional_image_results}. We use 10 \gls{MC} samples for predictions.
Ensembles use 5 elements for prediction. Ensemble elements differ from each other in their initialisation, which is sampled from the He initialisation distribution \citep{he2015delving}. We do not use adversarial training as, inline with \citet{ashukha2020pitfalls}, we do not find it to improve results.

\begin{figure}[htb]
    \centering
    \begin{tikzpicture}[every node/.style={scale=0.85}]
    	\node[block=depth1] (RB0) {$f_{0}$};
    	\node[block=depth2, below right=1ex and 4ex of RB0] (RB1) {$f_{1}$};
    	\node[block=depth3, below right=1ex and 4ex of RB1] (RB2) {$f_{2}$};
    	\node[block=depth4, below right=1ex and 4ex of RB2] (RB3) {$f_{3}$};
    	\node[block=depth5, below right=1ex and 4ex of RB3] (RB4) {$f_{4}$};
    	\node[block=teal, below right=3ex and 10ex of RB4, text width=2.5ex] (RBD) {$f_{D}$};

    	\node[left=4ex of RB0] (X) {$\mathbf{x}$};
    	
    	\draw[rounded corners, nicecolor=gray!50!white] ($(RB0.north west) + (14.2ex, 0)$)  rectangle ($(RB1.south east)  + (7.ex, 0)$) node[pos=.5] (U1) {$a_1$};
        \draw[rounded corners, nicecolor=gray!70!white] ($(RB0.north west) + (28.4ex, 0)$)  rectangle ($(RB3.south east)  + (7.ex, 0)$) node[pos=.5] (U3) {$a_2$};
        \draw[rounded corners, nicecolor=gray!90!white] ($(RB0.north west) + (41.5ex, 0)$)  rectangle ($(RB4.south east)  + (13.5ex, -1.5ex)$) node[pos=.5] (U4) {$a_K$};

        \draw[arrow, snake] (RBD) -- ([xshift=4ex] RBD.east) node[coordinate] (end) {};
        \draw[arrow, snake] (RB0) -- node[above right=2ex and 2ex] (C1) {$l_0$} (RB0 -| U1.west);
        \draw[arrow, snake] (RB0 -| U1.east) -- node[above right=2ex and 2ex] (C3) {$l_1$} (RB0 -| U3.west);
        \draw[snake, line] (RB0 -| U3.east) -- ($(RB0 -| U3.east) + (1.5ex,0)$);
        \draw[dashed, line] ($(RB0 -| U3.east) + (1.5ex,0)$) -- ($(RB0 -| U3.east) + (6ex,0)$);
        \draw[arrow, snake] ($(RB0 -| U3.east) + (6ex,0)$) -- node[above=2ex] (CD) {$l_{K-1}$} (RB0 -| U4.west);
        \draw[arrow, snake] (RB0 -| U4.east) -- node[above=2ex] (CD1) {$l_{K}$} (RB0 -| end);
        
        \draw[line, gray!30] (C1) -- (C1 |- RBD.south);
        \draw[line, gray!30] (C3) -- (C3 |- RBD.south);
        \draw[line, gray!30] (CD) -- (CD |- RBD.south);
        \draw[line, gray!30] (CD1) -- (CD1 |- RBD.south);
        
        \draw[arrow, snake] (RB1) -- (RB1 -| U1.west);
        \draw[arrow, snake] (RB1 -| U1.east) -- (RB1 -| U3.west);
        \draw[snake, line] (RB1 -| U3.east) -- ($(RB1 -| U3.east) + (1.5ex,0)$);
        \draw[dashed, line] ($(RB1 -| U3.east) + (1.5ex,0)$) -- ($(RB1 -| U3.east) + (6ex,0)$);
        \draw[arrow, snake] ($(RB1 -| U3.east) + (6ex,0)$) -- (RB1 -| U4.west);
        \draw[arrow, snake] (RB1 -| U4.east) -- (RB1 -| end);
        
        \draw[arrow, snake] (RB2) -- (RB2 -| U3.west);
        \draw[snake, line] (RB2 -| U3.east) -- ($(RB2 -| U3.east) + (1.5ex,0)$);
        \draw[dashed, line] ($(RB2 -| U3.east) + (1.5ex,0)$) -- ($(RB2 -| U3.east) + (6ex,0)$);
        \draw[arrow, snake] ($(RB2 -| U3.east) + (6ex,0)$) -- (RB2 -| U4.west);
        \draw[arrow, snake] (RB2 -| U4.east) -- (RB2 -| end);
        
        \draw[arrow, snake] (RB3) -- (RB3 -| U3.west);
        \draw[snake, line] (RB3 -| U3.east) -- ($(RB3 -| U3.east) + (1.5ex,0)$);
        \draw[dashed, line] ($(RB3 -| U3.east) + (1.5ex,0)$) -- ($(RB3 -| U3.east) + (6ex,0)$);
        \draw[arrow, snake] ($(RB3 -| U3.east) + (6ex,0)$) -- (RB3 -| U4.west);
        \draw[arrow, snake] (RB3 -| U4.east) -- (RB3 -| end);
        
        \draw[snake, line] (RB4 -| U3.east) -- ($(RB4 -| U3.east) + (1.5ex,0)$);
        \draw[dashed, line] ($(RB4 -| U3.east) + (1.5ex,0)$) -- ($(RB4 -| U3.east) + (6ex,0)$);
        \draw[arrow, snake] ($(RB4 -| U3.east) + (6ex,0)$) -- (RB4 -| U4.west);
        \draw[arrow, snake] (RB4 -| U4.east) -- (RB4 -| end);

        \draw [decorate,decoration={brace,amplitude=1ex, raise=.5ex},xshift=2ex, thick] (RB0 -| end.west) -- (end) node [block=blue!80!white, midway, xshift=6ex, text width=5ex, thin] (OB) {$f_{D+1}$};
        
        \draw[arrow] (X) -- (RB0);
    	\draw[arrow] (RB0) |- (RB1);
    	\draw[arrow] (RB1) |- (RB2);
    	\draw[arrow] (RB2) |- (RB3);
    	\draw[arrow] (RB3) |- (RB4);
    	\draw[line] (RB4) -- ([yshift=-1.5ex] RB4.south);
    	\draw[line, dashed] ([yshift=-1.5ex] RB4.south) |- ([xshift=-4.ex] RBD.west);
    	\draw[arrow] ([xshift=-4.ex] RBD.west) -- (RBD);
        
    	\node[right=4ex of OB] (YD) {$\mathbf{\hat y}_{i}$};
        	
    	\draw[arrow] (OB.east) -- (YD);

    \end{tikzpicture}
    \caption{For network architectures in which the input and output number of channels or dimensions is not constant, we add \emph{adaption layers} to the computational model shown in \cref{fig:nn_structure_graph_model}. The $n^{\text{th}}$ adaption layer $a_n$ takes a number of channels/dimensions $l_{n-1}$ and outputs $l_{n}$ channels/dimensions. Later adaption layers are reused multiple times, reducing the number of parameters required. Note that block sizes are unrelated to their number of parameters.}
    \label{fig:upscaling_model}
\end{figure}
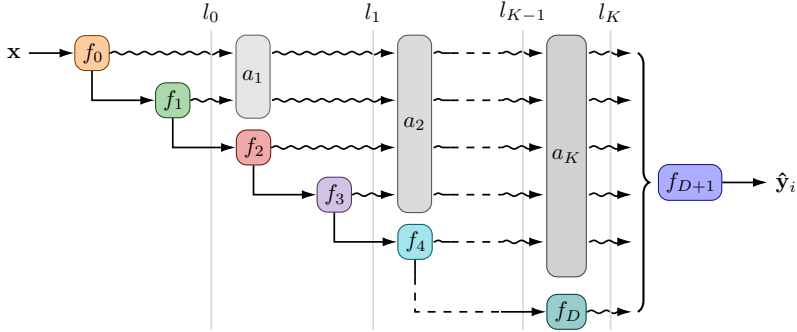

We modify the standard ResNet-50 architecture such that the first $7\times7$ convolution is replaced with a $3\times3$ convolution. Additionally, we remove the first max-pooling layer.
Following \cite{goyal2017accurate}, we zero-initialise the last batch normalisation layer in residual blocks so that they act as identity functions at the start of training. 
Because the output block of a ResNet expects to receive activations with a fixed number of channels, we add \emph{adaption layers}. We implement these using $1\times1$ convolutions. See \cref{app:resnet_impl} for an example implementation. \Cref{fig:upscaling_model} shows this modified computational model.
Note, however, that this channel mismatch issue is a specific instance of a more general problem of shape and size mismatch between layers in a \gls{DUN}. Consider the following cases where constructing a DUN requires adaption layers:
\begin{itemize}
    \item A \gls{NN} consisting of series of dense layers of different dimensions. E.g. an auto-encoder or U-Net.
    \item A \gls{NN} consisting of a mix of convolutional and dense layers. E.g. LeNet~\citep{lecun1998gradient}.
\end{itemize}
In the first case, we will have dimensionality mismatches between the different dense layers. In the second case, we will have shape mismatches between the 3D convolutional layers and the 1D dense layers, in addition to the potential size mismatches between layers of the same type. 
Fortunately, adaption layers can be used to solve any shape and size mismatches. Size mismatches can be naively solved by either padding or cropping tensors as appropriate. Another solution is to use (parameter) cheap\footnote{For ResNet-50, which contains 23.52~M parameters, a DUN over the first 13 blocks (using $1\times1$ convolution adaption layers) contains 26.28~M parameters which is an increase of only 1.17\%.} $1\times1$ convolution layers and low-rank\footnote{A low rank dense layer with input of size $n_1$ and output of size $n_2$ can be constructed by composing two standard dense layers of size $n_1 \times n_3$ and $n_3 \times n_2$ where $n_3 << n_1, n_2$.} dense layers in the case of mismatches between numbers of channels and numbers of dimensions, respectively.

For the MNIST and Fashion-MNIST datasets, we train DUNs with a fixed approximate posterior $q_{\balpha}(d) = p_{\bbeta}(d)$ for the first 3 epochs. These are the simplest image dataset we work with and can be readily solved with shallower models than ResNet-50. By fixing, $q_{\balpha}(d)$ for the first epochs, we ensure all layers receive strong gradients and become useful for making predictions. 

\subsubsection{Evaluation}

All methods are trained 5 times on each dataset, allowing for error bars in experiments. We report mean values and standard deviations.

To evaluate the methods' resilience to out of distribution data, we follow \cite{snoek1029can}. We train each method on MNIST and evaluate their predictive distributions on increasingly rotated digits.
We also train models on CIFAR10 and evaluate them on data submitted to 16 different corruptions \citep{hendrycks2019benchmarking} with 5 levels of severity each. Per severity results are provided.

We simulate a realistic \gls{OOD} rejection scenario \citep{oatml2019bdlb} by jointly evaluating our models on an in-distribution and an \gls{OOD} test set. We allow our methods to reject increasing proportions of the data based on predictive entropy before classifying the rest. All predictions on \gls{OOD} samples are treated as incorrect. In the main text we provide results with CIFAR10-SVHN as the in-out of distribution dataset pair. Results for the other pairs are found in \cref{app:additional_image_results}. We also perform \gls{OOD} detection experiments, where we evaluate methods' capacity to distinguish in-distribution and \gls{OOD} points using predictive entropy.

For all datasets, we compute run times per batch of size of 256 samples on two P100 GPUs. Results are obtained as averages of 5 independent runs. Ensembles and Dropout require multiple forward passes per batch. We report the time taken for all passes to be made. For Ensembles, we also include network loading time. This is because, in most cases, keeping 5 ResNet-50's in memory is unrealistic.

\subsection{Datasets}

We employ the following datasets in \cref{sec:experiments}. These are summarized in \cref{tab:datasets}.

\paragraph{Regression:}
\begin{itemize}
    \item UCI with standard splits \citep{hernandez2015probabilistic}
    \item UCI with gap splits \citep{foong2019inbetween}
    \item Flights (Airline Delay) \citep{hensman2013Gaussian}
\end{itemize}

\paragraph{Image Classification:}
\begin{itemize}
    \item MNIST \citep{lecun1998gradient}
    \item Fashion-MNIST \citep{xiao2017fashion}
    \item Kuzushiji-MNIST \citep{clanuwat2018deep}
    \item CIFAR10/100 \citep{krizhevsky2009learning} and Corrupted CIFAR \citep{hendrycks2019benchmarking}
    \item SVHN \citep{netzer2011reading}
\end{itemize}

\begin{table}[h]
    \centering
    \caption{Summary of datasets. For non-UCI datasets, the test and train set sizes are shown in brackets, e.g. (test \& train). For the standard UCI splits 90\% of the data is used for training and 10\% for validation. For the gap splits 66\% is used for training and 33\% for validation. Note that for the UCI datasets, only the standard number of splits are given since the number of gap splits is equal to the input dimensionality. }
    \resizebox{\textwidth}{!}{\begin{tabular}{l|ccccc}
    \toprule
         \textsc{Name} & \textsc{Size} & \textsc{Input Dim.} & \textsc{No. Classes} & \textsc{No. Splits} \\
         \midrule
         Boston Housing & 506 & 13 & -- & 20 \\
         Concrete Strength & 1,030 & 8 & -- & 20 \\
         Energy Efficiency & 768 & 8 & -- & 20 \\
         Kin8nm & 8,192 & 8 & -- & 20 \\
         Naval Propulsion & 11,934 & 16 & -- & 20 \\
         Power Plant & 9,568 & 4 & -- & 20 \\ 
         Protein Structure & 45,730 & 9 & -- & 5 \\
         Wine Quality Red & 1,599 & 11 & -- & 20 \\
         Yacht Hydrodynamics & 308 & 6 & -- & 20 \\
         \midrule
         Airline Delay & 2,055,733 (1,955,733 \& 100,000) & 8 & -- & 2 \\
         \midrule
         MNIST & 70,000 (60,000 \& 10,000) & 784 ($28\,\times\,28$) & 10 & 2 \\
         Fashion-MNIST & 70,000 (60,000 \& 10,000) & 784 ($28\,\times\,28$) & 10 & 2 \\
         Kuzushiji-MNIST & 70,000 (60,000 \& 10,000) & 784 ($28\,\times\,28$) & 10 & 2 \\
         CIFAR10 & 60,000 (50,000 \& 10,000) & 3072 ($32\,\times\,32\,\times\,3$) & 10 & 2 \\
         CIFAR100 & 60,000 (50,000 \& 10,000) & 3072 ($32\,\times\,32\,\times\,3$) & 100 & 2 \\
         SVHN & 99,289 (73,257 \& 26,032) & 3072 ($32\,\times\,32\,\times\,3$) & 10 & 2 \\
         \bottomrule
    \end{tabular}}
    \label{tab:datasets}
\end{table}

\section{Additional Results}\label{app:experiment_appendix}

\subsection{Toy Datasets}\label{app:toy_datasets}

\begin{figure}[h]
\vspace{-0.1cm}
 \centering
    \includegraphics[width=\linewidth]{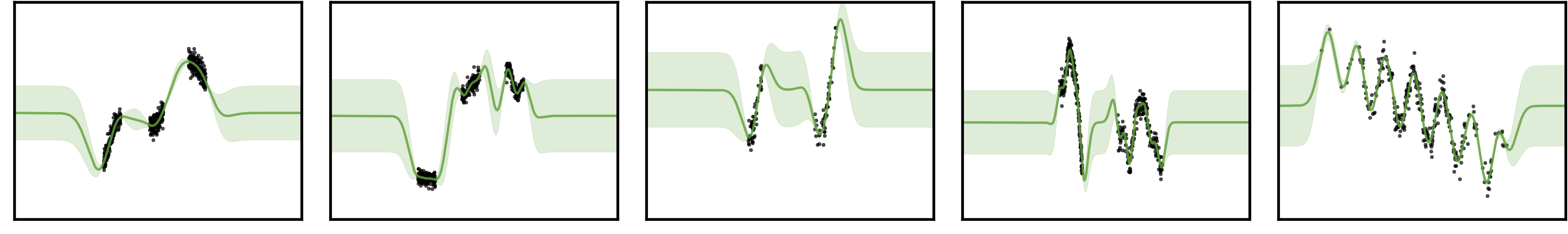}
    \caption{Fit obtained by a \gls{GP} with an RBF covariance function on the following datasets, from left to right: Simple\_1d, \citep{izmailov2019subspace}, \citep{foong2019inbetween}, Matern, Wiggle. Error bars represent the standard deviations of the distributions over functions.}
    \label{fig:appendix_GP_1d}
\vspace{-0.1in}
\end{figure}

In addition to the 1D toy dataset from \cite{izmailov2019subspace} and the Wiggle dataset introduced in \cref{sec:toy_datasets}, we conduct experiments on another three 1D toy datasets. Similarly to that of \cite{izmailov2019subspace}, the first of these datasets is composed of three disjoint clusters of inputs. However, these are arranged such that they can be fit by slower varying functions. We dub it ``Simple\_1d''. The second is the toy dataset used by \citet{foong2019inbetween} to evaluate the capacity of \gls{NN} approximate inference techniques to express model uncertainty in between disjoint clusters of data, also know as ``in-between'' uncertainty. The third is generated by sampling a function from a \gls{GP} with a Matern kernel with additive Gaussian noise. We dub it ``Matern''. We show all 5 1D toy datasets in \cref{fig:appendix_GP_1d}, where we fit them with a GP.

\subsubsection{Different Depths}\label{app:1d_toy_depths}
\begin{figure}[h]
\vspace{-0.1cm}
 \centering
    \includegraphics[width=0.8\linewidth]{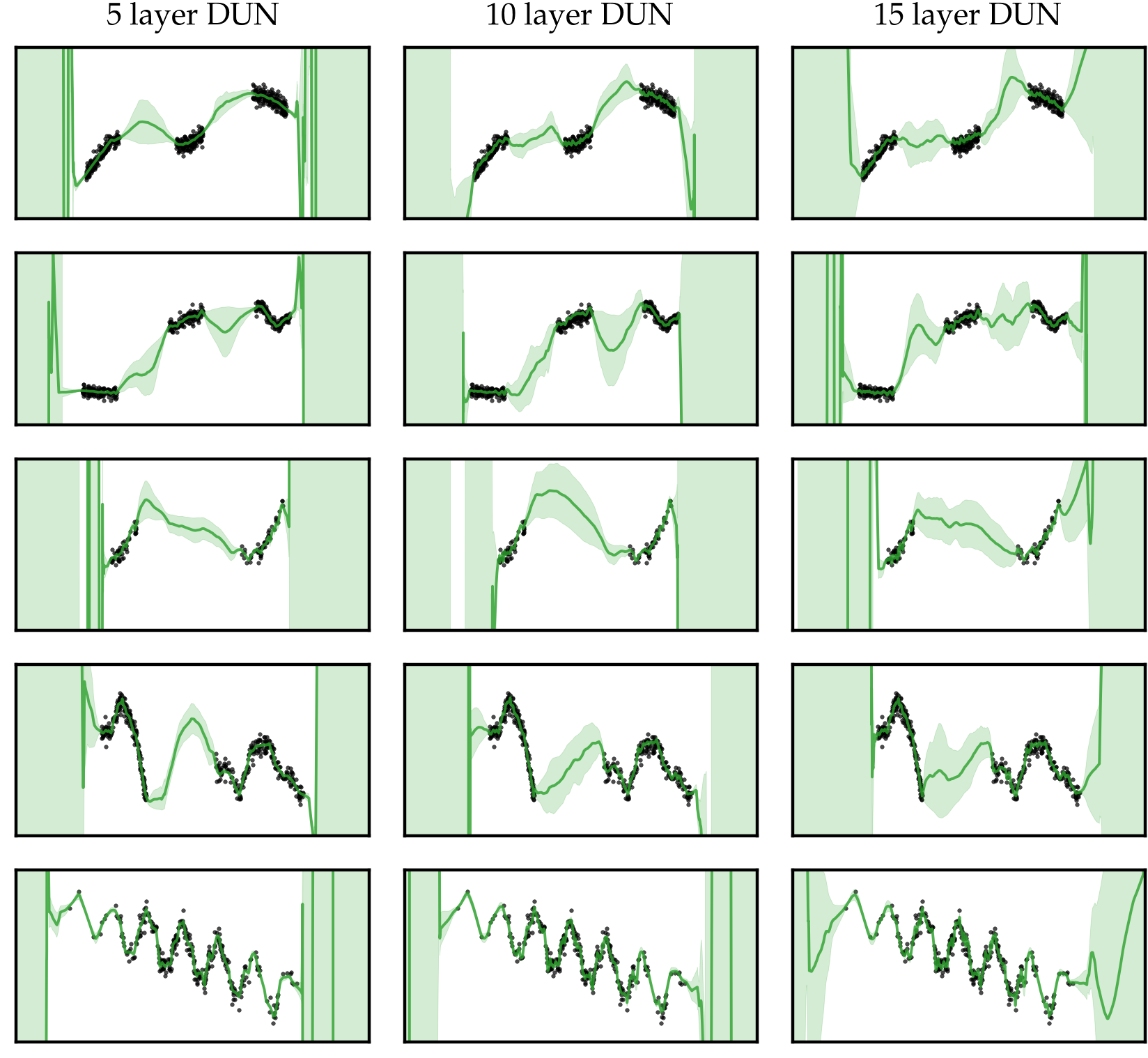}
    \caption{Increasing depth \glspl{DUN} trained on all 1d toy datasets. Each row corresponds to a different dataset. From top to bottom: Simple\_1d, \citep{izmailov2019subspace}, \citep{foong2019inbetween}, Matern, Wiggle.}
    \label{fig:appendix_DUN_toy_depths}
\vspace{-0.1in}
\end{figure}

In this section, we evaluate the effects of network depth on uncertainty estimates. We first train \glspl{DUN} of depths 5, 10 and 15 on all 1d toy datasets. The results are shown in \cref{fig:appendix_DUN_toy_depths}. \glspl{DUN} are able to fit all of the datasets well. However, the 5 layer versions provide noticeably smaller uncertainty estimates in between clusters of data. The uncertainty estimates from \glspl{DUN} benefit from depth for 2 reasons: increased depth means increasing the number of explanations of the data over which we perform \gls{BMA} and deeper subnetworks are able to express faster varying functions, which are more likely to disagree with each other. 

\begin{figure}[h]
\vspace{-0.1cm}
 \centering
    \includegraphics[width=0.71\linewidth]{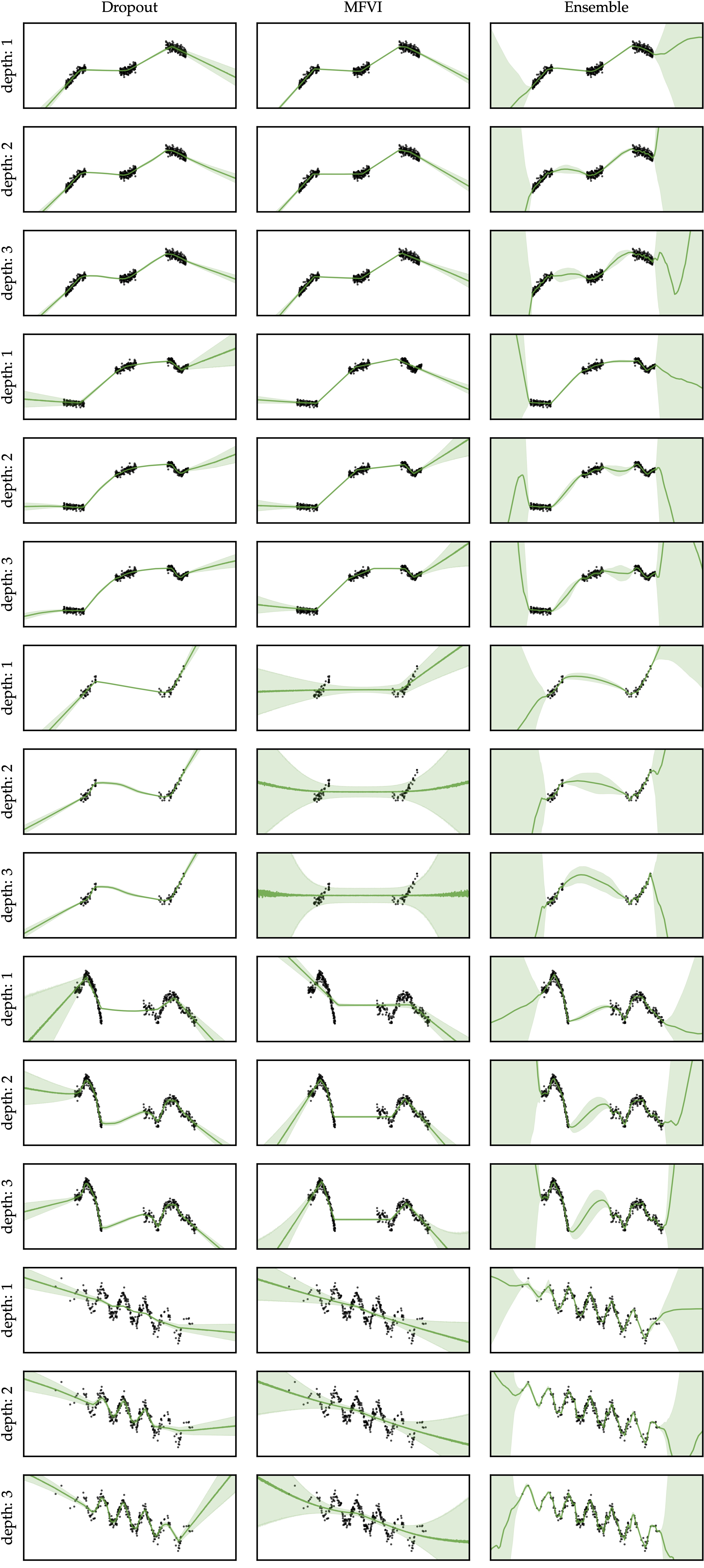}
    \caption{\gls{NN} baselines fit on all toy datasets. }
    \label{fig:appendix_all_depth_baselines}
\vspace{-0.1in}
\end{figure}

We also train each of our \gls{NN}-based baselines with depths 1, 2 and 3 on each of these datasets. Recall that by depth, we refer to the number of hidden layers. Results are shown in \cref{fig:appendix_all_depth_baselines}.

\cite{foong2019inbetween} prove that single hidden layer MFVI and dropout networks are unable to express high uncertainty in between regions of low uncertainty. Indeed, we observe this in our results. Further inline with the author's empirical observations, we find that deeper networks also fail to represent uncertainty in between clusters of points when making use of these approximate inference methods. Interestingly, the size of the error bars in the extrapolation regime seems to grow with depth for MFVI but shrink when using dropout. The amount of in-between and extrapolation uncertainty expressed by deep ensembles grows with depth. We attribute this to deeper models being able to express a wider range of functions, thus creating more opportunities for disagreement.  

Shallower dropout models tend to underfit faster varying functions, like Matern and Wiggle. For the latter, even the 3 hidden layer model underfits slightly, failing to capture the effects of the faster varying, lower amplitude sinusoid. MFVI completely fails to fit fast varying functions, even for deeper networks. Additionally, the functions it learns look piecewise linear. This might be the result of variational overprunning \citep{trippe2018overpruning}. Ensembles are able to fit all datasets well.


\clearpage

\subsubsection{Overcounting Data with MFVI}\label{app:overcounting_MFVI}

\begin{figure}[h]
\vspace{-0.1cm}
 \centering
    \includegraphics[width=0.9\linewidth]{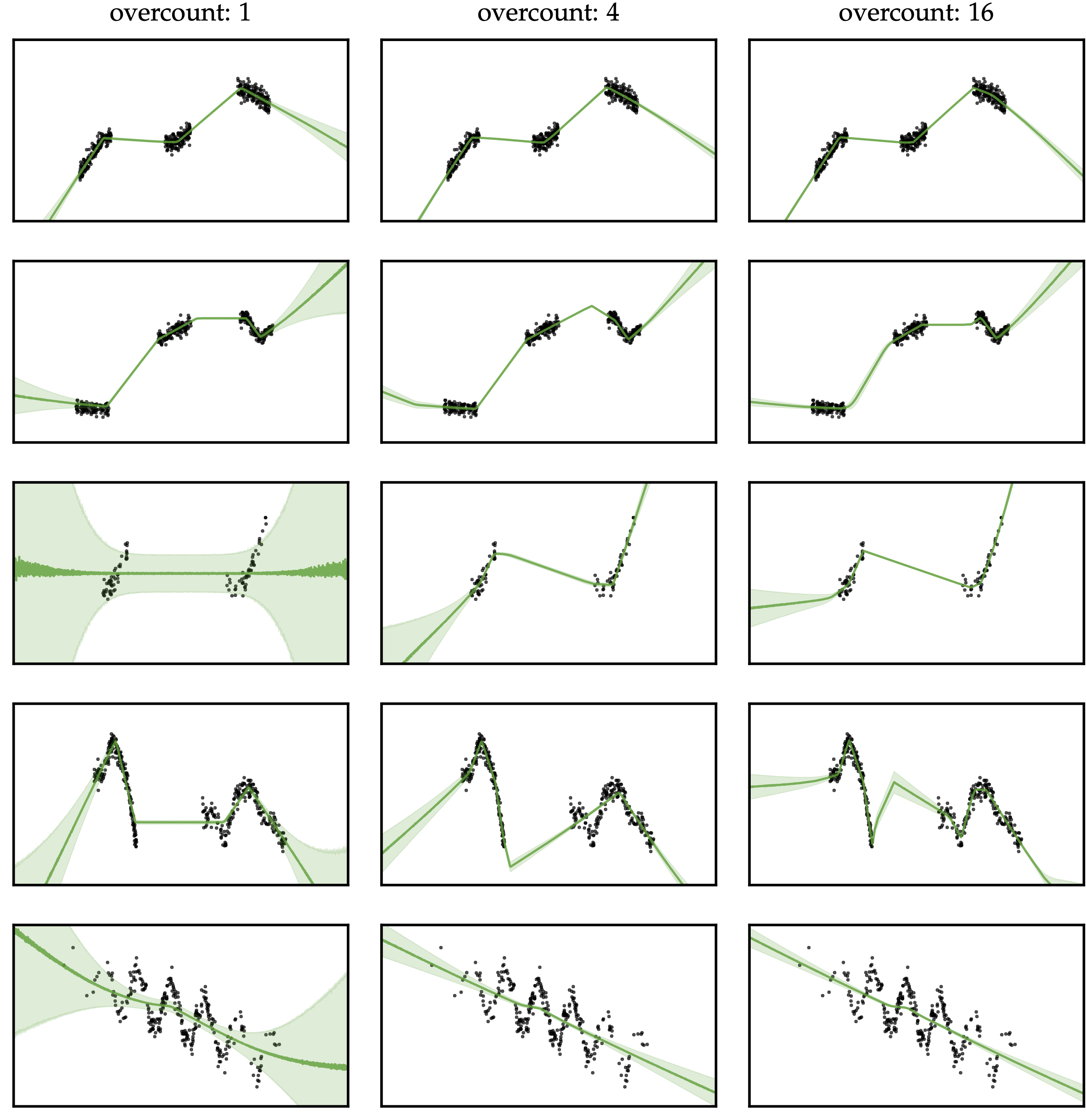}
    \caption{MFVI networks fit on all toy datasets for different overcount settings.}
    \label{fig:appendix_overcount_MFVI}
\vspace{-0.1in}
\end{figure}

In an attempt to fit MFVI networks to faster varying functions, we overcount the importance of the data in the ELBO. This type of objective targets what is often referred to as a tempered posterior \citep{wenzel2020good}: $p_{\text{overcount}}(\bff{w} | \dset) \propto p(\dset|\bff{w})^{T} p(\bff{w})$.
\begin{gather*}
    \mathrm{ELBO}_{\text{overcount}} = -KL(q(\bff{w}) || p(\bff{w})) + T \cdot \EX_{q(\bff{w})}[\sum_{n=1}^{N} p(y^{(n)} | \bff{x}^{(n)}, \bff{w})]
\end{gather*}
We experiment by setting the overcounting factor $T$ to the values: 1, 4 and 16. The results are shown in \cref{fig:appendix_overcount_MFVI}. Although increasing the relative importance of the data dependent likelihood term in the ELBO helps MFVI fit the Matern dataset and the dataset from \citet{foong2019inbetween}, the method still fails to fit Wiggle. Overcounting the data results in smaller error bars. 

\subsubsection{2d Toy Datasets}\label{app:2d_toy}

\begin{figure}[h]
\vspace{-0.1cm}
 \centering
    \includegraphics[width=\linewidth]{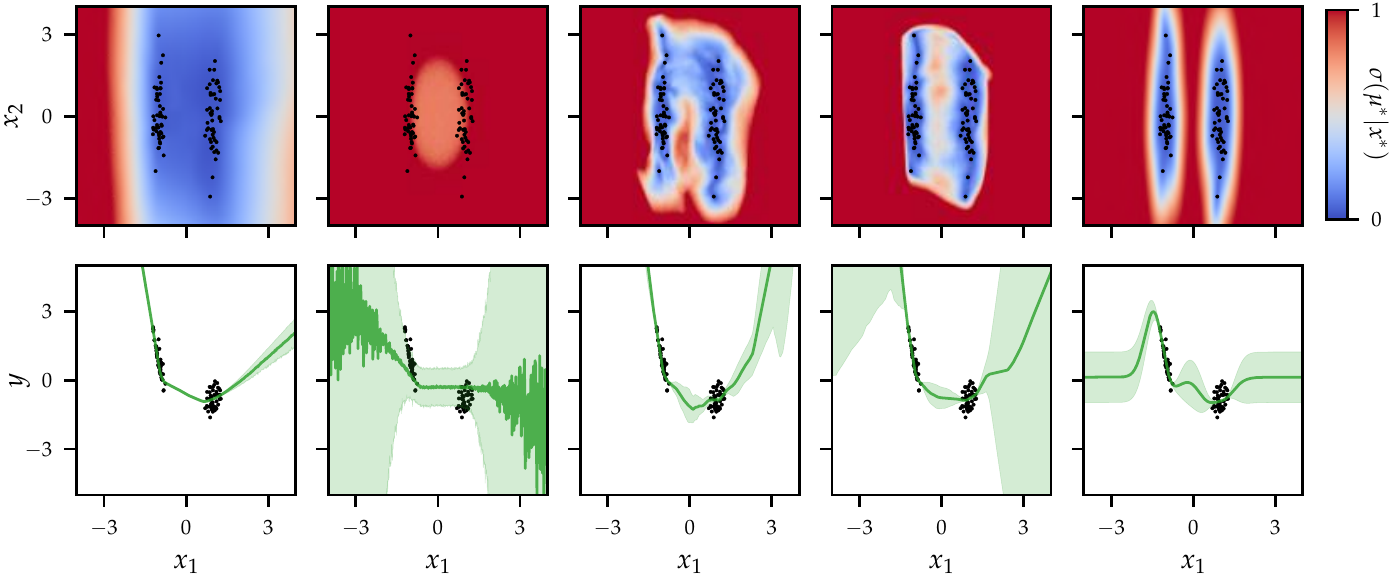}
    \caption{All methods under consideration trained on the Axis dataset. The top row shows the standard deviation values provided by each method for each point in the 2d input space. The bottom plot shows each method's predictions on a cross section of the input space at $x_2{=}0$. From left to right, the following methods are shown: dropout, MFVI, \gls{DUN}, deep ensembles, \gls{GP}.}
    \label{fig:appendix_axis}
\end{figure}

We evaluate the approaches under consideration on two 2d toy datasets, taken from \cite{foong2019pathologies}. These are dubbed Axis, \cref{fig:appendix_axis}, and Origin, \cref{fig:fig:appendix_origin}. We employ 15 hidden layers with \glspl{DUN} and 3 hidden layers with all other approaches. 

\glspl{DUN} and ensembles do not provide significantly increased uncertainty estimates in the regions between clusters of data on the Axis dataset. Both methods perform well on Origin. Otherwise, all methods display similar properties as in previous sections.

\begin{figure}[h]
\vspace{-0.1cm}
 \centering
    \includegraphics[width=\linewidth]{plots/appendix/appendix_origin.pdf}
    \caption{All methods under consideration trained on the Origin dataset. The top row shows the standard deviation values provided by each method for each point in the 2d input space. The bottom plot shows each method's predictions on a cross section of the input space. From left to right, the following methods are shown: dropout, MFVI, \gls{DUN}, deep ensembles, \gls{GP}.}
    \label{fig:fig:appendix_origin}
\vspace{-0.1in}
\end{figure}
\subsubsection{Non-residual Models}\label{app:non_residual_DUN}

\begin{figure}[h]
\vspace{-0.1cm}
 \centering
    \includegraphics[width=0.8\linewidth]{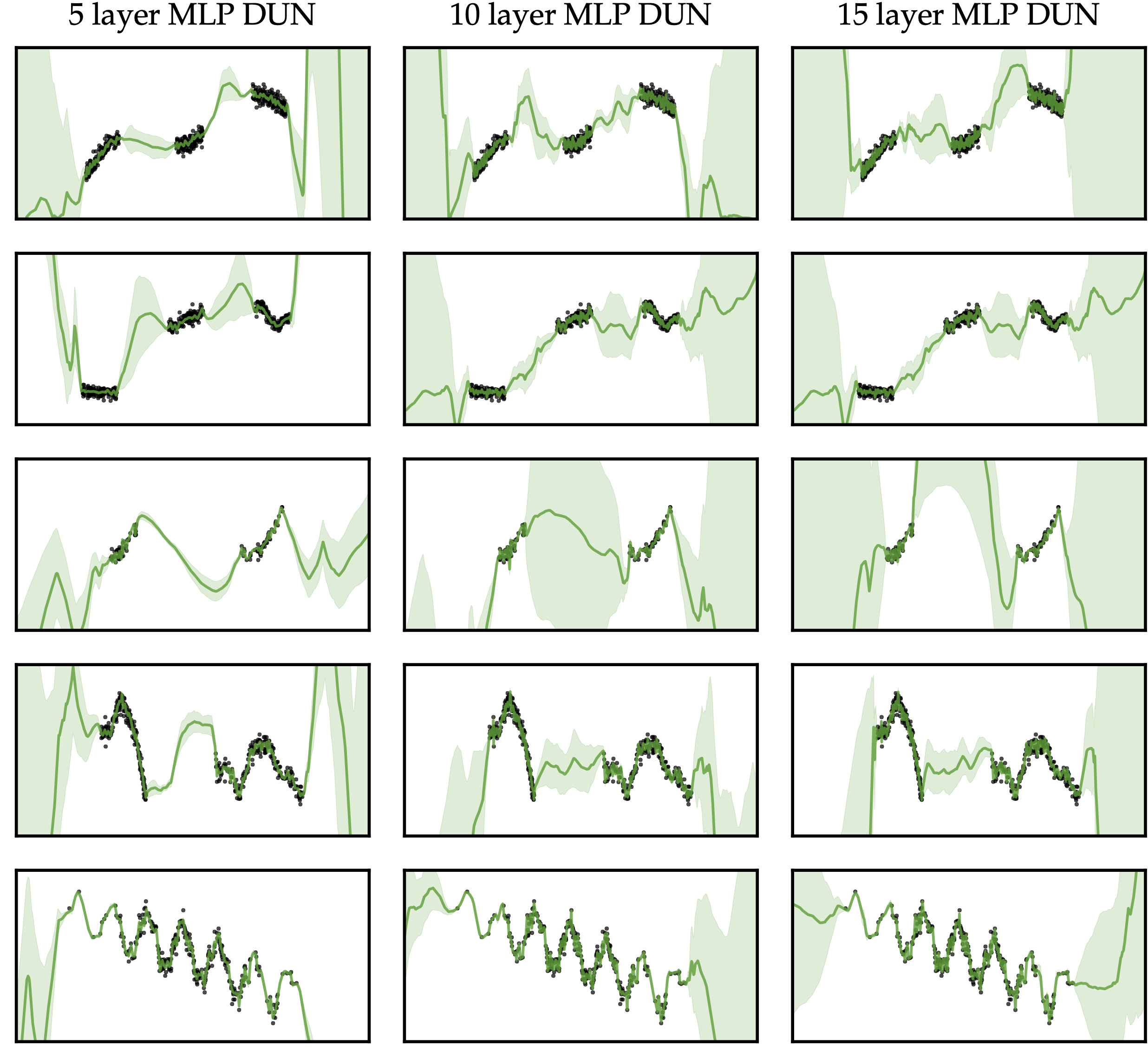}
    \caption{\glspl{DUN} with an MLP architecture trained on 1d toy datasets.}
    \label{fig:appendix_MLP_DUN}
\vspace{-0.1in}
\end{figure}

We employ residual architectures for most experiments in this work. This subsection explores the effect of residual connections on \glspl{DUN}. We first fit non-residual (MLP) \glspl{DUN} on all of our 1d toy datasets. The results are given in \cref{fig:appendix_MLP_DUN}. The learnt functions resemble those obtained with residual networks in \cref{fig:appendix_DUN_toy_depths}. However, non-residual \glspl{DUN} tend to provide less consistent uncertainty estimates in the extrapolation regime, especially when working with shallower models.

\begin{figure}[h]
\vspace{-0.1cm}
 \centering
    \includegraphics[width=0.8\linewidth]{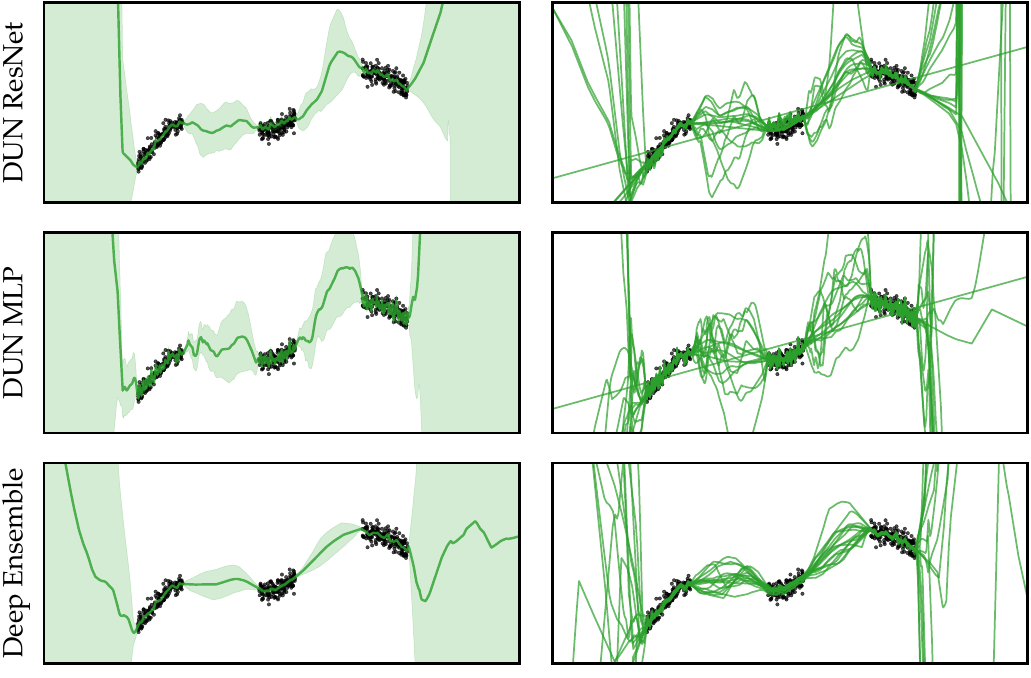}
    \caption{We fit the Simple\_1d toy dataset with 15 a layer MLP \gls{DUN}, a 15 layer residual \gls{DUN} and a 20 network deep ensemble with 3 hidden layers per network. The leftmost plot shows mean predictions and standard deviations corresponding to model uncertainty. The rightmost plot shows individual predictions from \gls{DUN} subnetworks and ensemble elements.}
    \label{fig:appendix_res_MLP_ensemble}
\vspace{-0.1in}
\end{figure}

We further compare the in-distribution fits from residual \glspl{DUN}, MLP \glspl{DUN}, and deep ensembles in \cref{fig:appendix_res_MLP_ensemble}. Ensemble elements differ slightly from each other in their predictions within the data dense regions. These predictions are averaged, making for mostly smooth functions. Functions expressed at most depths of the MLP \glspl{DUN} seem to vary together rapidly within the data region. Their mean prediction also varies rapidly, suggesting overfitting. In an MLP architecture, each successive layer only receives the previous one's output as its input. We hypothesize that, because of this structure, once a layer overfits a data point, the following layer is unlikely to modify the function in the area of that data point, as that would increase the training loss. This leads to most subnetworks only disagreeing about their predictions out of distribution. Functions expressed by residual \glspl{DUN} differ somewhat in-distribution, allowing some robustness to overfitting. We hypothesize that this occurs because each layer takes a linear combination of all previous layers' activations as its input. This prevents re-using the previous subnetworks' fits.  

Ensembles provide diverse explanations both in and out of distribution. This results in both better accuracy and predictive uncertainty than single models. \glspl{DUN} provide explanations which differ from each other mostly out of distribution. They provide uncertainty estimates out of distribution but their accuracy on in-distribution points is similar to that of a single model. 

\FloatBarrier
\subsection{Regression} \label{app:additional_regression_results}

In \cref{sec:tabresres}, we discussed the performance of \glspl{DUN} compared with \gls{SGD}, Dropout, Ensembles, and \gls{MFVI}, in terms of \gls{LL}, \gls{RMSE}, \gls{TCE}, and batch time. In this section, we elaborate by providing an additional metric: \gls{RCE}, discussed in \cref{app:evaluating_uncertainties}. We also further investigate the predictive performance to prediction time trade-off and provide results for the UCI gap splits.

\begin{figure}
    \centering
    \includegraphics[width=\textwidth]{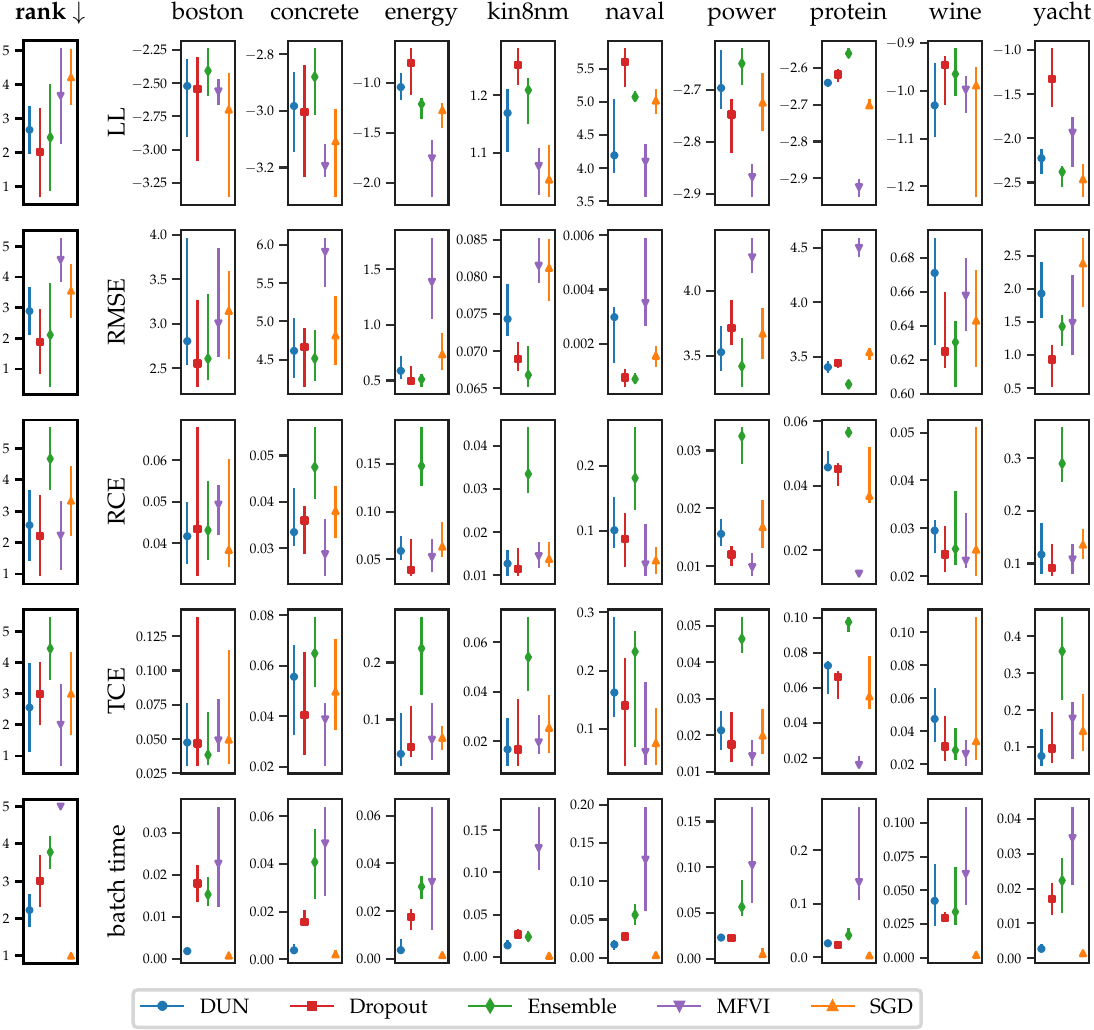}
    \caption{Quartiles for results on UCI regression datasets across \emph{standard splits}. Average ranks are computed across datasets. For LL, higher is better. Otherwise, lower is better.}
    \label{fig:uci_std_res_app}
\end{figure}

UCI standard split results are found in \cref{fig:uci_std_res_app}. As before, we rank methods from 1 to 5 based on mean performance, reporting mean ranks and standard deviations. Dropout obtains the best mean rank in terms of \gls{RMSE}, followed closely by Ensembles. \glspl{DUN} are third, significantly ahead of MFVI and SGD. Even so, \glspl{DUN} outperform Dropout and Ensembles in terms of \gls{TCE}, i.e. \glspl{DUN} more reliably assign large error bars to points on which they make incorrect predictions. Consequently, in terms of LL, a metric which considers both uncertainty and accuracy, \glspl{DUN} perform competitively (the LL rank distributions for all three methods overlap almost completely). 
However, on an alternate uncertainty metric, \gls{RCE}, Dropout tends to outperform \glspl{DUN}. This is indicative that the Dropout predictive posterior is better approximated by a Gaussian than DUNs' predictive posterior.  Ensembles still performs poorly and is only better than \gls{SGD}.
MFVI provides the best calibrated uncertainty estimates according to \gls{TCE} and ties with Dropout according to \gls{RCE}. Despite this, its mean predictions are inaccurate, as evidenced by it being last in terms of RMSE. This leads to \gls{MFVI}'s LL rank only being better than SGD's.

\begin{figure}
    \centering
    \includegraphics[width=\textwidth]{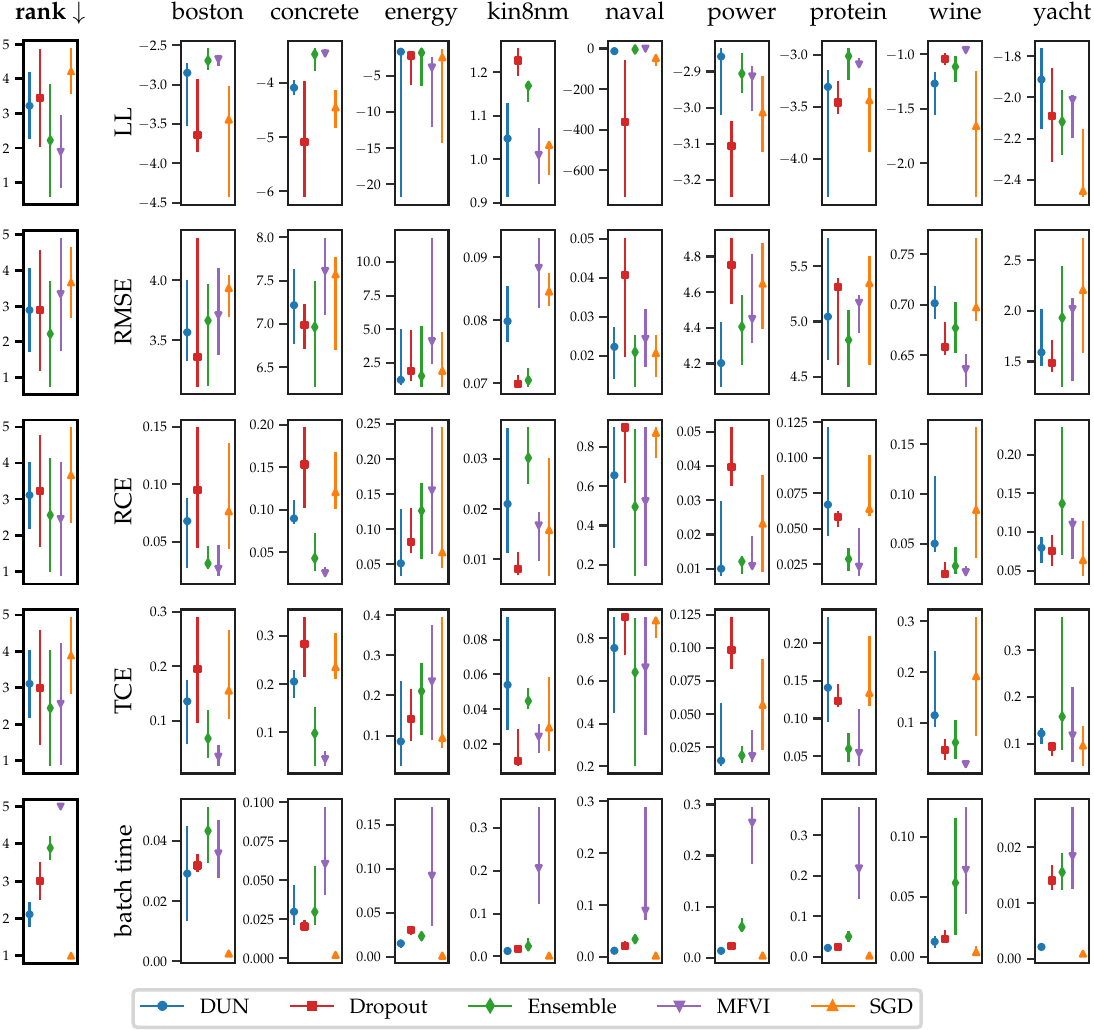}
    \caption{Quartiles for results on UCI regression datasets across \emph{gap splits}. Average ranks are computed across datasets. For LL, higher is better. Otherwise, lower is better.}
    \label{fig:uci_gap_res}
\end{figure}

\cref{fig:uci_gap_res} shows results for gap splits, designed to evaluate methods' capacity to express in-between uncertainty. All methods tend to perform worse in terms of the predictive performance metrics, indicating that the gap splits represent a more challenging problem. This trend is exemplified in the naval, and to a lesser extent, the energy datasets.
Here, \glspl{DUN} outperform Dropout in terms of LL rank. However, they are both outperformed by MFVI and Ensembles. DUNs consistently outperform multiple forward pass methods in terms of prediction time.

\begin{figure}
    \centering
    \includegraphics[width=\textwidth]{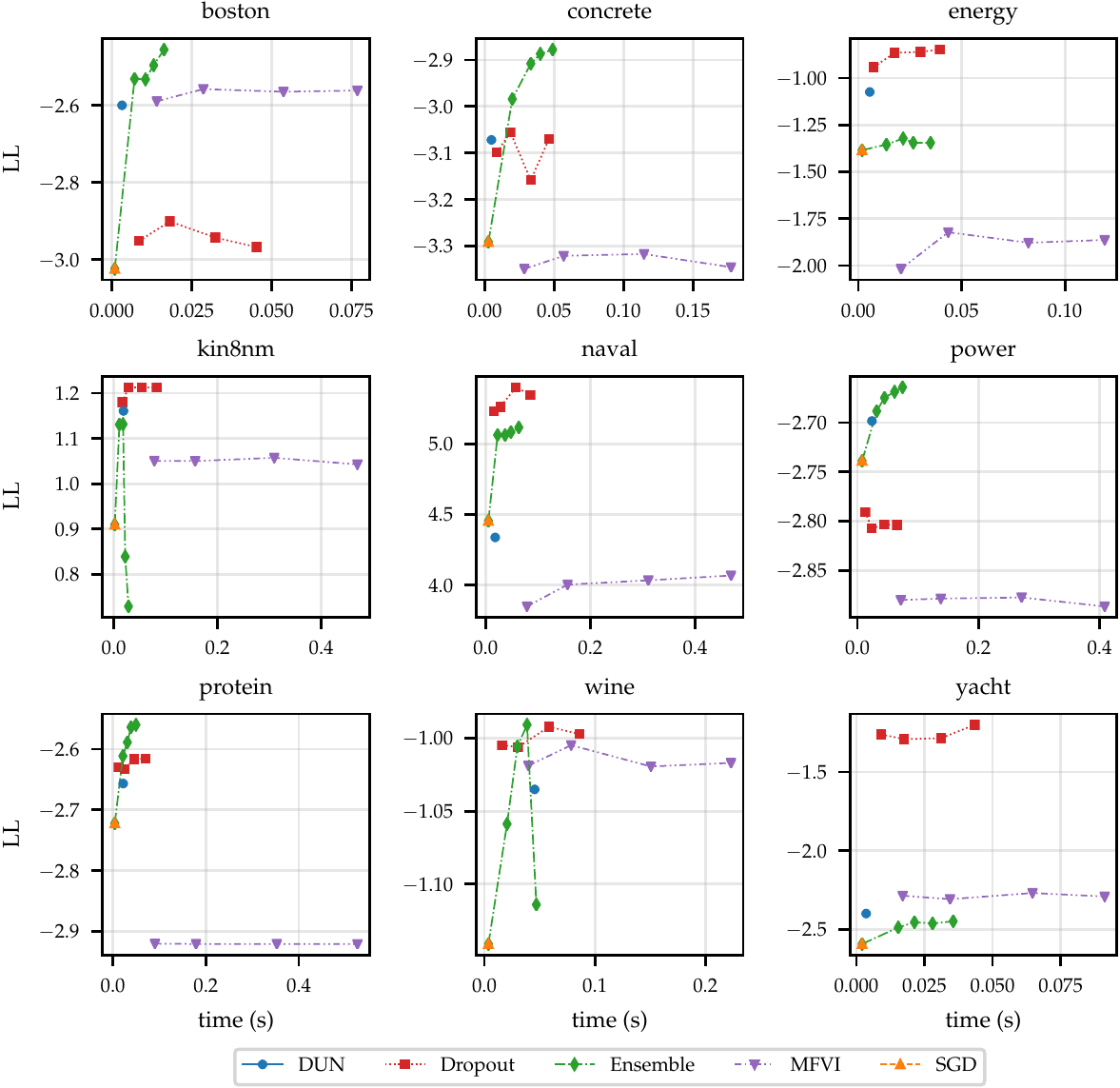}
    \caption{Pareto frontiers showing mean LL vs batch prediction time on the UCI datasets with standard splits. MFVI and Dropout are shown for 5, 10, 20, and 30 samples. Ensembles are shown with 1, 2, 3, 4, and 5 elements. Note that a single element ensemble is equivalent to SGD. Top left is better. Bottom right is worse. Timing includes overhead such as ensemble element loading. }
    \label{fig:uci_timing}
\end{figure}

In \cref{fig:uci_timing}, we show \gls{LL} vs batch time Pareto curves for all methods under consideration on the UCI datasets with standard splits. DUNs are Pareto efficient in 5 datasets, performing competitively in all of them. Dropout and Ensembles also tend to perform well. 

Finally, in \cref{tab:std_reg_results} and \cref{tab:gap_reg_res}, we provide mean and standard deviation results for both UCI standard and gap splits.

\begin{table}[h]
    \centering
    \caption{Mean values and standard deviations for results on UCI regression datasets across \emph{standard splits}. Bold blue text denotes the best mean value for each dataset and each metric. Bold red text denotes the worst mean value. Note that in some cases the best/worst mean values are within error of other mean values.}
    \label{tab:std_reg_results}
    \resizebox{\textwidth}{!}{\begin{tabular}{ll|rrrrrr}
    \toprule
               & \textsc{Method} & \multicolumn{1}{c}{\textsc{DUN}} &   \multicolumn{1}{c}{\textsc{DUN (MLP)}} & \multicolumn{1}{c}{\textsc{Dropout}} & \multicolumn{1}{c}{\textsc{Ensemble}} & \multicolumn{1}{c}{\textsc{MFVI}} & \multicolumn{1}{c}{\textsc{SGD}} \\
    \textsc{Metric} & \textsc{Dataset} &                                  &                                  &                                  &                                  &                                  &                                  \\
    \midrule
    \multirow{9}{*}{LL $\uparrow$} & boston &  $-2.604 \scriptstyle \pm 0.351$ &  $-2.604 \scriptstyle \pm 0.368$ &  $-2.882 \scriptstyle \pm 1.028$ &  $\hbest{-2.454} \scriptstyle \pm 0.275$ &  $-2.573 \scriptstyle \pm 0.136$ &  $\hworst{-2.942} \scriptstyle \pm 0.676$ \\
               & concrete &  $-3.005 \scriptstyle \pm 0.212$ &  $-3.051 \scriptstyle \pm 0.278$ &  $-3.051 \scriptstyle \pm 0.308$ &  $\hbest{-2.886} \scriptstyle \pm 0.153$ &  $-3.190 \scriptstyle \pm 0.110$ &  $\hworst{-3.214} \scriptstyle \pm 0.399$ \\
               & energy &  $-1.037 \scriptstyle \pm 0.159$ &  $-1.564 \scriptstyle \pm 0.383$ &  $\hbest{-0.975} \scriptstyle \pm 0.509$ &  $-1.298 \scriptstyle \pm 0.210$ &  $\hworst{-1.961} \scriptstyle \pm 0.648$ &  $-1.348 \scriptstyle \pm 0.225$ \\
               & kin8nm &   $1.151 \scriptstyle \pm 0.083$ &   $1.111 \scriptstyle \pm 0.103$ &   $\hbest{1.231} \scriptstyle \pm 0.086$ &   $0.813 \scriptstyle \pm 1.224$ &   $1.055 \scriptstyle \pm 0.084$ &   $\hworst{0.905} \scriptstyle \pm 0.778$ \\
               & naval &   $4.245 \scriptstyle \pm 1.108$ &   $4.472 \scriptstyle \pm 1.239$ &   $\hbest{5.429} \scriptstyle \pm 0.735$ &   $5.081 \scriptstyle \pm 0.156$ &   $\hworst{3.389} \scriptstyle \pm 2.891$ &   $4.821 \scriptstyle \pm 0.621$ \\
               & power &  $-2.695 \scriptstyle \pm 0.086$ &  $-2.719 \scriptstyle \pm 0.069$ &  $-2.790 \scriptstyle \pm 0.118$ &  $\hbest{-2.663} \scriptstyle \pm 0.055$ &  $\hworst{-2.877} \scriptstyle \pm 0.041$ &  $-2.733 \scriptstyle \pm 0.081$ \\
               & protein &  $-2.657 \scriptstyle \pm 0.044$ &  $-2.692 \scriptstyle \pm 0.020$ &  $-2.623 \scriptstyle \pm 0.036$ &  $\hbest{-2.561} \scriptstyle \pm 0.026$ &  $\hworst{-2.929} \scriptstyle \pm 0.038$ &  $-2.717 \scriptstyle \pm 0.064$ \\
               & wine &  $-1.031 \scriptstyle \pm 0.119$ &  $\hbest{-0.979} \scriptstyle \pm 0.113$ &  $-1.003 \scriptstyle \pm 0.128$ &  $-1.116 \scriptstyle \pm 0.582$ &  $-1.007 \scriptstyle \pm 0.063$ &  $\hworst{-1.212} \scriptstyle \pm 0.485$ \\
               & yacht &  $-2.420 \scriptstyle \pm 0.523$ &  $-2.463 \scriptstyle \pm 0.197$ &  $\hbest{-1.330} \scriptstyle \pm 0.436$ &  $-2.441 \scriptstyle \pm 0.189$ &  $-2.238 \scriptstyle \pm 0.952$ &  $\hworst{-2.525} \scriptstyle \pm 0.354$ \\
    \midrule
    \multirow{9}{*}{RMSE $\downarrow$} & boston &   $3.200 \scriptstyle \pm 0.978$ &   $3.157 \scriptstyle \pm 0.885$ &   $\hbest{2.832} \scriptstyle \pm 0.768$ &   $2.835 \scriptstyle \pm 0.808$ &   $\hworst{3.218} \scriptstyle \pm 0.837$ &   $\hworst{3.218} \scriptstyle \pm 0.904$ \\
               & concrete &   $4.613 \scriptstyle \pm 0.607$ &   $4.571 \scriptstyle \pm 0.703$ &   $4.610 \scriptstyle \pm 0.572$ &   $\hbest{4.552} \scriptstyle \pm 0.582$ &   $\hworst{5.894} \scriptstyle \pm 0.742$ &   $4.983 \scriptstyle \pm 0.914$ \\
               & energy &   $0.612 \scriptstyle \pm 0.157$ &   $0.948 \scriptstyle \pm 0.474$ &   $0.571 \scriptstyle \pm 0.204$ &   $\hbest{0.507} \scriptstyle \pm 0.110$ &   $\hworst{1.686} \scriptstyle \pm 1.016$ &   $0.797 \scriptstyle \pm 0.283$ \\
               & kin8nm &   $0.076 \scriptstyle \pm 0.005$ &   $0.077 \scriptstyle \pm 0.006$ &   $\hbest{0.070} \scriptstyle \pm 0.005$ &   $\hworst{0.304} \scriptstyle \pm 0.991$ &   $0.084 \scriptstyle \pm 0.007$ &   $0.202 \scriptstyle \pm 0.544$ \\
               & naval &   $0.003 \scriptstyle \pm 0.002$ &   $0.002 \scriptstyle \pm 0.001$ &   $\hbest{0.001} \scriptstyle \pm 0.001$ &   $\hbest{0.001} \scriptstyle \pm 0.000$ &   $\hworst{0.005} \scriptstyle \pm 0.005$ &   $0.002 \scriptstyle \pm 0.001$ \\
               & power &   $3.573 \scriptstyle \pm 0.254$ &   $3.671 \scriptstyle \pm 0.247$ &   $3.823 \scriptstyle \pm 0.350$ &   $\hbest{3.444} \scriptstyle \pm 0.238$ &   $\hworst{4.286} \scriptstyle \pm 0.179$ &   $3.697 \scriptstyle \pm 0.272$ \\
               & protein &   $3.402 \scriptstyle \pm 0.058$ &   $3.412 \scriptstyle \pm 0.076$ &   $3.425 \scriptstyle \pm 0.070$ &   $\hbest{3.260} \scriptstyle \pm 0.074$ &   $\hworst{4.511} \scriptstyle \pm 0.145$ &   $3.589 \scriptstyle \pm 0.174$ \\
               & wine &   $0.659 \scriptstyle \pm 0.061$ &   $0.629 \scriptstyle \pm 0.047$ &   $\hbest{0.642} \scriptstyle \pm 0.049$ &   $\hworst{1.934} \scriptstyle \pm 5.708$ &   $0.660 \scriptstyle \pm 0.040$ &   $0.652 \scriptstyle \pm 0.054$ \\
               & yacht &   $2.514 \scriptstyle \pm 1.985$ &   $2.465 \scriptstyle \pm 0.841$ &   $\hbest{0.876} \scriptstyle \pm 0.411$ &   $1.429 \scriptstyle \pm 0.483$ &   $\hworst{3.419} \scriptstyle \pm 7.333$ &   $2.352 \scriptstyle \pm 0.905$ \\
    \midrule
    \multirow{9}{*}{RCE $\downarrow$} & boston &   $0.045 \scriptstyle \pm 0.016$ &   $\hbest{0.043} \scriptstyle \pm 0.013$ &   $\hworst{0.058} \scriptstyle \pm 0.037$ &   $0.046 \scriptstyle \pm 0.015$ &   $0.049 \scriptstyle \pm 0.014$ &   $0.052 \scriptstyle \pm 0.032$ \\
               & concrete &   $0.037 \scriptstyle \pm 0.011$ &   $0.039 \scriptstyle \pm 0.011$ &   $0.036 \scriptstyle \pm 0.011$ &   $\hworst{0.053} \scriptstyle \pm 0.020$ &   $\hbest{0.030} \scriptstyle \pm 0.008$ &   $0.040 \scriptstyle \pm 0.016$ \\
               & energy &   $0.064 \scriptstyle \pm 0.031$ &   $0.120 \scriptstyle \pm 0.069$ &   $\hbest{0.059} \scriptstyle \pm 0.047$ &   $\hworst{0.157} \scriptstyle \pm 0.052$ &   $0.070 \scriptstyle \pm 0.051$ &   $0.072 \scriptstyle \pm 0.031$ \\
               & kin8nm &   $\hbest{0.014} \scriptstyle \pm 0.007$ &   $0.021 \scriptstyle \pm 0.014$ &   $\hbest{0.014} \scriptstyle \pm 0.006$ &   $\hworst{0.090} \scriptstyle \pm 0.199$ &   $0.016 \scriptstyle \pm 0.007$ &   $0.028 \scriptstyle \pm 0.051$ \\
               & naval &   $0.134 \scriptstyle \pm 0.102$ &   $0.094 \scriptstyle \pm 0.123$ &   $0.100 \scriptstyle \pm 0.074$ &   $\hworst{0.191} \scriptstyle \pm 0.079$ &   $0.087 \scriptstyle \pm 0.108$ &   $\hbest{0.072} \scriptstyle \pm 0.073$ \\
               & power &   $0.016 \scriptstyle \pm 0.004$ &   $0.018 \scriptstyle \pm 0.005$ &   $0.015 \scriptstyle \pm 0.012$ &   $\hworst{0.032} \scriptstyle \pm 0.005$ &   $\hbest{0.010} \scriptstyle \pm 0.003$ &   $0.017 \scriptstyle \pm 0.005$ \\
               & protein &   $0.048 \scriptstyle \pm 0.005$ &   $0.045 \scriptstyle \pm 0.003$ &   $0.043 \scriptstyle \pm 0.006$ &   $\hworst{0.055} \scriptstyle \pm 0.007$ &   $\hbest{0.014} \scriptstyle \pm 0.003$ &   $0.041 \scriptstyle \pm 0.011$ \\
               & wine &   $0.030 \scriptstyle \pm 0.009$ &   $0.031 \scriptstyle \pm 0.013$ &   $\hbest{0.027} \scriptstyle \pm 0.009$ &   $\hworst{0.100} \scriptstyle \pm 0.214$ &   $0.028 \scriptstyle \pm 0.009$ &   $0.083 \scriptstyle \pm 0.195$ \\
               & yacht &   $0.141 \scriptstyle \pm 0.078$ &   $0.177 \scriptstyle \pm 0.066$ &   $\hbest{0.117} \scriptstyle \pm 0.068$ &   $\hworst{0.311} \scriptstyle \pm 0.089$ &   $0.156 \scriptstyle \pm 0.190$ &   $0.153 \scriptstyle \pm 0.085$ \\
    \midrule
    \multirow{9}{*}{TCE $\downarrow$} & boston &   $0.053 \scriptstyle \pm 0.034$ &   $\hbest{0.047} \scriptstyle \pm 0.030$ &   $\hworst{0.089} \scriptstyle \pm 0.076$ &   $0.055 \scriptstyle \pm 0.038$ &   $0.060 \scriptstyle \pm 0.035$ &   $0.082 \scriptstyle \pm 0.075$ \\
               & concrete &   $0.054 \scriptstyle \pm 0.027$ &   $0.048 \scriptstyle \pm 0.025$ &   $0.047 \scriptstyle \pm 0.028$ &   $\hworst{0.067} \scriptstyle \pm 0.032$ &   $\hbest{0.036} \scriptstyle \pm 0.020$ &   $0.060 \scriptstyle \pm 0.045$ \\
               & energy &   $\hbest{0.072} \scriptstyle \pm 0.073$ &   $0.103 \scriptstyle \pm 0.112$ &   $0.088 \scriptstyle \pm 0.085$ &   $\hworst{0.221} \scriptstyle \pm 0.101$ &   $0.097 \scriptstyle \pm 0.090$ &   $0.083 \scriptstyle \pm 0.057$ \\
               & kin8nm &   $\hbest{0.024} \scriptstyle \pm 0.022$ &   $0.042 \scriptstyle \pm 0.038$ &   $0.025 \scriptstyle \pm 0.021$ &   $\hworst{0.065} \scriptstyle \pm 0.054$ &   $\hbest{0.024} \scriptstyle \pm 0.015$ &   $0.031 \scriptstyle \pm 0.022$ \\
               & naval &   $\hworst{0.212} \scriptstyle \pm 0.159$ &   $0.127 \scriptstyle \pm 0.162$ &   $0.153 \scriptstyle \pm 0.128$ &   $\hworst{0.212} \scriptstyle \pm 0.147$ &   $0.118 \scriptstyle \pm 0.143$ &   $\hbest{0.112} \scriptstyle \pm 0.118$ \\
               & power &   $0.020 \scriptstyle \pm 0.007$ &   $0.022 \scriptstyle \pm 0.011$ &   $0.024 \scriptstyle \pm 0.026$ &   $\hworst{0.045} \scriptstyle \pm 0.010$ &   $\hbest{0.015} \scriptstyle \pm 0.006$ &   $0.020 \scriptstyle \pm 0.009$ \\
               & protein &   $0.069 \scriptstyle \pm 0.012$ &   $0.058 \scriptstyle \pm 0.011$ &   $0.063 \scriptstyle \pm 0.011$ &   $\hworst{0.094} \scriptstyle \pm 0.014$ &   $\hbest{0.020} \scriptstyle \pm 0.008$ &   $0.061 \scriptstyle \pm 0.017$ \\
               & wine &   $0.051 \scriptstyle \pm 0.028$ &   $0.047 \scriptstyle \pm 0.033$ &   $0.040 \scriptstyle \pm 0.028$ &   $0.088 \scriptstyle \pm 0.201$ &   $\hbest{0.027} \scriptstyle \pm 0.013$ &   $\hworst{0.109} \scriptstyle \pm 0.197$ \\
               & yacht &   $\hbest{0.122} \scriptstyle \pm 0.119$ &   $0.169 \scriptstyle \pm 0.131$ &   $0.131 \scriptstyle \pm 0.114$ &   $\hworst{0.341} \scriptstyle \pm 0.176$ &   $0.196 \scriptstyle \pm 0.207$ &   $0.175 \scriptstyle \pm 0.130$ \\
    \midrule
    \multirow{9}{*}{batch time $\downarrow$} & boston &   $0.003 \scriptstyle \pm 0.003$ &   $\hbest{0.001} \scriptstyle \pm 0.000$ &   $0.018 \scriptstyle \pm 0.006$ &   $0.016 \scriptstyle \pm 0.004$ &   $\hworst{0.029} \scriptstyle \pm 0.021$ &   $\hbest{0.001} \scriptstyle \pm 0.000$ \\
               & concrete &   $0.005 \scriptstyle \pm 0.003$ &   $\hbest{0.002} \scriptstyle \pm 0.001$ &   $0.019 \scriptstyle \pm 0.007$ &   $0.050 \scriptstyle \pm 0.035$ &   $\hworst{0.055} \scriptstyle \pm 0.042$ &   $0.003 \scriptstyle \pm 0.002$ \\
               & energy &   $0.007 \scriptstyle \pm 0.008$ &   $0.005 \scriptstyle \pm 0.002$ &   $0.017 \scriptstyle \pm 0.007$ &   $0.037 \scriptstyle \pm 0.020$ &   $\hworst{0.043} \scriptstyle \pm 0.037$ &   $\hbest{0.002} \scriptstyle \pm 0.001$ \\
               & kin8nm &   $0.019 \scriptstyle \pm 0.014$ &   $0.011 \scriptstyle \pm 0.008$ &   $0.029 \scriptstyle \pm 0.009$ &   $0.026 \scriptstyle \pm 0.008$ &   $\hworst{0.157} \scriptstyle \pm 0.097$ &   $\hbest{0.002} \scriptstyle \pm 0.001$ \\
               & naval &   $0.019 \scriptstyle \pm 0.010$ &   $0.012 \scriptstyle \pm 0.005$ &   $0.029 \scriptstyle \pm 0.009$ &   $0.065 \scriptstyle \pm 0.032$ &   $\hworst{0.156} \scriptstyle \pm 0.128$ &   $\hbest{0.005} \scriptstyle \pm 0.003$ \\
               & power &   $0.024 \scriptstyle \pm 0.007$ &   $0.016 \scriptstyle \pm 0.006$ &   $0.023 \scriptstyle \pm 0.006$ &   $0.074 \scriptstyle \pm 0.038$ &   $\hworst{0.138} \scriptstyle \pm 0.106$ &   $\hbest{0.007} \scriptstyle \pm 0.005$ \\
               & protein &   $0.022 \scriptstyle \pm 0.008$ &   $0.018 \scriptstyle \pm 0.002$ &   $0.024 \scriptstyle \pm 0.004$ &   $0.051 \scriptstyle \pm 0.022$ &   $\hworst{0.178} \scriptstyle \pm 0.099$ &   $\hbest{0.004} \scriptstyle \pm 0.002$ \\
               & wine &   $0.046 \scriptstyle \pm 0.026$ &   $0.028 \scriptstyle \pm 0.009$ &   $0.031 \scriptstyle \pm 0.006$ &   $0.046 \scriptstyle \pm 0.034$ &   $\hworst{0.078} \scriptstyle \pm 0.048$ &   $\hbest{0.004} \scriptstyle \pm 0.003$ \\
               & yacht &   $0.004 \scriptstyle \pm 0.003$ &   $0.003 \scriptstyle \pm 0.002$ &   $0.017 \scriptstyle \pm 0.005$ &   $0.035 \scriptstyle \pm 0.035$ &   $\hworst{0.038} \scriptstyle \pm 0.022$ &   $\hbest{0.002} \scriptstyle \pm 0.002$ \\
    \bottomrule
    \end{tabular}}
\end{table}

\begin{table}[]
    \centering
    \caption{Mean values and standard deviations for results on UCI regression datasets across \emph{gap splits}. Bold blue text denotes the best mean value for each dataset and each metric. Bold red text denotes the worst mean value. Note that in some cases the best/worst mean values are within error of other mean values.}
    \label{tab:gap_reg_res}
    \resizebox{\textwidth}{!}{\begin{tabular}{ll|rrrrrr}
    \toprule
               & \textsc{Method} & \multicolumn{1}{c}{\textsc{DUN}} &   \multicolumn{1}{c}{\textsc{DUN (MLP)}} & \multicolumn{1}{c}{\textsc{Dropout}} & \multicolumn{1}{c}{\textsc{Ensemble}} & \multicolumn{1}{c}{\textsc{MFVI}} & \multicolumn{1}{c}{\textsc{SGD}} \\
    \textsc{Metric} & \textsc{Dataset} &                                  &                                  &                                  &                                  &                                  &                                  \\
    \midrule
        \multirow{9}{*}{LL $\uparrow$} & boston &    $-3.107 \scriptstyle \pm 0.593$ &    $-3.033 \scriptstyle \pm 0.409$ &      $-4.001 \scriptstyle \pm 1.814$ &  $-3.106 \scriptstyle \pm 1.481$ &    $\hbest{-2.703} \scriptstyle \pm 0.072$ &    $\hworst{-4.217} \scriptstyle \pm 1.876$ \\
               & concrete &    $-4.222 \scriptstyle \pm 0.818$ &    $-4.152 \scriptstyle \pm 0.433$ &      $\hworst{-5.170} \scriptstyle \pm 1.376$ &  $-3.631 \scriptstyle \pm 0.523$ &    $\hbest{-3.460} \scriptstyle \pm 0.177$ &    $-4.839 \scriptstyle \pm 1.585$ \\
               & energy &  $-10.730 \scriptstyle \pm 13.477$ &    $-6.477 \scriptstyle \pm 7.516$ &     $-8.102 \scriptstyle \pm 13.796$ &  $\hbest{-5.423} \scriptstyle \pm 7.290$ &   $-9.093 \scriptstyle \pm 10.573$ &  $\hworst{-15.295} \scriptstyle \pm 26.058$ \\
               & kin8nm &     $1.029 \scriptstyle \pm 0.133$ &     $1.110 \scriptstyle \pm 0.073$ &       $\hbest{1.215} \scriptstyle \pm 0.049$ &   $\hworst{0.315} \scriptstyle \pm 2.397$ &     $0.942 \scriptstyle \pm 0.240$ &     $0.991 \scriptstyle \pm 0.131$ \\
               & naval &  $-16.279 \scriptstyle \pm 19.437$ &  $-19.165 \scriptstyle \pm 11.324$ &  $\hworst{-523.856} \scriptstyle \pm 570.116$ &  $\hbest{-4.573} \scriptstyle \pm 7.496$ &  $-15.208 \scriptstyle \pm 43.758$ &  $-60.470 \scriptstyle \pm 53.213$ \\
               & power &    $-2.998 \scriptstyle \pm 0.325$ &    $-2.961 \scriptstyle \pm 0.089$ &      $\hworst{-3.178} \scriptstyle \pm 0.224$ &  $\hbest{-2.904} \scriptstyle \pm 0.110$ &    $-2.980 \scriptstyle \pm 0.184$ &    $-3.022 \scriptstyle \pm 0.141$ \\
               & protein &    $\hworst{-3.835} \scriptstyle \pm 0.998$ &    $-3.553 \scriptstyle \pm 0.371$ &      $-3.459 \scriptstyle \pm 0.444$ &  $-\hbest{3.071} \scriptstyle \pm 0.261$ &    $-3.083 \scriptstyle \pm 0.086$ &    $-3.554 \scriptstyle \pm 0.408$ \\
               & wine &    $-1.417 \scriptstyle \pm 0.474$ &    $\hworst{-2.121} \scriptstyle \pm 1.227$ &      $-1.267 \scriptstyle \pm 0.677$ &  $-1.126 \scriptstyle \pm 0.137$ &    $\hbest{-0.965} \scriptstyle \pm 0.033$ &    $-2.026 \scriptstyle \pm 1.130$ \\
               & yacht &    $-2.122 \scriptstyle \pm 0.584$ &    $-2.165 \scriptstyle \pm 0.235$ &      $-2.344 \scriptstyle \pm 0.995$ &  $\hworst{-2.568} \scriptstyle \pm 1.250$ &    $\hbest{-2.114} \scriptstyle \pm 0.399$ &    $-2.442 \scriptstyle \pm 0.520$ \\
    \midrule
    \multirow{9}{*}{RMSE $\downarrow$} & boston &     $3.636 \scriptstyle \pm 0.493$ &     $3.585 \scriptstyle \pm 0.517$ &       $3.597 \scriptstyle \pm 0.684$ &   $\hbest{3.512} \scriptstyle \pm 0.573$ &     $3.756 \scriptstyle \pm 0.418$ &     $\hworst{4.593} \scriptstyle \pm 2.927$ \\
               & concrete &     $7.196 \scriptstyle \pm 0.821$ &     $7.461 \scriptstyle \pm 0.948$ &       $7.064 \scriptstyle \pm 0.921$ &   $\hbest{6.853} \scriptstyle \pm 0.796$ &     $\hworst{7.548} \scriptstyle \pm 0.865$ &     $7.367 \scriptstyle \pm 0.866$ \\
               & energy &     $2.938 \scriptstyle \pm 3.017$ &     $3.606 \scriptstyle \pm 3.927$ &       $\hbest{2.874} \scriptstyle \pm 2.254$ &   $3.364 \scriptstyle \pm 3.696$ &     $\hworst{8.614} \scriptstyle \pm 9.390$ &     $3.061 \scriptstyle \pm 2.880$ \\
               & kin8nm &     $0.080 \scriptstyle \pm 0.006$ &     $0.078 \scriptstyle \pm 0.005$ &       $\hbest{0.071} \scriptstyle \pm 0.003$ &   $\hworst{1.632} \scriptstyle \pm 4.418$ &     $0.095 \scriptstyle \pm 0.025$ &     $0.085 \scriptstyle \pm 0.007$ \\
               & naval &     $0.022 \scriptstyle \pm 0.014$ &     $0.021 \scriptstyle \pm 0.007$ &       $\hworst{0.034} \scriptstyle \pm 0.018$ &   $\hbest{0.018} \scriptstyle \pm 0.009$ &     $0.033 \scriptstyle \pm 0.041$ &     $0.020 \scriptstyle \pm 0.009$ \\
               & power &     $\hbest{4.299} \scriptstyle \pm 0.416$ &     $4.584 \scriptstyle \pm 0.356$ &       $\hworst{4.688} \scriptstyle \pm 0.335$ &   $4.369 \scriptstyle \pm 0.383$ &     $4.680 \scriptstyle \pm 0.703$ &     $4.621 \scriptstyle \pm 0.339$ \\
               & protein &     $\hworst{5.206} \scriptstyle \pm 0.780$ &     $5.101 \scriptstyle \pm 0.526$ &       $5.133 \scriptstyle \pm 0.636$ &   $\hbest{4.801} \scriptstyle \pm 0.599$ &     $5.115 \scriptstyle \pm 0.298$ &     $5.171 \scriptstyle \pm 0.632$ \\
               & wine &     $0.697 \scriptstyle \pm 0.043$ &     $0.692 \scriptstyle \pm 0.041$ &       $0.660 \scriptstyle \pm 0.040$ &   $0.673 \scriptstyle \pm 0.039$ &     $\hbest{0.632} \scriptstyle \pm 0.029$ &     $\hworst{0.731} \scriptstyle \pm 0.070$ \\
               & yacht &     $1.851 \scriptstyle \pm 0.750$ &     $1.852 \scriptstyle \pm 0.623$ &       $\hworst{2.290} \scriptstyle \pm 2.108$ &   $\hbest{1.841} \scriptstyle \pm 0.836$ &     $1.836 \scriptstyle \pm 0.712$ &     $2.214 \scriptstyle \pm 0.793$ \\
    \midrule
    \multirow{9}{*}{RCE $\downarrow$} & boston &     $0.072 \scriptstyle \pm 0.047$ &     $0.068 \scriptstyle \pm 0.038$ &       $0.126 \scriptstyle \pm 0.103$ &   $0.103 \scriptstyle \pm 0.240$ &     $\hbest{0.031} \scriptstyle \pm 0.014$ &     $\hworst{0.156} \scriptstyle \pm 0.236$ \\
               & concrete &     $0.108 \scriptstyle \pm 0.066$ &     $0.097 \scriptstyle \pm 0.034$ &       $\hworst{0.155} \scriptstyle \pm 0.072$ &   $0.057 \scriptstyle \pm 0.043$ &     $\hbest{0.030} \scriptstyle \pm 0.016$ &     $0.134 \scriptstyle \pm 0.075$ \\
               & energy &     $0.120 \scriptstyle \pm 0.149$ &     $\hbest{0.095} \scriptstyle \pm 0.067$ &       $0.117 \scriptstyle \pm 0.094$ &   $0.128 \scriptstyle \pm 0.076$ &     $\hworst{0.187} \scriptstyle \pm 0.158$ &     $0.162 \scriptstyle \pm 0.180$ \\
               & kin8nm &     $0.027 \scriptstyle \pm 0.021$ &     $0.015 \scriptstyle \pm 0.012$ &       $\hbest{0.012} \scriptstyle \pm 0.010$ &   $\hworst{0.137} \scriptstyle \pm 0.308$ &     $0.017 \scriptstyle \pm 0.011$ &     $0.024 \scriptstyle \pm 0.023$ \\
               & naval &     $0.580 \scriptstyle \pm 0.321$ &     $0.649 \scriptstyle \pm 0.293$ &       $0.719 \scriptstyle \pm 0.314$ &   $\hbest{0.499} \scriptstyle \pm 0.347$ &     $0.525 \scriptstyle \pm 0.353$ &     $\hworst{0.732} \scriptstyle \pm 0.278$ \\
               & power &     $0.027 \scriptstyle \pm 0.039$ &     $0.013 \scriptstyle \pm 0.006$ &       $\hworst{0.046} \scriptstyle \pm 0.026$ &   $\hbest{0.010} \scriptstyle \pm 0.005$ &     $0.019 \scriptstyle \pm 0.017$ &     $0.023 \scriptstyle \pm 0.017$ \\
               & protein &     $0.087 \scriptstyle \pm 0.053$ &     $0.076 \scriptstyle \pm 0.027$ &       $0.062 \scriptstyle \pm 0.030$ &   $\hbest{0.034} \scriptstyle \pm 0.021$ &     $\hworst{0.217} \scriptstyle \pm 0.387$ &     $0.079 \scriptstyle \pm 0.034$ \\
               & wine &     $0.076 \scriptstyle \pm 0.057$ &     $\hworst{0.134} \scriptstyle \pm 0.096$ &       $0.047 \scriptstyle \pm 0.068$ &   $0.033 \scriptstyle \pm 0.017$ &     $\hbest{0.023} \scriptstyle \pm 0.009$ &     $0.114 \scriptstyle \pm 0.094$ \\
               & yacht &     $0.085 \scriptstyle \pm 0.036$ &     $\hbest{0.075} \scriptstyle \pm 0.024$ &       $0.137 \scriptstyle \pm 0.173$ &   $\hworst{0.249} \scriptstyle \pm 0.312$ &     $0.102 \scriptstyle \pm 0.052$ &     $0.077 \scriptstyle \pm 0.045$ \\
    \midrule
    \multirow{9}{*}{TCE $\downarrow$} & boston &     $0.138 \scriptstyle \pm 0.092$ &     $0.132 \scriptstyle \pm 0.065$ &       $0.221 \scriptstyle \pm 0.154$ &   $0.134 \scriptstyle \pm 0.235$ &     $\hbest{0.037} \scriptstyle \pm 0.023$ &     $\hworst{0.234} \scriptstyle \pm 0.231$ \\
               & concrete &     $0.212 \scriptstyle \pm 0.094$ &     $0.199 \scriptstyle \pm 0.055$ &       $\hworst{0.280} \scriptstyle \pm 0.098$ &   $0.107 \scriptstyle \pm 0.088$ &     $\hbest{0.052} \scriptstyle \pm 0.040$ &     $0.248 \scriptstyle \pm 0.107$ \\
               & energy &     $0.175 \scriptstyle \pm 0.211$ &     $\hbest{0.161} \scriptstyle \pm 0.132$ &       $0.180 \scriptstyle \pm 0.149$ &   $0.216 \scriptstyle \pm 0.137$ &     $\hworst{0.267} \scriptstyle \pm 0.208$ &     $0.227 \scriptstyle \pm 0.238$ \\
               & kin8nm &     $0.064 \scriptstyle \pm 0.051$ &     $0.030 \scriptstyle \pm 0.031$ &       $\hbest{0.025} \scriptstyle \pm 0.030$ &   $\hworst{0.150} \scriptstyle \pm 0.303$ &     $0.029 \scriptstyle \pm 0.027$ &     $0.048 \scriptstyle \pm 0.051$ \\
               & naval &     $0.650 \scriptstyle \pm 0.284$ &     $0.714 \scriptstyle \pm 0.229$ &       $0.744 \scriptstyle \pm 0.296$ &   $\hbest{0.560} \scriptstyle \pm 0.334$ &     $0.585 \scriptstyle \pm 0.336$ &     $\hworst{0.763} \scriptstyle \pm 0.253$ \\
               & power &     $0.055 \scriptstyle \pm 0.088$ &     $0.033 \scriptstyle \pm 0.017$ &       $\hworst{0.109} \scriptstyle \pm 0.049$ &   $\hbest{0.020} \scriptstyle \pm 0.015$ &     $0.034 \scriptstyle \pm 0.039$ &     $0.057 \scriptstyle \pm 0.040$ \\
               & protein &     $0.167 \scriptstyle \pm 0.088$ &     $0.157 \scriptstyle \pm 0.052$ &       $0.129 \scriptstyle \pm 0.057$ &   $\hbest{0.071} \scriptstyle \pm 0.050$ &     $\hworst{0.240} \scriptstyle \pm 0.375$ &     $0.159 \scriptstyle \pm 0.064$ \\
               & wine &     $0.153 \scriptstyle \pm 0.104$ &     $\hworst{0.233} \scriptstyle \pm 0.135$ &       $0.084 \scriptstyle \pm 0.115$ &   $0.073 \scriptstyle \pm 0.046$ &     $\hbest{0.019} \scriptstyle \pm 0.007$ &     $0.209 \scriptstyle \pm 0.146$ \\
               & yacht &     $0.133 \scriptstyle \pm 0.058$ &     $\hbest{0.095} \scriptstyle \pm 0.073$ &       $0.097 \scriptstyle \pm 0.066$ &   $\hworst{0.289} \scriptstyle \pm 0.322$ &     $0.136 \scriptstyle \pm 0.107$ &     $0.098 \scriptstyle \pm 0.056$ \\
    \midrule
    \multirow{9}{*}{batch time $\downarrow$} & boston &     $0.029 \scriptstyle \pm 0.018$ &     $0.037 \scriptstyle \pm 0.020$ &       $0.032 \scriptstyle \pm 0.005$ &   $0.044 \scriptstyle \pm 0.016$ &     $\hworst{0.047} \scriptstyle \pm 0.030$ &     $\hbest{0.003} \scriptstyle \pm 0.001$ \\
               & concrete &     $0.033 \scriptstyle \pm 0.018$ &     $0.016 \scriptstyle \pm 0.010$ &       $0.022 \scriptstyle \pm 0.006$ &   $0.043 \scriptstyle \pm 0.030$ &     $\hworst{0.090} \scriptstyle \pm 0.078$ &     $\hbest{0.003} \scriptstyle \pm 0.003$ \\
               & energy &     $0.021 \scriptstyle \pm 0.021$ &     $0.023 \scriptstyle \pm 0.014$ &       $0.029 \scriptstyle \pm 0.005$ &   $0.025 \scriptstyle \pm 0.006$ &     $\hworst{0.117} \scriptstyle \pm 0.099$ &     $\hbest{0.002} \scriptstyle \pm 0.000$ \\
               & kin8nm &     $0.012 \scriptstyle \pm 0.007$ &     $0.008 \scriptstyle \pm 0.004$ &       $0.019 \scriptstyle \pm 0.007$ &   $0.040 \scriptstyle \pm 0.034$ &     $\hworst{0.244} \scriptstyle \pm 0.157$ &     $\hbest{0.003} \scriptstyle \pm 0.002$ \\
               & naval &     $0.013 \scriptstyle \pm 0.006$ &     $0.010 \scriptstyle \pm 0.005$ &       $0.024 \scriptstyle \pm 0.011$ &   $0.040 \scriptstyle \pm 0.018$ &     $\hworst{0.203} \scriptstyle \pm 0.196$ &     $\hbest{0.003} \scriptstyle \pm 0.002$ \\
               & power &     $0.013 \scriptstyle \pm 0.002$ &     $0.018 \scriptstyle \pm 0.005$ &       $0.023 \scriptstyle \pm 0.007$ &   $0.073 \scriptstyle \pm 0.037$ &     $\hworst{0.215} \scriptstyle \pm 0.125$ &     $\hbest{0.005} \scriptstyle \pm 0.003$ \\
               & protein &     $0.023 \scriptstyle \pm 0.006$ &     $0.016 \scriptstyle \pm 0.005$ &       $0.027 \scriptstyle \pm 0.007$ &   $0.057 \scriptstyle \pm 0.030$ &     $\hworst{0.258} \scriptstyle \pm 0.135$ &     $\hbest{0.005} \scriptstyle \pm 0.003$ \\
               & wine &     $0.013 \scriptstyle \pm 0.007$ &     $0.008 \scriptstyle \pm 0.002$ &       $0.018 \scriptstyle \pm 0.008$ &   $0.065 \scriptstyle \pm 0.051$ &     $\hworst{0.102} \scriptstyle \pm 0.094$ &     $\hbest{0.005} \scriptstyle \pm 0.004$ \\
               & yacht &     $0.003 \scriptstyle \pm 0.002$ &     $0.004 \scriptstyle \pm 0.002$ &       $0.015 \scriptstyle \pm 0.003$ &   $0.015 \scriptstyle \pm 0.004$ &     $\hworst{0.021} \scriptstyle \pm 0.010$ &     $\hbest{0.001} \scriptstyle \pm 0.000$ \\
    \bottomrule
    \end{tabular}}
\end{table}

\FloatBarrier
\subsection{Image Classification}\label{app:additional_image_results}

\Cref{tab:detailed_img_results} shows a detailed breakdown of the performance of \glspl{DUN}, as well as various benchmark methods, on image datasets.

\begin{table}[h]
    \centering
    \caption{Mean values and standard deviations for results on image datasets. Bold blue text denotes the best mean value for each dataset and each metric. Bold red text denotes the worst mean value. Note that in some cases the best/worst mean values are within error of other mean values.}
    \begin{tabular}{ll|ccccc}
        \toprule
             \textsc{Dataset} & \textsc{Method}  &   \textsc{LL} $\uparrow$ &  \textsc{Error} $\downarrow$&                  \textsc{Brier} $\downarrow$ & \textsc{ECE} $\downarrow$\\
        \midrule
        \multirow{8}{*}{CIFAR10} & DUN &  $\hworst{-0.240} \scriptstyle \pm 0.011$ &  $0.056 \scriptstyle \pm 0.002$ &  $0.092 \scriptstyle \pm 0.003$ &  $\hworst{0.034} \scriptstyle \pm 0.002$ \\
             & Depth-Ens (13) &  $\hbest{-0.143} \scriptstyle \pm 0.003$ &  $0.044 \scriptstyle \pm 0.000$ &  $0.068 \scriptstyle \pm 0.001$ &  $\hbest{0.006} \scriptstyle \pm 0.002$ \\
             & Depth-Ens (5) &  $-0.212 \scriptstyle \pm 0.003$ &  $\hworst{0.064} \scriptstyle \pm 0.001$ &  $\hworst{0.096} \scriptstyle \pm 0.001$ &  $0.010 \scriptstyle \pm 0.001$ \\
             & Dropout &  $-0.211 \scriptstyle \pm 0.004$ &  $0.051 \scriptstyle \pm 0.002$ &  $0.081 \scriptstyle \pm 0.002$ &  $0.028 \scriptstyle \pm 0.002$ \\
             & Dropout ($p = 0.3$) &  $-0.222 \scriptstyle \pm 0.006$ &  $0.053 \scriptstyle \pm 0.001$ &  $0.084 \scriptstyle \pm 0.001$ &  $0.027 \scriptstyle \pm 0.002$ \\
             & Ensemble &  $-0.145 \scriptstyle \pm 0.002$ &  $\hbest{0.042} \scriptstyle \pm 0.001$ &  $\hbest{0.063} \scriptstyle \pm 0.001$ &  $0.010 \scriptstyle \pm 0.001$ \\
             & SGD &  $-0.234 \scriptstyle \pm 0.008$ &  $0.051 \scriptstyle \pm 0.001$ &  $0.084 \scriptstyle \pm 0.002$ &  $0.033 \scriptstyle \pm 0.001$ \\
             & S-ResNet &  $-0.233 \scriptstyle \pm 0.009$ &  $0.057 \scriptstyle \pm 0.001$ &  $0.090 \scriptstyle \pm 0.002$ &  $0.030 \scriptstyle \pm 0.003$ \\
        \midrule
        \multirow{8}{*}{CIFAR100} & DUN &  $\hworst{-1.182} \scriptstyle \pm 0.018$ &  $0.246 \scriptstyle \pm 0.001$ &  $\hworst{0.377} \scriptstyle \pm 0.001$ &  $\hworst{0.135} \scriptstyle \pm 0.001$ \\
             & Depth-Ens (13) &  $\hbest{-0.796} \scriptstyle \pm 0.003$ &  $\hbest{0.210} \scriptstyle \pm 0.001$ &  $0.297 \scriptstyle \pm 0.001$ &  $\hbest{0.015} \scriptstyle \pm 0.002$ \\
             & Depth-Ens (5) &  $-1.046 \scriptstyle \pm 0.011$ &  $\hworst{0.264} \scriptstyle \pm 0.002$ &  $0.365 \scriptstyle \pm 0.002$ &  $0.038 \scriptstyle \pm 0.001$ \\
             & Dropout &  $-1.057 \scriptstyle \pm 0.020$ &  $0.233 \scriptstyle \pm 0.003$ &  $0.347 \scriptstyle \pm 0.004$ &  $0.110 \scriptstyle \pm 0.002$ \\
             & Dropout ($p = 0.3$) &  $-1.088 \scriptstyle \pm 0.007$ &  $0.243 \scriptstyle \pm 0.002$ &  $0.358 \scriptstyle \pm 0.003$ &  $0.111 \scriptstyle \pm 0.002$ \\
             & Ensemble &  $-0.821 \scriptstyle \pm 0.006$ &  $0.204 \scriptstyle \pm 0.001$ &  $\hbest{0.292} \scriptstyle \pm 0.001$ &  $0.047 \scriptstyle \pm 0.002$ \\
             & SGD &  $-1.155 \scriptstyle \pm 0.014$ &  $0.234 \scriptstyle \pm 0.002$ &  $0.363 \scriptstyle \pm 0.004$ &  $0.133 \scriptstyle \pm 0.002$ \\
             & S-ResNet &  $-1.108 \scriptstyle \pm 0.034$ &  $0.248 \scriptstyle \pm 0.004$ &  $0.368 \scriptstyle \pm 0.008$ &  $0.113 \scriptstyle \pm 0.008$ \\
        \midrule
        \multirow{8}{*}{Fashion} & DUN &  $-0.245 \scriptstyle \pm 0.029$ &  $0.051 \scriptstyle \pm 0.001$ &  $\hworst{0.087} \scriptstyle \pm 0.002$ &  $0.035 \scriptstyle \pm 0.002$ \\
             & Depth-Ens (13) &  $\hbest{-0.141} \scriptstyle \pm 0.002$ &  $\hbest{0.041} \scriptstyle \pm 0.001$ &  $\hbest{0.064} \scriptstyle \pm 0.000$ &  $\hbest{0.011} \scriptstyle \pm 0.001$ \\
             & Depth-Ens (5) &  $-0.152 \scriptstyle \pm 0.002$ &  $0.044 \scriptstyle \pm 0.001$ &  $0.069 \scriptstyle \pm 0.001$ &  $0.014 \scriptstyle \pm 0.001$ \\
             & Dropout &  $-0.208 \scriptstyle \pm 0.006$ &  $0.050 \scriptstyle \pm 0.001$ &  $0.081 \scriptstyle \pm 0.001$ &  $\hworst{0.050} \scriptstyle \pm 0.000$ \\
             & Dropout ($p = 0.3$) &  $-0.185 \scriptstyle \pm 0.004$ &  $0.049 \scriptstyle \pm 0.001$ &  $0.079 \scriptstyle \pm 0.002$ &  $\hworst{0.050} \scriptstyle \pm 0.000$ \\
             & Ensemble &  $-0.180 \scriptstyle \pm 0.006$ &  $0.044 \scriptstyle \pm 0.001$ &  $0.070 \scriptstyle \pm 0.002$ &  $\hworst{0.050} \scriptstyle \pm 0.000$ \\
             & SGD &  $\hworst{-0.272} \scriptstyle \pm 0.006$ &  $0.051 \scriptstyle \pm 0.001$ &  $\hworst{0.087} \scriptstyle \pm 0.002$ &  $0.048 \scriptstyle \pm 0.000$ \\
             & S-ResNet &  $-0.217 \scriptstyle \pm 0.025$ &  $\hworst{0.052} \scriptstyle \pm 0.003$ &  $0.084 \scriptstyle \pm 0.007$ &  $0.025 \scriptstyle \pm 0.006$ \\
        \midrule
        \multirow{8}{*}{MNIST} & DUN &  $-0.015 \scriptstyle \pm 0.004$ &  $0.004 \scriptstyle \pm 0.000$ &  $0.006 \scriptstyle \pm 0.000$ &  $0.005 \scriptstyle \pm 0.003$ \\
             & Depth-Ens (13) &  $\hbest{-0.009} \scriptstyle \pm 0.000$ &  $\hbest{0.003} \scriptstyle \pm 0.000$ &  $\hbest{0.004} \scriptstyle \pm 0.000$ &  $\hbest{0.001} \scriptstyle \pm 0.000$ \\
             & Depth-Ens (5) &  $\hbest{-0.009} \scriptstyle \pm 0.000$ &  $\hbest{0.003} \scriptstyle \pm 0.000$ &  $0.005 \scriptstyle \pm 0.000$ &  $\hbest{0.001} \scriptstyle \pm 0.000$ \\
             & Dropout &  $-0.011 \scriptstyle \pm 0.000$ &  $0.004 \scriptstyle \pm 0.000$ &  $0.005 \scriptstyle \pm 0.000$ &  $0.040 \scriptstyle \pm 0.022$ \\
             & Dropout ($p = 0.3$) &  $-0.010 \scriptstyle \pm 0.001$ &  $\hbest{0.003} \scriptstyle \pm 0.000$ &  $0.005 \scriptstyle \pm 0.000$ &  $0.030 \scriptstyle \pm 0.027$ \\
             & Ensemble &  $-0.010 \scriptstyle \pm 0.000$ &  $\hbest{0.003} \scriptstyle \pm 0.000$ &  $0.005 \scriptstyle \pm 0.000$ &  $\hworst{0.050} \scriptstyle \pm 0.000$ \\
             & SGD &  $-0.012 \scriptstyle \pm 0.001$ &  $0.004 \scriptstyle \pm 0.000$ &  $0.006 \scriptstyle \pm 0.000$ &  $\hworst{0.050} \scriptstyle \pm 0.000$ \\
             & S-ResNet &  $\hworst{-0.031} \scriptstyle \pm 0.008$ &  $\hworst{0.005} \scriptstyle \pm 0.004$ &  $\hworst{0.011} \scriptstyle \pm 0.005$ &  $0.015 \scriptstyle \pm 0.006$ \\
        \midrule
        \multirow{8}{*}{SVHN} & DUN &  $\hworst{-0.202} \scriptstyle \pm 0.021$ &  $\hworst{0.046} \scriptstyle \pm 0.005$ &  $\hworst{0.074} \scriptstyle \pm 0.007$ &  $0.019 \scriptstyle \pm 0.004$ \\
             & Depth-Ens (13) &  $\hbest{-0.114} \scriptstyle \pm 0.001$ &  $\hbest{0.027} \scriptstyle \pm 0.000$ &  $\hbest{0.042} \scriptstyle \pm 0.000$ &  $\hbest{0.005} \scriptstyle \pm 0.000$ \\
             & Depth-Ens (5) &  $-0.129 \scriptstyle \pm 0.001$ &  $0.030 \scriptstyle \pm 0.000$ &  $0.047 \scriptstyle \pm 0.000$ &  $0.006 \scriptstyle \pm 0.000$ \\
             & Dropout &  $-0.162 \scriptstyle \pm 0.014$ &  $0.036 \scriptstyle \pm 0.008$ &  $0.057 \scriptstyle \pm 0.011$ &  $\hworst{0.050} \scriptstyle \pm 0.000$ \\
             & Dropout ($p = 0.3$) &  $-0.138 \scriptstyle \pm 0.002$ &  $0.031 \scriptstyle \pm 0.001$ &  $0.049 \scriptstyle \pm 0.001$ &  $\hworst{0.050} \scriptstyle \pm 0.000$ \\
             & Ensemble &  $-0.123 \scriptstyle \pm 0.002$ &  $\hbest{0.027} \scriptstyle \pm 0.000$ &  $0.043 \scriptstyle \pm 0.000$ &  $\hworst{0.050} \scriptstyle \pm 0.000$ \\
             & SGD &  $-0.177 \scriptstyle \pm 0.003$ &  $0.033 \scriptstyle \pm 0.001$ &  $0.055 \scriptstyle \pm 0.001$ &  $0.049 \scriptstyle \pm 0.000$ \\
             & S-ResNet &  $-0.168 \scriptstyle \pm 0.005$ &  $0.035 \scriptstyle \pm 0.001$ &  $0.056 \scriptstyle \pm 0.001$ &  $0.018 \scriptstyle \pm 0.001$ \\
        \bottomrule
    \end{tabular}
    \label{tab:detailed_img_results}
\end{table}

\begin{figure}[h]
    \centering
    \includegraphics{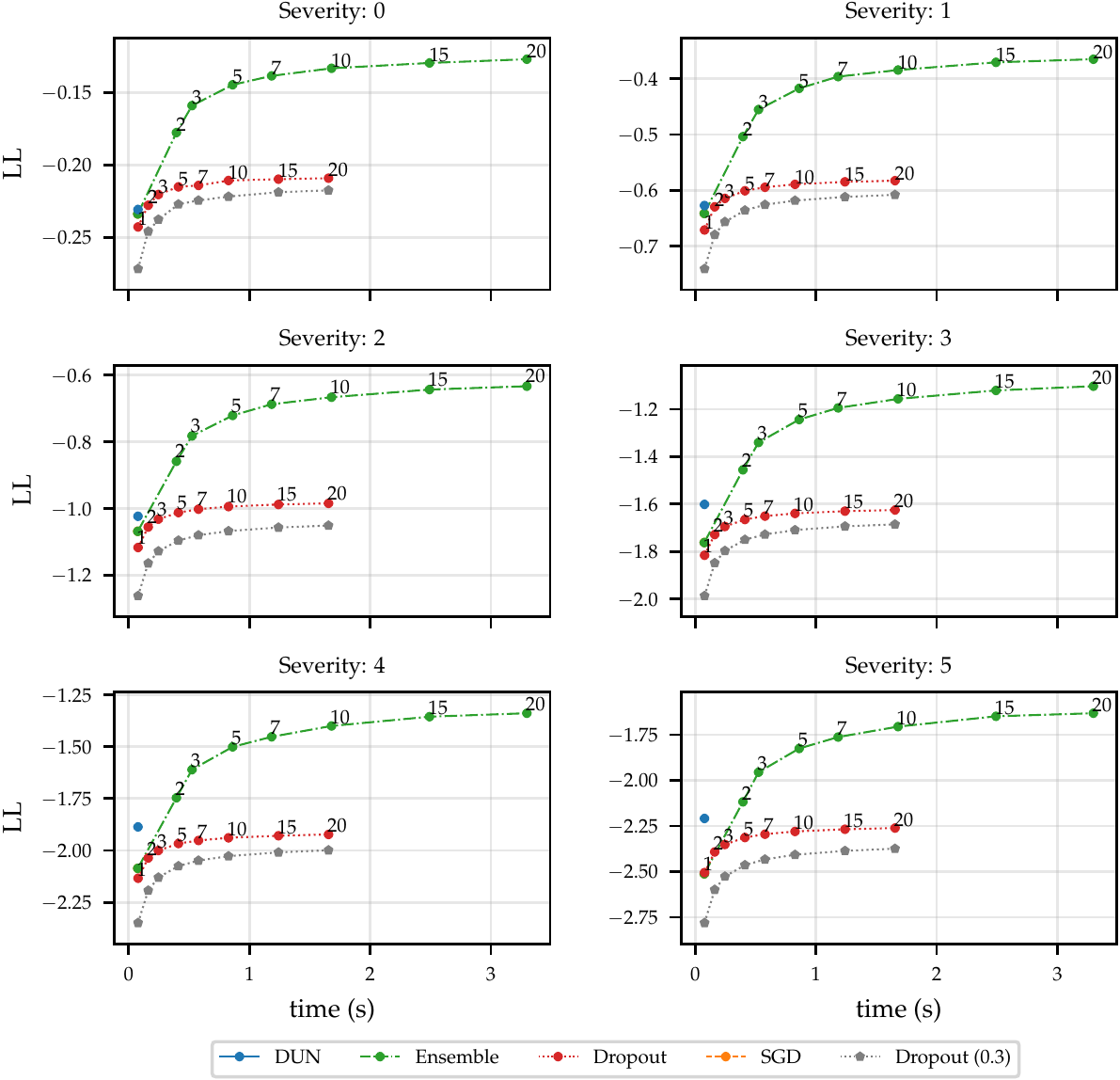}
    \caption{Pareto frontiers showing \gls{LL} for all CIFAR10 corruptions vs batch prediction time. Batch size is 256, split over 2 Nvidia P100 GPUs. Annotations show ensemble elements and Dropout samples. Note that a single element ensemble is equivalent to SGD.}
    \label{fig:app_corruption_pareto}
\end{figure}

\begin{figure}[h]
    \centering
    \includegraphics{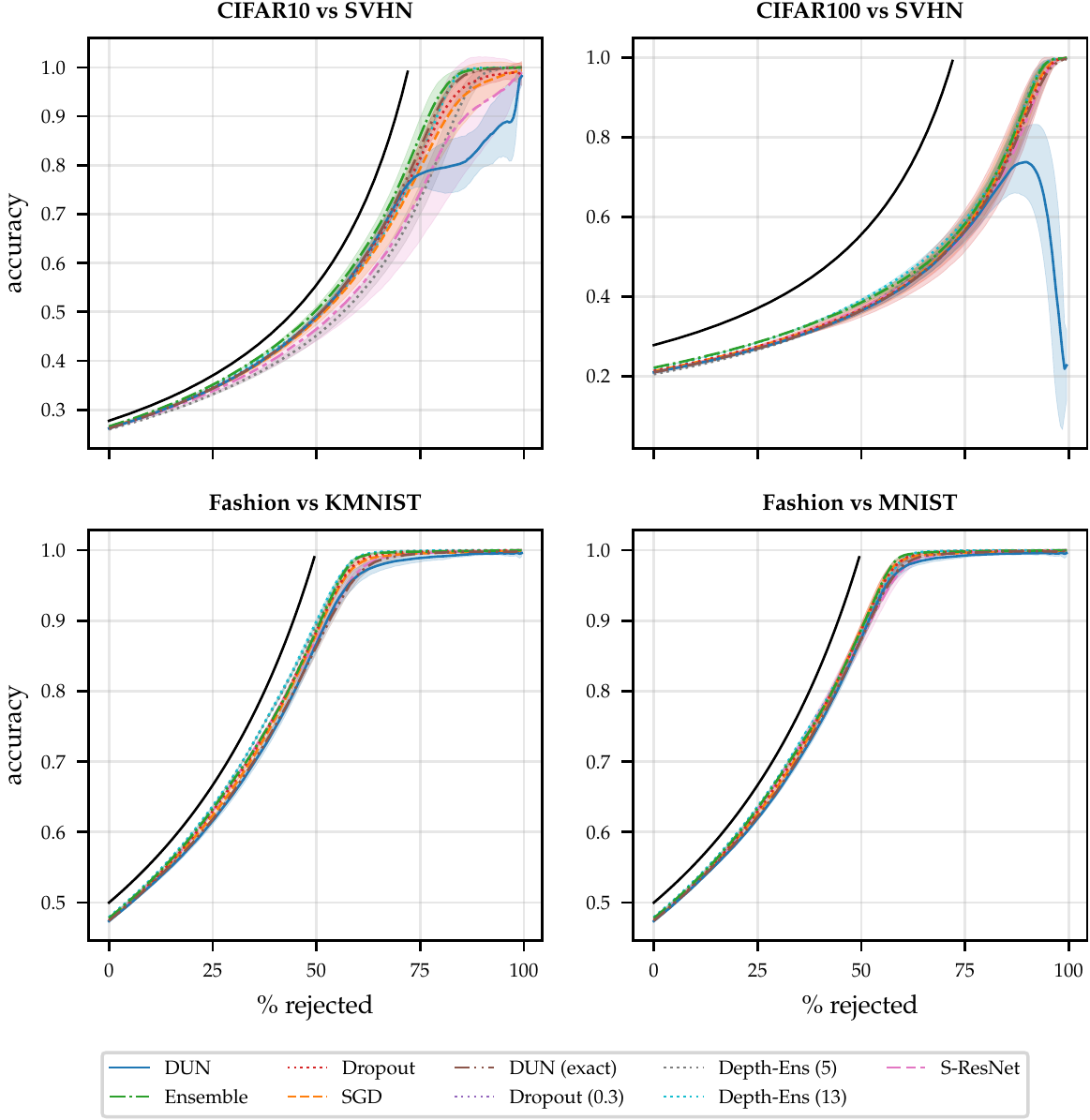}
    \caption{Rejection-classification plots. The black line denotes the theoretical maximum performance; all in-distribution samples are correctly classified and OOD samples are rejected first.}
    \label{fig:app_rejection_classification}
\end{figure}

\Cref{fig:app_corruption_pareto} compares methods' \gls{LL} performance vs batch time on increasingly corrupted CIFAR10 test data.
DUNs are competitive in all cases but their relative performance increases with corruption severity. Dropout shows a clear drop in \gls{LL} when using a drop rate of $0.3$.
\Cref{fig:app_rejection_classification} shows rejection classification plots for CIFAR10 and CIFAR100 vs SVHN and for Fashion MNIST vs MNIST and KMNIST. \Cref{tab:AUC_ROC} shows AUC-ROC values for entropy based in-distribution vs \gls{OOD} classification with all methods under consideration.

\begin{table}[h]
    \centering
     \caption{AUC-ROC values obtained for predictive entropy based separation of in and out of distribution test sets. Bold blue text denotes the best mean value for each dataset and each metric. Bold red text denotes the worst mean value. Note that in some cases the best/worst mean values are within error of other mean values.}
    \label{tab:AUC_ROC}
    \begin{tabular}{l|llllll}
\toprule
\textsc{Source} &                       \textsc{CIFAR10} &                      \textsc{CIFAR100} & \multicolumn{2}{c}{\textsc{Fashion}} &                         \textsc{MNIST} &                          \textsc{SVHN} \\
\textsc{Target} &                          \multicolumn{2}{c}{\textsc{SVHN}} &                                 \textsc{KMNIST} &                         \textsc{MNIST} &                       \textsc{Fashion} &                       \textsc{CIFAR10} \\
\midrule
DUN            &  $\hworst{0.84} \scriptstyle \pm 0.06$  &  $0.76 \scriptstyle \pm 0.04$             &  $0.95 \scriptstyle \pm 0.01$            &  $\hworst{0.95} \scriptstyle \pm 0.01$    &  $\hworst{0.86} \scriptstyle \pm 0.03$    &  $0.92 \scriptstyle \pm 0.02$ \\
DUN (exact)    &  $0.90 \scriptstyle \pm 0.03$          &  $0.77 \scriptstyle \pm 0.03$             &  $\hworst{0.94} \scriptstyle \pm 0.01$    &  $\hworst{0.95} \scriptstyle \pm 0.01$    &  $0.87 \scriptstyle \pm 0.03$             &  $\hworst{0.90} \scriptstyle \pm 0.03$ \\
Depth-Ens (13) &  $0.91 \scriptstyle \pm 0.00$          &  $\hbest{0.80} \scriptstyle \pm 0.01$     &  $\hbest{0.97} \scriptstyle \pm 0.00$     &  $0.96 \scriptstyle \pm 0.00$            &  $\hbest{0.99} \scriptstyle \pm 0.00$     &  $\hbest{0.98} \scriptstyle \pm 0.00$ \\
Depth-Ens (5)  &  $\hworst{0.84} \scriptstyle \pm 0.02$ &  $0.79 \scriptstyle \pm 0.02$             &  $\hbest{0.97} \scriptstyle \pm 0.00$     &  $0.96 \scriptstyle \pm 0.00$            &  $0.94 \scriptstyle \pm 0.04$             &  $0.97 \scriptstyle \pm 0.00$ \\
Dropout        &  $0.90 \scriptstyle \pm 0.01$          &  $\hworst{0.75} \scriptstyle \pm 0.05$    &  $0.96 \scriptstyle \pm 0.01$             &  $\hbest{0.97} \scriptstyle \pm 0.01$    &  $0.95 \scriptstyle \pm 0.04$             &  $0.94 \scriptstyle \pm 0.01$ \\
Dropout (0.3)  &  $0.88 \scriptstyle \pm 0.01$          &  $0.76 \scriptstyle \pm 0.03$             &  $0.96 \scriptstyle \pm 0.01$             &  $0.96 \scriptstyle \pm 0.01$            &  $0.91 \scriptstyle \pm 0.06$             &  $0.95 \scriptstyle \pm 0.01$ \\
Ensemble       &  $\hbest{0.93} \scriptstyle \pm 0.02$  &  $0.77 \scriptstyle \pm 0.01$             &  $0.96 \scriptstyle \pm 0.00$             &  $0.96 \scriptstyle \pm 0.00$            &  $0.98 \scriptstyle \pm 0.00$             &  $0.97 \scriptstyle \pm 0.00$ \\
S-ResNet       &  $0.87 \scriptstyle \pm 0.05$          &  $0.79 \scriptstyle \pm 0.03$             &  $0.96 \scriptstyle \pm 0.01$             &  $0.96 \scriptstyle \pm 0.01$            &  $0.86 \scriptstyle \pm 0.04$             &  $0.93 \scriptstyle \pm 0.01$ \\
SGD            &  $0.89 \scriptstyle \pm 0.02$          &  $0.76 \scriptstyle \pm 0.03$             &  $0.95 \scriptstyle \pm 0.01$             &  $0.96 \scriptstyle \pm 0.01$            &  $0.94 \scriptstyle \pm 0.04$             &  $0.93 \scriptstyle \pm 0.01$ \\
\bottomrule
\end{tabular}
\end{table}

\begin{figure}[h]
    \centering
    \includegraphics[width=0.5\linewidth]{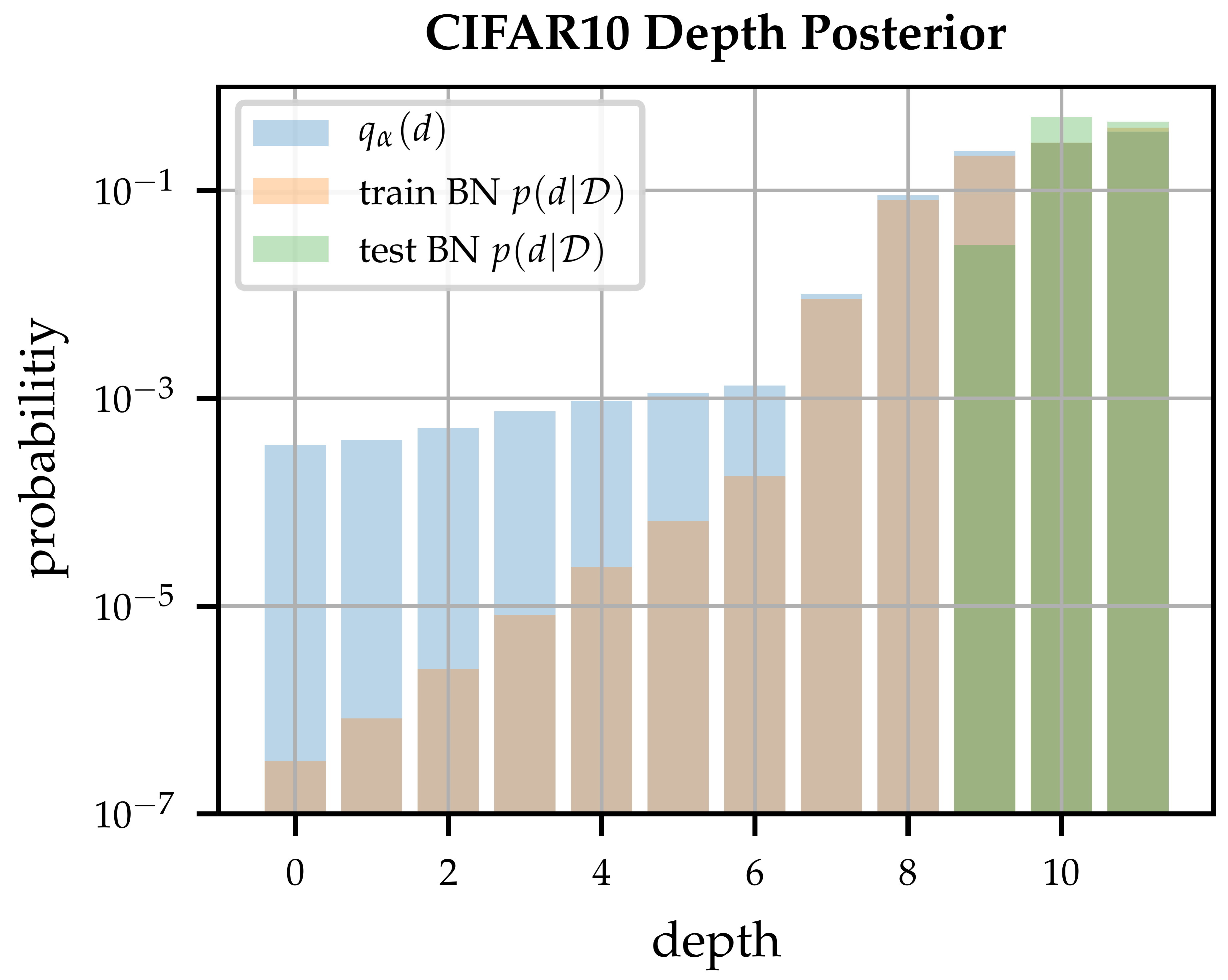}
    \caption{Posterior probabilities produced by VI, exact inference with batch-norm in train mode and exact inference with batch-norm in test mode on the CIFAR10 train-set.}
    \label{fig:rebuttal_BN_probs}
\end{figure}

In some cases, similarly to \cref{sec:experiments}, we find that using the exact posterior in DUNs is necessary to reduce underconfidence in-distribution. We further investigate this by plotting the posterior probabilities produced by VI, exact inference with batch-norm in train mode and exact inference with batch-norm in test mode in \cref{fig:rebuttal_BN_probs}. All three approaches assign most probability mass to the final three layers. However, exact inference with BN in test mode differs in that it assigns vanishing low probability mass to all other layers. This is in contrast to VI and train mode BN, where the probability mass assigned to each shallower layer decreases gradually. Predictions are made with BN in test mode. This changes the BN layers from using batch statistics for normalisation to using weighed moving averages computed from the train set. \Cref{fig:rebuttal_BN_probs} suggests that this shift in network behavior results in predictions from earlier layers being worse and weighing them too heavily in our predictive posterior results in underconfidence.

\begin{figure}[h]
    \centering
    \includegraphics[width=0.5\linewidth]{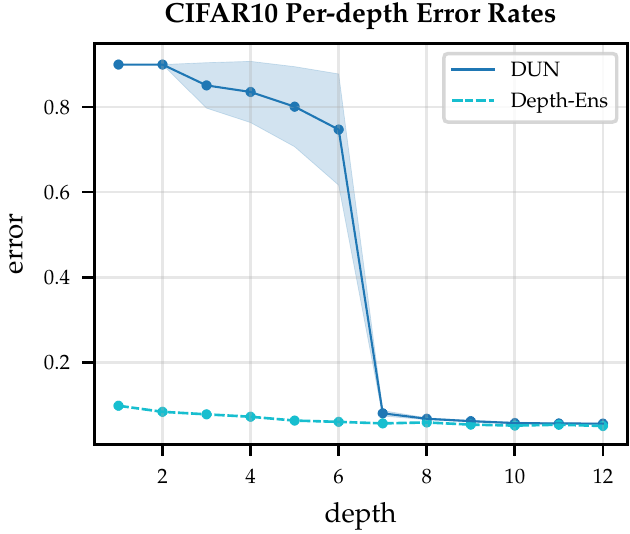}
    \caption{Comparison of per-depth error rates for the layers of a \gls{DUN} and depth-ensemble elements of the same depth.}
    \label{fig:per_d_err}
\end{figure}

In \cref{sec:prob_model}, \cref{sec:VI_vs_MLE} and \cref{app:vI_vs_MLE}, we discussed how \glspl{DUN} trade off expressively and explanation diversity automatically. \Cref{fig:per_d_err} shows confirms that this mechanism is due to earlier layers obtaining low accuracy. These layers perform representation learning instead. In turn, these layers are assigned low posterior probabilities such that they contribute negligibly to predictions.

\FloatBarrier
\section{DUNs for Neural Architecture Search} \label{app:duns_for_nas}

In this section, we briefly explore the application of \glspl{DUN} to \gls{NAS} and how architecture hyperparameters affect the posterior over depth. 
This section is based on the previous work \citep{antoran2020variational}. Please see that paper for more information, including further experimental evaluation and analysis as well as contextualisation of this technique in the \gls{NAS} literature.  

After training a \gls{DUN}, as described in \cref{sec:inference}, $q_{\balpha}(d{=}i)\,{=}\,\alpha_{i}$ represents our confidence that the number of blocks we should use is $i$. We would like to use this information to prune our network such that we reduce computational cost while maintaining performance. Recall our training objective \cref{eq:var_objective}:
\begin{gather*}
    \mathcal{L}(\balpha, \btheta) = \textstyle{\sum}_{n=1}^{N} \EX_{q_{\balpha}(d)}\left[ \log p(\bff{y}^{(n)}|\bff{x}^{(n)}, d; \btheta)\right] - \text{KL}( q_{\balpha}(d)\,\|\,p_{\bbeta}(d)).
\end{gather*}
In low data regimes, where both the log-likelihood and KL divergence terms are of comparable scale, we obtain a posterior with a clear maximum. We choose 
\begin{gather} \label{eqn:d_opt_exact}
    d_{\text{opt}}{=}\argmax_{i}\alpha_{i}.
\end{gather} 
as our fixed depth. In medium-to-big data regimes, where the log-likelihood dominates our objective, we find that the values of $\alpha_{i}$ flatten out after reaching an appropriate depth. For examples of this phenomenon, compare the approximate posteriors over depth shown in \cref{fig:spiral_depth_barplot} and \cref{fig:image_depth_dist}. We propose a heuristics for choosing $d_\mathrm{opt}$ in this case. 
We choose the smallest depth with a probability larger that 95\% of the maximum of $q$:
\begin{gather} \label{eqn:d_opt_heur95}
    d_\mathrm{opt} = \min_{i}\{i : q(d{=}i) \geq 0.95 \max_{j} q(d{=}j)\}.
\end{gather}
Both heuristics aim to keep the minimum number of blocks needed to explain the data well. We prune all blocks after $d_{opt}$ by setting $q_{\balpha}(d{=}d_{\text{opt}})\,{=}\, q_{\balpha}(d{\geq}d_{\text{opt}})$ and then $q_{\balpha}(d{>}d_{\text{opt}})\,{=}\,0$. Instead of also discarding the learnt probabilities over shallower networks, we incorporate them when making predictions on new data points $\bff{x}^{*}$ through marginalisation:
\begin{gather}\label{eq:predict_cutoff_depth}
    p(\bff{y}^{*} | \bff{x}^{*}) \approx \sum^{d_{\text{opt}}}_{i=0} p(\bff{y}^{*} | \bff{x}^{*}, d{=}i; \btheta) q_{\balpha}(d{=}i).
\end{gather}
We refer to pruned \glspl{DUN} as \glspl{LDN} and study them, contrasting them with (standard) \glspl{DDN} in the following experiments.

\subsection{Toy Experiments}

\begin{figure}[htbp]
\begin{center}
\centerline{\includegraphics[width=0.95\textwidth]{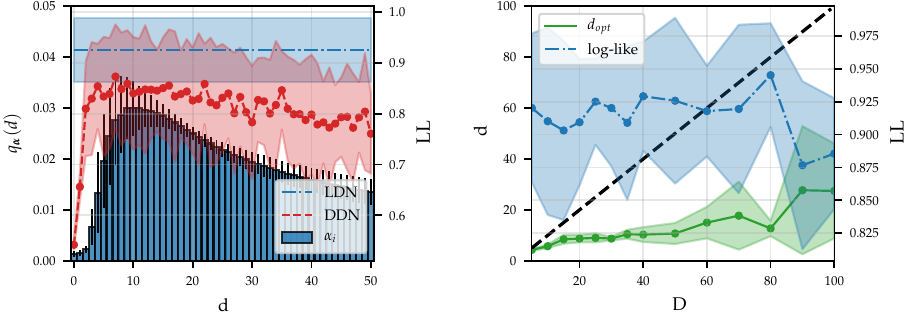}}
\vskip -0.1in
\caption{Left: posterior over depths for a \gls{LDN} of $D\,{=}\,50$ trained on our spirals dataset. Test log-likelihood values obtained for \glspl{DDN} at every depth are overlaid with the log-likelihood value obtained with a \gls{LDN} when marginalising over $d_{opt}\,{=}\,9$ layers. Right: the \gls{LDN}'s depth, chosen using \cref{eqn:d_opt_exact}, and test performance remain stable as $D$ increases up until $D\,{\approx}\,50$.}
\label{fig:spiral_depth_barplot}
\end{center}
\vskip -0.2in
\end{figure}

We generate a 2d training set by drawing 200 samples from a 720{\degree} rotation 2-armed spiral function with additive Gaussian noise of $\sigma\,{=}\,0.15$. The test set is composed of an additional 1800 samples. Choosing a relatively small width for each hidden layer $w\,{=}\,20$ to ensure the task can not be solved with a shallow model, we train fully-connected \glspl{LDN} with varying maximum depths $D$ and \glspl{DDN} of all depths up to $D{=}100$. \cref{fig:spiral_depth_barplot} shows how the depths to which \glspl{LDN} assign larger probabilities match those at which \glspl{DDN} perform best. Predictions from \glspl{LDN} pruned to $d_{opt}$ layers outperform \glspl{DDN} at all depths. The chosen $d_{opt}$ remains stable for increasing maximum depths up to $D\approx50$. The same is true for test performance, showing some robustness to overfitting. After this point, training starts to become unstable. We repeat experiments 6 times and report standard deviations as error bars. 

\begin{figure}[htb]
\begin{center}
\centerline{\includegraphics[width=0.65\textwidth]{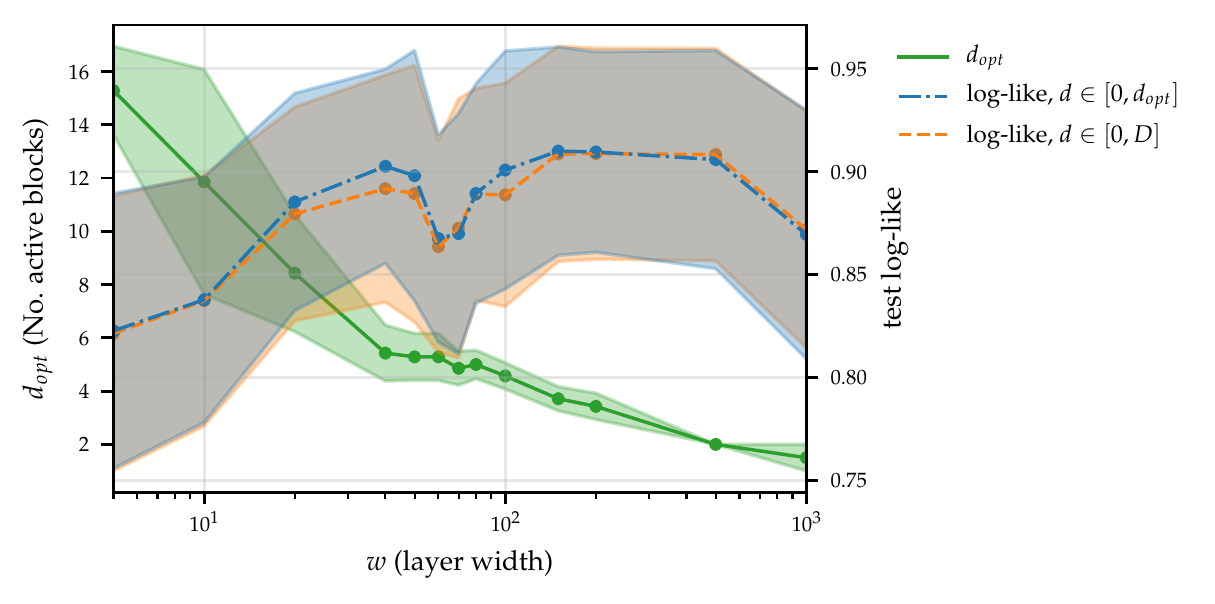}}
\vskip -0.1in
\caption[]{Evolution of LDNs' chosen depth and test performance as their layer width $w$ increases. The results obtained when making predictions by marginalising over all $D\,{=}\,20$ layers overlap with those obtained when only using the first $d_{opt}$ layers. The x-axis is presented in logarithmic scale.}
\label{fig:spiral_width_increase}
\end{center}
\vskip -0.2in
\end{figure}

We further explore the properties of LDNs in the small data regime by varying the layer width $w$. As shown in \cref{fig:spiral_width_increase}, very small widths result in very deep LDNs and worse test performance. Increasing layer width gives our models more representation capacity at each layer, causing the optimal depth to decrease rapidly. Test performance remains stable for widths in the range of $20$ to $500$, showing that our approach adapts well to changes in this parameter. The test log-likelihood starts to decrease for widths of $1000$, possibly due to training instabilities.

\begin{figure}[htb]
\begin{center}
\centerline{\includegraphics[width=0.9\textwidth]{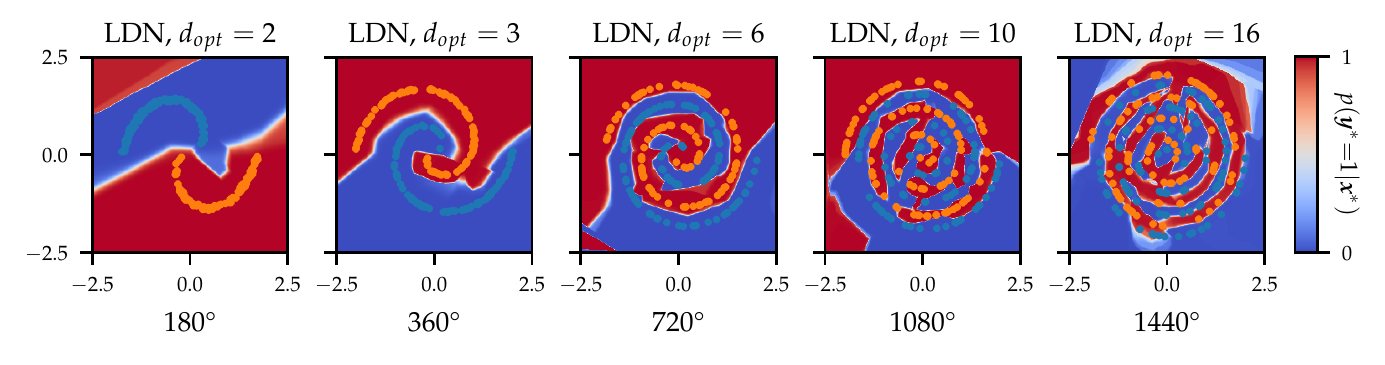}}
\vskip -0.1in
\caption[]{Functions learnt at each depth of a LDN on increasingly complex spirals. Note that single depth settings are being evaluated in this plot. We are not marginalising all layers up to $d_{opt}$.}
\label{fig:spiral_data_complex_input_scan}
\end{center}
\vskip -0.2in
\end{figure}

\begin{figure}[htb]
\begin{center}
\centerline{\includegraphics[width=0.9\textwidth]{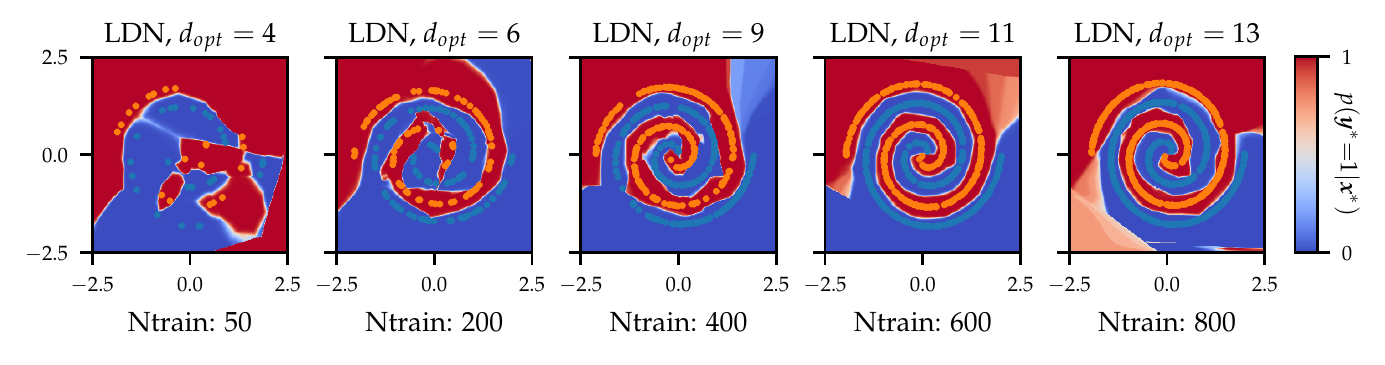}}
\vskip -0.1in
\caption[]{Functions learnt by LDNs trained on increasingly large spiral datasets. Note that single depth settings are being evaluated in this plot. We are not marginalising all layers up to $d_{opt}$.}
\label{fig:spiral_Ndata_input_scan}
\end{center}
\vskip -0.2in
\end{figure}

\begin{figure}[htb]
\begin{center}
\centerline{\includegraphics[width=0.75\textwidth]{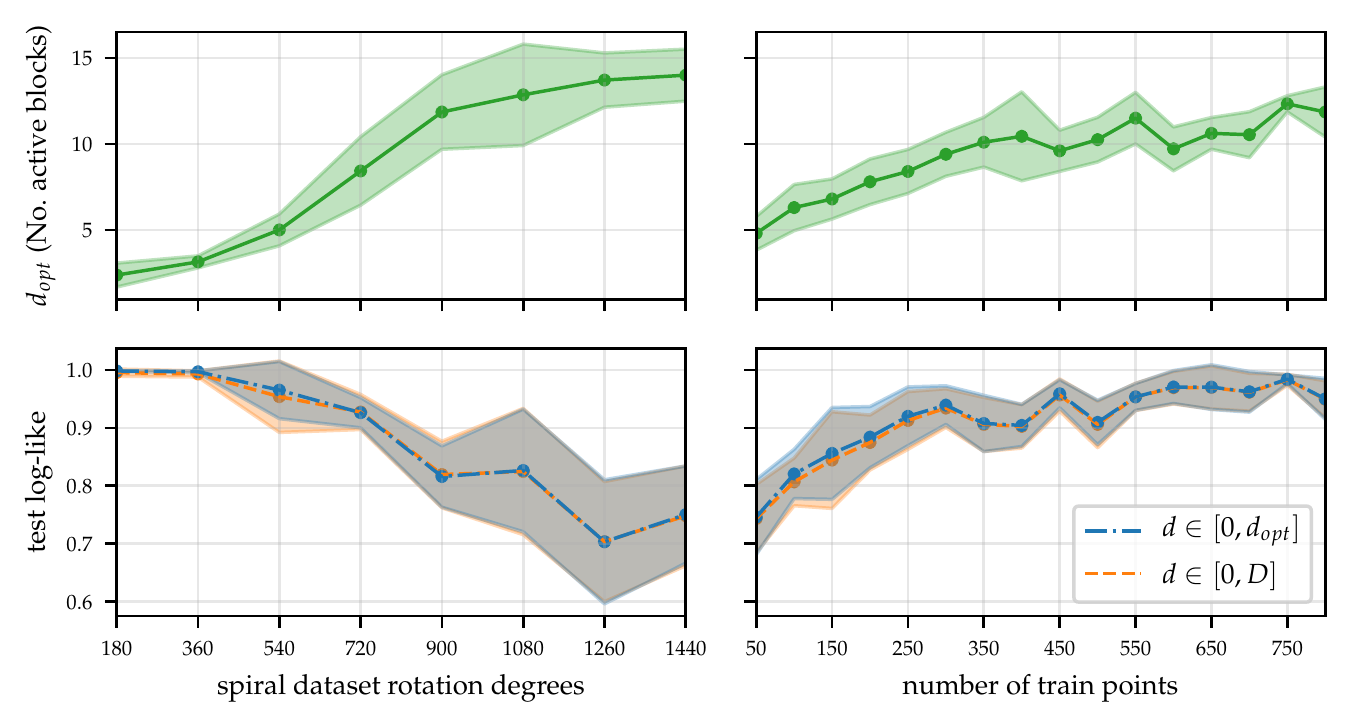}}
\vskip -0.1in
\caption[]{The left-side plots show the evolution of test performance and learnt depth as the data complexity increases. The right side plots show changes in the same variables as the number of train points increases. The results obtained when making predictions by marginalising over all $D\,{=}\,20$ layers overlap with those obtained when only using the first $d_{opt}$ layers. Best viewed in colour.}
\label{fig:rotation_Npoints}
\end{center}
\vskip -0.15in
\end{figure}

Setting $w$ back to $20$, we generate spiral datasets with varying degrees of rotation while keeping the number of train points fixed to $200$. In \cref{fig:rotation_Npoints}, we see how LDNs increase their depth to match the increasing complexity of the underlying generative process of the data. For rotations larger than 720\degree, $w\,{=}\,20$ may be excessively restrictive. Test performance starts to suffer significantly. \cref{fig:spiral_data_complex_input_scan} shows how our LDNs struggle to fit these datasets well.

Returning to 720\degree spirals, we vary the number of training points in our dataset. We plot the LDNs' learnt functions in \cref{fig:spiral_Ndata_input_scan}. LDNs overfit the 50 point train set but, as displayed in \cref{fig:spiral_data_complex_input_scan}, learn very shallow network configurations. Increasingly large training sets allow the LDNs to become deeper while increasing test performance. Around 500 train points seem to be enough for our models to fully capture the generative process of the data. After this point $d_{opt}$ oscillates around 11 layers and the test performance remains constant. Marginalising over $D$ layers consistently produces the same test performance as only considering the first $d_{opt}$. All figures are best viewed in colour.



\subsection{Small Image Datasets}

\begin{figure}[htbp]
\begin{center}
\centerline{\includegraphics[width=\textwidth]{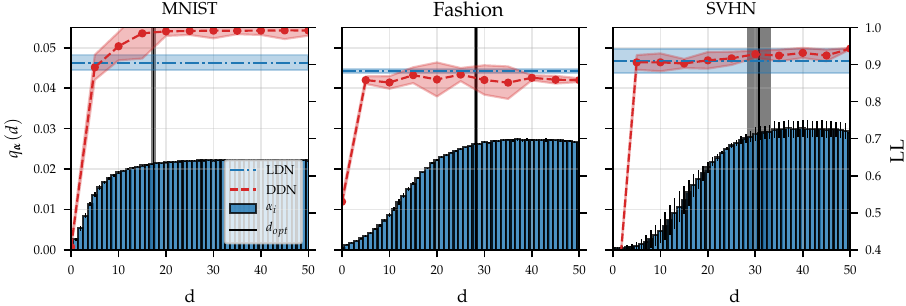}}
\vskip -0.1in
 \caption{Approximate posterior over depths for \glspl{LDN} of $D\,{=}\,50$ trained on image datasets. Test log-likelihoods obtained for \glspl{DDN} at various depths are overlaid with those from our \glspl{LDN} when marginalising over the first $d_{opt}$ layers. The depth was chosen using \cref{eqn:d_opt_heur95}}
\label{fig:image_depth_dist}
\end{center}
\vskip -0.2in
\end{figure}

We further evaluate \glspl{LDN} on MNIST, Fashion-MNIST and SVHN. 
Note that the network architecture used for these experiments is different from that used for experiments on the same datasets in \cref{sec:imgres} and \cref{app:additional_image_results}. It is described below. 
Each experiment is repeated 4 times to produce error bars. The results obtained with $D\,{=}\,50$ are shown in \cref{fig:image_depth_dist}. 
The larger size of these datasets diminishes the effect of the prior on the \gls{ELBO}.
Models that explain the data well obtain large probabilities, regardless of their depth.
For MNIST, the probabilities assigned to each depth in our \gls{LDN} grow quickly and flatten out around $d_{opt}\,{\approx}\,18$.
For Fashion-MNIST, depth probabilities grow slower. We obtain $d_{opt}\,{\approx}\,28$. For SVHN, probabilities flatten out around $d_{opt}\,{\approx}\,30$. These distributions and $d_{opt}$ values correlate with dataset complexity. 
In most cases, \glspl{LDN} achieve test log-likelihoods competitive with the best performing \glspl{DDN}.

\begin{figure}[htbp]
\begin{center}
\centerline{\includegraphics[width=\textwidth]{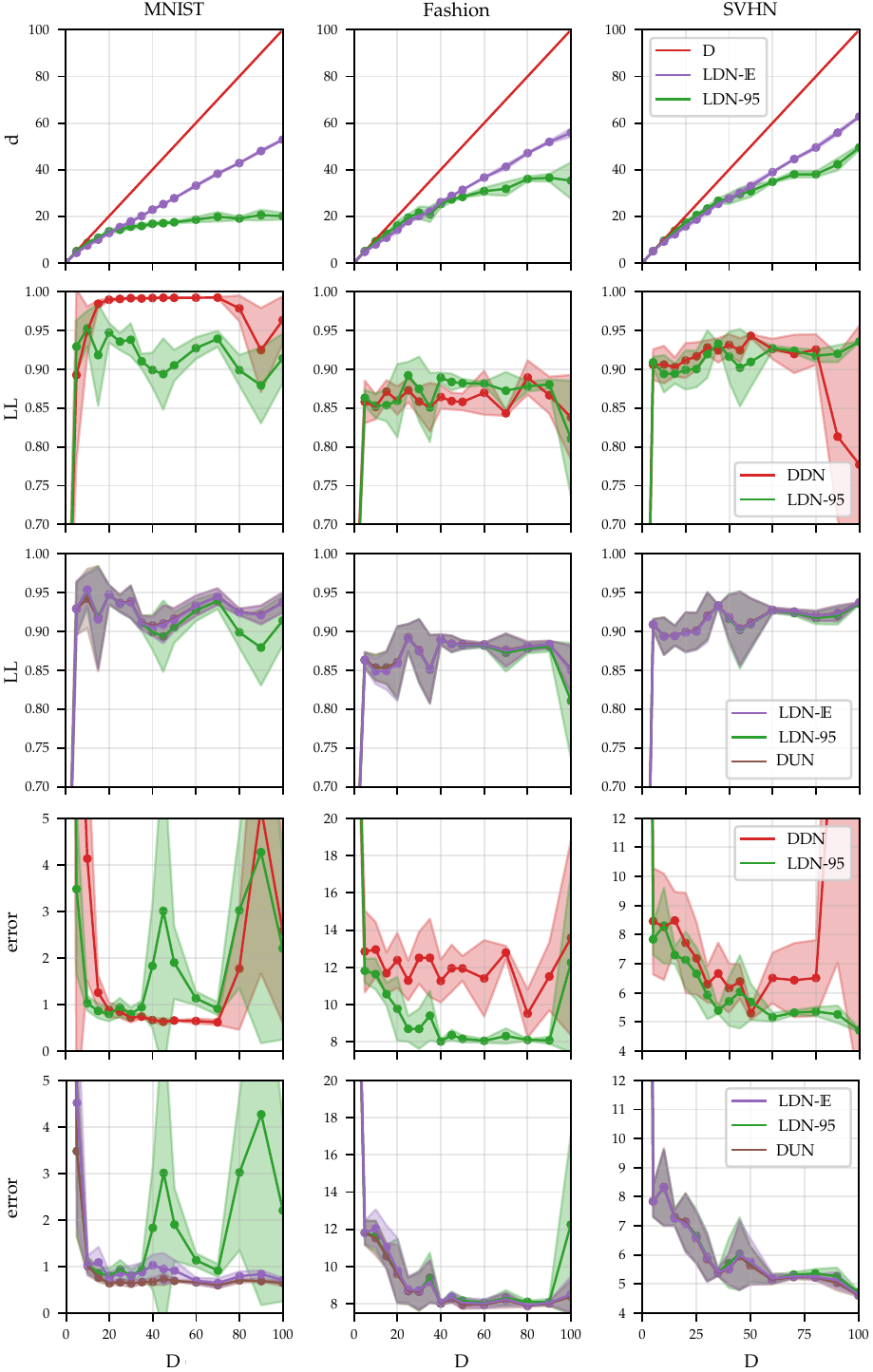}}
\vskip -0.13in
 \caption{\small{Comparisons of DDNs and LDNs using different pruning strategies and maximum depths. \emph{LDN-95} and \emph{LDN-$\mathbb{E}$} refer to the pruning strategy described in \cref{eqn:d_opt_heur95} and $d_\mathrm{opt} = \mathrm{round}(\EX[{q_{\balpha}}(d)])$, respectively. \nth{1} row: comparison of $d_\mathrm{opt}$. \nth{2} row: comparison of test log-likelihoods for DDNs and LDNs with 95 percent pruning. \nth{3} row: comparison of test log-likelihoods for LDN pruning methods. \nth{4} and \nth{5} rows: as above but for test error.}}
\label{fig:image_depth_dist2}
\end{center}
\vskip -0.2in
\end{figure}

\Cref{fig:image_depth_dist2} shows more detailed experiments comparing LDNs with DDNs on image datasets. We introduce expected depth $d_{opt}\,{=}\,\text{round}(\EX_{q_{\balpha}}[d])$ as an alternative to the \nth{95} percent heuristic introduced above. The first row of the figure adds further evidence that the depth learnt by LDNs corresponds to dataset complexity. For any maximum depth, and both pruning approaches, the LDN's learnt depth is smaller for MNIST than Fashion-MNIST and likewise smaller for Fashion-MNIST than SVHN. For MNIST, Fashion-MNIST and, to a lesser extent, SVHN the depth given by the \nth{95} percent pruning tends to saturate. On the other hand, the expected depth grows with $D$, making it a less suitable pruning strategy. 

As shown in rows 2 to 5, for SVHN and Fashion-MNIST, \nth{95} percentile-pruned LDNs suffer no loss in predictive performance compared to expected depth-pruned and even non-pruned LDNs. They outperform DDNs. For MNIST, \nth{95} percent pruning gives results with high variance and reduced predictive performance in some cases. Here, DDNs yield better log-likelihood and accuracy results. Expected depth is more resilient in this case, matching full-depth LDNs and DDNs in terms of accuracy.

\subsection{Experimental Setup for \gls{NAS}}
For experiments on the spirals dataset, our input $f_{0}$ and output $f_{D+1}$ blocks consist of linear layers. These map from input space to the selected width $w$ and from $w$ to the output size respectively. Thus, selecting $d=0 \Rightarrow b_{i}{=}0\,\forall i \in [1, D]$ results in a linear model.  The functions applied in residual blocks, $f_{i}(\cdot)\,\forall i \in [1, D]$, consist of a fully connected layer followed by a ReLU activation function and Batch Normalization~\citep{ioffe2015batch}.

Our architecture for the image experiments uses a $5{\times}5$ convolutional layer together with a $2{\times}2$ average pooling layer as an input block $f_{0}$. No additional down-sampling layers are used. The output block, $f_{D+1}$, is composed of a global average pooling layer followed by a fully connected residual block, as described in the previous paragraph, and a linear layer. The function applied in the residual blocks, $f_{i}(\cdot)\,\forall i \in [1, D]$, matches the preactivation bottleneck residual function described by \citet{he2016identity} and uses $3{\times}3$ convolutions. The outer number of channels is set to 64 and the bottleneck number is 32.

\FloatBarrier
\section{Implementing a DUN} \label{app:dun_impl}

In this section we demonstrate how to implement a \glspl{DUN} computational model by modifying standard feed-forward \glspl{NN} written in PyTorch. First, we show this for a simple \gls{MLP} and then for the more realistic case of a \gls{ResNet}, starting from the default PyTorch implementation.

\subsection{Multi-Layer Perceptron} \label{app:mlp_impl}

Converting a simple \gls{MLP} to a \gls{DUN} requires only around 8 lines of changes, depending on the specific implementation. Only 4 of these changes, in the \texttt{forward} function, are significant differences. The following listing shows the \texttt{git diff} of a \gls{MLP} implementation before and after being converted.

\begin{lstlisting}[style=pydiff]
 import torch
 import torch.nn as nn

 class MLP(nn.Module):
     def __init__(self, input_dim, hidden_dim, output_dim, num_layers):
         super(MLP, self).__init__()
 
+        self.output_dim = output_dim 

         layers = [nn.Sequential(nn.Linear(input_dim , hidden_dim),
                   nn.ReLU())]

         for _ in range(num_layers):
             layers.append(nn.Sequential(nn.Linear(hidden_dim, hidden_dim),
                           nn.ReLU()))

-        layers.append(nn.Linear(hidden_dim, output_dim))
+        self.output_layer = nn.Linear(hidden_dim, output_dim)

-        self.layers = nn.Sequential(*layers) 
+        self.layers = nn.ModuleList(layers)
 
     def forward(self, x):
+        act_vec = x.new_zeros(len(self.layers), x.shape[0], self.output_dim)

+        for idx, layer in enumerate(self.layers):
+            x = self.layers[idx](x)
+            act_vec[idx] = self.output_layer(x)

-        return self.layers(x)
+        return act_vec
\end{lstlisting}

\subsection{PyTorch ResNet} \label{app:resnet_impl}

To convert the official PyTorch \gls{ResNet} implementation\footnote{\url{https://github.com/pytorch/vision/blob/master/torchvision/models/resnet.py}} into a \gls{DUN}, we just need to make 17 changes. Many of these changes involve changing only a few characters on each line. Rather than looking at the whole file, which is over 350 lines long, we'll look only at the changes.

The first change that needs to be made is to the \texttt{\_make\_layer} function on line 177 of \texttt{resnet.py}. This function now needs to return a list of layers rather than a \texttt{nn.Sequential} container.
\begin{lstlisting}[style=pydiff]
                      base_width=self.base_width, dilation=self.dilati
                      norm_layer=norm_layer))
 
- return nn.Sequential(*layers)
+ return layers
\end{lstlisting}

With that change, we can modify the \texttt{\_\_init\_\_} function of the \texttt{ResNet} class on line 124 of \texttt{resnet.py}. We will create a \texttt{ModuleList} container to hold all of the layers of the \gls{ResNet}. This change has been made so that our \texttt{forward} function has access to the each layer individually. 

\begin{lstlisting}[style=pydiff]
                                 dilate=replace_stride_with_dilation[1]
  self.layer4 = self._make_layer(block, 512, layers[3], stride=2,
                                 dilate=replace_stride_with_dilation[2]
+ self.layers = nn.ModuleList(self.layer1 + self.layer2 + 
+                             self.layer3 + self.layer4)
\end{lstlisting}

Before implementing the forward function, we need to implement the adaption layers that ensure that inputs to the output block always have the correct number of filters. This is also done in the \texttt{\_\_init\_\_} function. Each adaption layer up-scales the number of filters by a factor of 2. Some layers need to have their outputs up-scaled multiple times which is kept track of by \texttt{self.num\_adaptions}.

\begin{lstlisting}[style=pydiff]
 self.avgpool = nn.AdaptiveAvgPool2d((1, 1))
 self.fc = nn.Linear(512 * block.expansion, num_classes)
 
+ self.num_adaptions = [0] * layers[0] + [1] * layers[1] + \
+                      [2] * layers[2] + [3] * layers[3]
+ adapt0 = nn.Sequential(
+     conv1x1(64*block.expansion, 128*block.expansion, stride=2),
+     self._norm_layer(128*block.expansion), self.relu)
+ adapt1 = nn.Sequential(
+     conv1x1(128*block.expansion, 256*block.expansion, stride=2),
+     self._norm_layer(256*block.expansion), self.relu)
+ adapt2 = nn.Sequential(
+     conv1x1(256*block.expansion, 512*block.expansion, stride=2),
+     self._norm_layer(512*block.expansion), self.relu)
+ adapt3 = nn.Identity()
+ self.adapt_layers = nn.Sequential(adapt0, adapt1, adapt2, adapt3)
\end{lstlisting}

The changes to the \texttt{\_forward\_impl} function on line 201 of \texttt{resnet.py} involve iterating over the layer list, up scaling layer outputs, and saving all of the activations of the output block. 

\begin{lstlisting}[style=pydiff]
  x = self.relu(x)
  x = self.maxpool(x)
 
- x = self.layer1(x)
- x = self.layer2(x)
- x = self.layer3(x)
- x = self.layer4(x)
+ act_vec = x.new_zeros(len(self.layers), x.shape[0], self.n_classes)
+ for layer_idx, layer in enumerate(self.layers):
+     x = layer(x)
+     y = self.adapt_layers[self.num_adaptions[layer_idx]:](x)
- x = self.avgpool(x)
+     y = self.avgpool(y)
- x = torch.flatten(x, 1)
+     y = torch.flatten(y, 1)
- x = self.fc(x)
+     y = self.fc(y)
+     act_vec[layer_idx] = y

- return x
+ return act_vec
\end{lstlisting}

The final change is to store the number of classes in the \texttt{\_\_init\_\_} function so that the \texttt{\_forward\_impl} function can pre-allocate a tensor of the correct size.

\begin{lstlisting}[style=pydiff]
  self.inplanes = 64
  self.dilation = 1
+ self.n_classes = num_classes
\end{lstlisting}

\section{Negative Results} \label{app:neg_res}

Here we briefly discus some ideas that seemed promising but were ultimately dead-ends.

\paragraph{Non-local Priors} 

\begin{figure}[htb]
    \centering
    \includegraphics[width=0.6\textwidth]{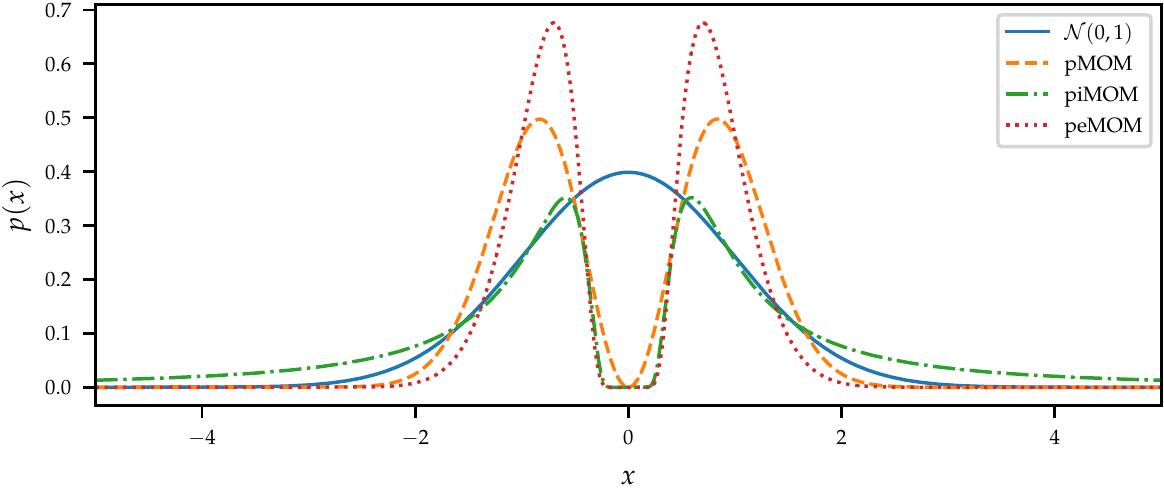}
    \caption{Comparison of non-local priors with the standard normal distribution.}
    \label{fig:mom_priors}
\end{figure}

These priors are ones which have zero density in the region of the \emph{null value} (often zero). Examples of such priors include the \gls{pMOM}, \gls{piMOM}, and \gls{peMOM} priors \citep{johnson2012bayesian, rossell2013high}, shown in \cref{fig:mom_priors}.

We attempted to train \glspl{DUN} with these priors, hoping that enforcing that each weight in the network was non-zero would, in turn, force each block of the \gls{DUN} to make a significantly different prediction to the previous block. 
Unfortunately training with non-local priors was unstable and resulted in poor performance. 

\paragraph{MLE training}

As described in \cref{app:vI_vs_MLE}, \gls{MLL} training on \glspl{DUN} tends to get stuck in local optima in which the posterior over depth collapses to a single arbitrary depth. In practice we found that \gls{VI} training greatly reduces this problem. 

\begin{wrapfigure}{r}{0.45\textwidth}
    \vspace{-0.1in}
  \begin{center}
    \includegraphics[width=0.45\textwidth]{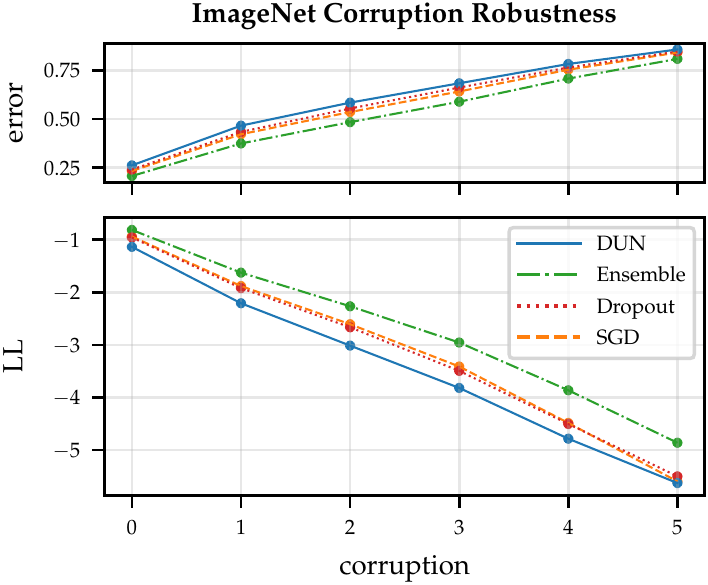}
  \end{center}
  \vspace{-0.1in}
  \caption{Error and LL for ImageNet at varying degrees of corruption. Due to computational costs only a single model was trained and evaluated for each method. As a result, we do not provide error bars.}
  \label{fig:imgnet_res}
  \vspace{-0.2in}
\end{wrapfigure}

\paragraph{Concat Pooling}

This technique combines the average and max pooling operations by concatenating their results. We tried to apply it before the final linear layer in ResNet-50. We suspected that for \glspl{DUN} based on \glspl{ResNet} this would be useful because the output block needs to work for predictions at multiple resolutions. Unfortunately, we found that the extra information provided by concat pooling over the standard average pooling resulted in strong overfitting. 


\paragraph{Scaling ResNet-50 to ImageNet}

We trained a ResNet-50 \gls{DUN} on the ImageNet dataset. However, in line with \citet{havasi2020training}, we found that a ResNet-50 does not have enough capacity to provide multiple explanations of the complex ImageNet dataset. As a result, the depth posterior for ResNet-50 invariably collapsed to a delta distribution. \gls{DUN} performance on ImageNet is poorer than standard SGD training, as shown in \cref{fig:imgnet_res}. Note that, while it is clear that \gls{DUN} performance in this setting is not strong, we only had enough computational resources to train each model one time. Without error bars, it is difficult to draw strong conclusions about the results. We hypothesize that a more heavily over-parameterised ResNet variant, such as ResNet-152 or a wide ResNet-50, would be able to support a \gls{DUN}.


\end{document}

%% file: glossary.tex
\newglossaryentry{DUN}
{
  name={DUN},
  description={Depth Uncertainty Network},
  first={\glsentrydesc{DUN} (\glsentrytext{DUN})},
  plural={DUNs},
  descriptionplural={Depth Uncertainty Networks},
  firstplural={\glsentrydescplural{DUN} (\glsentryplural{DUN})}
} 

\newglossaryentry{BNN}
{
  name={BNN},
  description={Bayesian Neural Network},
  first={\glsentrydesc{BNN} (\glsentrytext{BNN})},
  plural={BNNs},
  descriptionplural={Bayesian Neural Networks},
  firstplural={\glsentrydescplural{BNN} (\glsentryplural{BNN})}
}

\newglossaryentry{MLE}
{
  name={MLE},
  description={Maximum Likelihood Estimation},
  first={\glsentrydesc{MLE} (\glsentrytext{MLE})}
}

\newglossaryentry{VI}
{
  name={VI},
  description={Variational Inference},
  first={\glsentrydesc{VI} (\glsentrytext{VI})}
}

\newglossaryentry{MFVI}
{
  name={MFVI},
  description={Mean Field VI},
  first={\glsentrydesc{MFVI} (\glsentrytext{MFVI})}
}

\newglossaryentry{KLD}{
  name={KLD},
  description={Kullback-Leibler Divergence},
  first={\glsentrydesc{KLD} (\glsentrytext{KLD})}
}

\newglossaryentry{EP}
{
  name={EP},
  description={Expectation Propagation},
  first={\glsentrydesc{EP} (\glsentrytext{EP})}
}

\newglossaryentry{EM}
{
  name={EM},
  description={Expectation Maximisation},
  first={\glsentrydesc{EM} (\glsentrytext{EM})}
}

\newglossaryentry{MC}
{
  name={MC},
  description={Monte Carlo},
  first={MC}
}

\newglossaryentry{MCMC}
{
  name={MCMC},
  description={Markov Chain Monte Carlo},
  first={MCMC}
}

\newglossaryentry{BMA}
{
  name={BMA},
  description={Bayesian Model Averaging},
  first={\glsentrydesc{BMA} (\glsentrytext{BMA})}
}

\newglossaryentry{ELBO}
{
  name={ELBO},
  description={Evidence Lower BOund},
  first={\glsentrydesc{ELBO} (\glsentrytext{ELBO})}
}

\newglossaryentry{BO}
{
  name={BO},
  description={Bayesian Optimisation},
  first={\glsentrydesc{BO} (\glsentrytext{BO})}
}

\newglossaryentry{BOHB}
{
  name={BOHB},
  description={Bayesian Optimisation and Hyperband},
  first={\glsentrydesc{BOHB} (\glsentrytext{BOHB})}
}

\newglossaryentry{HPO}
{
  name={HPO},
  description={Hyperparameter Optimisation},
  first={\glsentrydesc{HPO} (\glsentrytext{HPO})}
}

\newglossaryentry{GP}
{
  name={GP},
  description={Gaussian Process},
  first={\glsentrydesc{GP} (\glsentrytext{GP})},
  plural={GPs},
  descriptionplural={Gaussian processes},
  firstplural={\glsentrydescplural{GP} (\glsentryplural{GP})}
}

\newglossaryentry{OOD}
{
  name={OOD},
  description={Out-of-distribution},
  first={\glsentrydesc{OOD} (\glsentrytext{OOD})}
}

\newglossaryentry{HMC}
{
  name={HMC},
  description={Hamiltonian Monte Carlo},
  first={\glsentrydesc{HMC} (\glsentrytext{HMC})}
}

\newglossaryentry{SGHMC}
{
  name={SGHMC},
  description={Stochastic Gradient HMC},
  first={\glsentrydesc{SGHMC} (\glsentrytext{SGHMC})}
}

\newglossaryentry{MLL}
{
  name={MLL},
  description={Marginal Log Likelihood},
  first={\glsentrydesc{MLL} (\glsentrytext{MLL})}
}

\newglossaryentry{NLL}
{
  name={NLL},
  description={Negative Log Likelihood},
  first={\glsentrydesc{NLL} (\glsentrytext{NLL})}
}

\newglossaryentry{LL}
{
  name={LL},
  description={Log Likelihood},
  first={\glsentrydesc{LL} (\glsentrytext{LL})}
}

\newglossaryentry{RMSE}
{
  name={RMSE},
  description={Root Mean Squared Error},
  first={\glsentrydesc{RMSE} (\glsentrytext{RMSE})}
}

\newglossaryentry{RCE}
{
  name={RCE},
  description={Regression Calibration Error},
  first={\glsentrydesc{RCE} (\glsentrytext{RCE})}
}

\newglossaryentry{TCE}
{
  name={TCE},
  description={Tail Calibration Error},
  first={\glsentrydesc{TCE} (\glsentrytext{TCE})}
}

\newglossaryentry{ECE}
{
  name={ECE},
  description={Expected Calibration Error},
  first={\glsentrydesc{ECE} (\glsentrytext{ECE})}
}

\newglossaryentry{NAS}
{
  name={NAS},
  description={Neural Architecture Search},
  first={\glsentrydesc{NAS} (\glsentrytext{NAS})}
}

\newglossaryentry{LDN}
{
  name={LDN},
  description={Learnt Depth Network},
  first={\glsentrydesc{LDN} (\glsentrytext{LDN})}
  plural={LDNs},
  descriptionplural={Learnt Depth Networks},
  firstplural={\glsentrydescplural{LDN} (\glsentryplural{LDN})}
}

\newglossaryentry{DDN}
{
  name={DDN},
  description={Determinisitc Depth Network},
  first={\glsentrydesc{DDN} (\glsentrytext{DDN})}
  plural={DDNs},
  descriptionplural={Determinisitc Depth Networks},
  firstplural={\glsentrydescplural{DDN} (\glsentryplural{DDN})}
}

\newglossaryentry{pMOM}
{
  name={pMOM},
  description={product moments},
  first={\glsentrytext{pMOM}}
}

\newglossaryentry{piMOM}
{
  name={piMOM},
  description={product inverse moments},
  first={\glsentrytext{piMOM}}
}

\newglossaryentry{peMOM}
{
  name={peMOM},
  description={product exponential moments},
  first={\glsentrytext{peMOM}}
}

\newglossaryentry{SGD}
{
  name={SGD},
  description={Stochastic Gradient Descent},
  first={\glsentrytext{SGD}}
}

\newglossaryentry{AI}
{
  name={AI},
  description={Artificial Intelligence},
  first={\glsentrytext{AI}}
}

\newglossaryentry{NN}
{
  name={NN},
  description={Neural Network},
  first={\glsentrydesc{NN} (\glsentrytext{NN})},
  plural={NNs},
  descriptionplural={Neural Networks},
  firstplural={\glsentrydescplural{NN} (\glsentryplural{NN})}
}

\newglossaryentry{CNN}
{
  name={CNN},
  description={Convolutional Neural Network},
  first={\glsentrytext{CNN}},
  plural={CNNs},
  descriptionplural={Convolutional Neural Networks},
  firstplural={\glsentryplural{CNN}}
}

\newglossaryentry{ResNet}
{
  name={ResNet},
  description={Residual Network},
  first={\glsentrytext{ResNet}},
  plural={ResNets},
  descriptionplural={Residual Networks},
  firstplural={\glsentryplural{ResNet}}
}

\newglossaryentry{MLP}
{
  name={MLP},
  description={Multi-Layer Perceptron},
  first={\glsentrytext{MLP}},
  plural={MLP},
  descriptionplural={Multi-Layer Perceptrons},
  firstplural={\glsentryplural{MLP}}
}

\newglossaryentry{CDF}
{
  name={CDF},
  description={Cumulative Density Function},
  first={\glsentrytext{CDF}},
  plural={CDFs},
  descriptionplural={Cumulative Density Functions},
  firstplural={\glsentryplural{CDF}}
}